\documentclass[a4paper,12pt]{article}

\RequirePackage{amsthm,amsmath,amsfonts,amssymb}
\RequirePackage[colorlinks,citecolor=blue,urlcolor=blue]{hyperref}
\RequirePackage{graphicx}

\usepackage[top=2cm,bottom=2cm,left=3cm,right=3cm,marginparwidth=1.75cm]{geometry}

\usepackage{booktabs}
\usepackage{mathtools}
\usepackage{mathabx}
\usepackage{graphicx,psfrag,epsf,multirow}
\usepackage{enumerate}

\usepackage{lmodern}
\usepackage{enumitem}
\usepackage{algorithm}
\usepackage{algpseudocode}
\usepackage{cleveref}
\usepackage{subcaption}
\usepackage[normalem]{ulem}
\usepackage{authblk}
\usepackage[numbers]{natbib}

\providecommand{\keywords}[1]
{
  \small	
  \textbf{\textit{Keywords: }} #1
}

\theoremstyle{plain}
\newtheorem{theorem}{Theorem}[section]

\newtheorem{corollary}{Corollary}[section]
\theoremstyle{remark}
\newtheorem{remark}{Remark}[section]
\newtheorem{assumption}{Assumption}[section]

\DeclareMathOperator{\dom}{\bf dom}
\DeclareMathOperator*{\argmin}{arg\,min}

\DeclareMathOperator{\tr}{\bf tr}

\DeclareMathOperator{\E}{\bf E}

\newcommand{\bx}{\boldsymbol{x}}
\newcommand{\by}{\boldsymbol{y}}

\newcommand{\bG}{\boldsymbol{G}}

\newcommand{\bI}{\boldsymbol{I}}

\newcommand{\bL}{\boldsymbol{L}}
\newcommand{\bM}{\boldsymbol{M}}

\newcommand{\bS}{\boldsymbol{S}}

\newcommand{\bX}{\boldsymbol{X}}
\newcommand{\bY}{\boldsymbol{Y}}
\newcommand{\bZ}{\boldsymbol{Z}}

\newcommand{\bGamma}{\boldsymbol{\Gamma}}
\newcommand{\bDelta}{\boldsymbol{\Delta}}

\newcommand{\bTheta}{\boldsymbol{\Theta}}

\newcommand{\bOmega}{\boldsymbol{\Omega}}
\newcommand{\bUpsilon}{\boldsymbol{\Upsilon}}

\newcommand{\bSigma}{\boldsymbol{\Sigma}}

\numberwithin{equation}{section}

\title{Learning Massive-scale Partial Correlation Networks in Clinical Multi-omics Studies with HP-ACCORD\thanks{Accepted for publication in Annals of Applied Statistics.}}
\date{September, 2025}

\author[1]{Sungdong~Lee}
\author[2]{Joshua~Bang}
\author[1]{Youngrae~Kim}
\author[1]{Hyungwon~Choi}
\author[2]{Sang-Yun~Oh}
\author[3]{Joong-Ho~Won\footnote{won.j@stats.snu.ac.kr}}

\affil[1]{Department of Medicine, National University of Singapore}
\affil[2]{Department of Statistics and Applied Probability, University of California, Santa Barbara}
\affil[3]{Department of Statistics, Seoul National University}

\begin{document}
\maketitle

\begin{abstract}
Graphical model estimation from multi-omics data requires a balance between statistical estimation performance and computational scalability. We introduce a novel pseudolikelihood-based graphical model framework 
that reparameterizes the target precision matrix while preserving the sparsity pattern and estimates it by minimizing an $\ell_1$-penalized empirical risk based on a new loss function. The proposed estimator maintains estimation and selection consistency in various metrics under high-dimensional assumptions. The associated optimization problem allows for a provably fast computation algorithm using a novel operator-splitting approach and communication-avoiding distributed matrix multiplication. A high-performance computing implementation of our framework was tested using simulated data with up to one million variables, demonstrating complex dependency structures similar to those found in biological networks. Leveraging this scalability, we estimated a partial correlation network from a dual-omic liver cancer data set. The co-expression network estimated from the ultrahigh-dimensional data demonstrated superior specificity in prioritizing key transcription factors and co-activators by excluding the impact of epigenetic regulation, thereby highlighting the value of computational scalability in multi-omic data analysis.  

\end{abstract}


\section{Introduction}\label{sec:intro}
Omics-scale technologies are valuable tools for surveying molecular features in biological samples such as tissues and biofluids in an unbiased manner. Individual omics modalities can generate high-dimensional data (large $p$) to characterize complex biological samples ($n \ll p$) and, more recently, isolated single cells or spatially connected micrometer-sized regions of tissue sections ($n \approx p$) \citep{vandereyken_2023}. 
Given the unprecedented opportunity to assemble multi-omic data as fine-resolution descriptors of gene expression regulation, cellular metabolism and signal transduction in a biological system, research studies in molecular medicine routinely employ a combination of omics platforms to study joint variation in the genome, epigenome, transcriptome, proteome and metabolome \citep{hasin2017, huang2017}.

In biomedical applications, it is customary to map multi-omic data to well-characterized biological pathways or networks for the integrative interpretation of results. In network-based analysis, data features are linked through a directed graph representing gene expression regulatory network, metabolic reaction pathways or signaling cascades, or an undirected graph such as protein-protein interaction network or gene co-expression network. Different networks facilitate the human interpretation of molecular interactions within their respective biological contexts. However, a downside of this approach is the requirement that network information relevant to a given study be available in the literature, covering all molecules measured in a study. This assumption is rarely satisfied in practice. In other cases, the biological network of interest may be adaptive, and it is therefore susceptible to rewiring under a specific condition, rendering the available static network information inapplicable. For these reasons, it is often necessary to complement existing networks by directly estimating new ones from data, ensuring that a realistic snapshot of the relational structure among variables is reflected in the downstream analysis or interpretation of the results. 

In omics-scale analysis, the biological network is typically represented as an adjacency matrix of data features based on marginal pairwise correlations. In weighted gene co-expression network analysis \citep[WGCNA,][]{zhang2005}, for example, Pearson correlation-based network modules are captured first, and the initial network is refined by topological analysis. Despite the success of WGCNA and other extensions in identifying gene communities from omics data, the adjacency of two molecular features as determined by marginal correlations may be a by-product of shared regulatory factors rather than a direct biological interaction. 
Hence, it is natural to estimate the co-expression network underlying a gene expression dataset by using a graph that represents the conditional dependencies, e.g., partial correlations. The same logic extends to the inference of other types of biological networks from multi-omic data, such as metabolic pathways and signaling cascades. 

It is well known that the precision matrix, or inverse covariance matrix, of a multivariate distribution encodes partial correlations. The most popular high-dimensional precision matrix estimation approach is the graphical lasso \citep{yuan2007model,friedman2008sparse}, which maximizes a multivariate Gaussian likelihood with an $\ell_1$-penalty. 
However, at the scale of modern multi-omics datasets, it is computationally infeasible to estimate desired networks using this strategy, mainly due to the computational bottleneck that is not easily resolved by using a more powerful computer. 
As we demonstrate later, our application data features close to 300,000 variables. In two high-performance computing (HPC) environments, the most scalable implementation of graphical lasso, BigQUIC \citep{hsieh2013big}, 
a scaled version of the quadratic approximation for sparse inverse covariance learning (QUIC) algorithm \citep{hsieh2011sparse},
could complete the computation only for the trivial cases in which the computed precision matrix estimates were diagonal.
Similarly, a fast implementation \citep{pang2014fastclime} of 
the constrained $\ell_1$-minimization for inverse matrix estimation \citep[CLIME,][]{cai2011constrained}, another popular methodology, was also not applicable on a HPC machine with 192GB of memory when the data set dimension exceeded 30,000. 
In the next section, we discuss in detail the inherent nature of the optimization problems from these popular methods that makes them difficult to scale up to modern omics-scale problems.

Feature screening procedures, which reduce the number of variables before the network estimation step, have been proposed as a remedy to address the lack of scalability of methods mentioned above \citep{luo2014sure,ZHENG2020104645}. 
Although feature screening could be suitable when a moderate dimensionality reduction is necessary, an order of magnitude reduction (300,000 or more to 30,000 or fewer features) precludes functionally important molecules from being considered for network estimation for computational tractability rather than a biologically motivated rationale.
It is also difficult to discern essential variables and dispensable ones \textit{a priori} solely based on numerical criteria chosen by an analyst; therefore, there is a clear merit in estimating the model with all possible input variables included from the start.

Another category of graphical model structure learning approach is based on optimizing an $\ell_1$-penalized pseudolikelihood-based objective function \citep{peng2009partial,khare2015convex}. Although the structure of pseudolikelihood-based objectives can be advantageous from an optimization perspective \citep{koanantakool2018communication}, existing estimators do not possess the statistical properties desirable for multi-omics analyses, as we also discuss in the next section.

In summary, a statistical framework for partial correlation graph estimation, scalable to contemporary multi-omics studies, is currently lacking. To address this gap, we propose a new estimation framework called ACCORD that aims to strike a balance between computational scalability and statistical performance. In ACCORD, target precision matrix is reparameterized and estimated by minimizing the $\ell_1$-penalized empirical risk based on a new loss function. The associated optimization problem enables a massively scalable and provably fast computation algorithm, achieved through a novel operator splitting algorithm and communication-avoiding distributed matrix multiplication. We then show that the ACCORD estimator has estimation consistency in $\ell_1$ and $\ell_2$ norms under standard high-dimensional assumptions and selection consistency under an irrepresentability condition. In this way, we strike a balance between statistical performance and computational scalability in massively large-scale settings. 

We demonstrate that the HPC implementation of ACCORD, termed HP-ACCORD, scales well to handle dimensions up to one million. Leveraging on the scalability, we estimate a partial correlation network in the multi-omic liver cancer dataset from The Cancer Genome Atlas (TCGA). Using a combination of epigenomic and transcriptomic data sets with a total of 285,358 variables, we successfully recapitulated the local and global correlation structures of the variables within the same omics platform and identified a network of co-transcribed genes and DNA methylation events in upstream regulatory regions. More importantly, we show that the graph estimated from the ultrahigh-dimensional dual-omic data enabled us to identify \textit{bona fide} transcription factors driving the co-expression network with greater specificity than and equivalent sensitivity to the alternative graph estimated from the transcriptomic data with 15,598 variables only. The analysis clearly highlights the merit of performing graph estimation in the whole feature space. 

The organization of this paper is as follows. \Cref{sec:challenges} explains the challenges that existing methods face when the scale of the problem becomes massively large. \Cref{sec:alg} introduces the ACCORD framework. In \Cref{sec:meth} we study the statistical properties of the ACCORD estimator. 
\Cref{sec:exp} is devoted to numerical experiments demonstrating the performance of the estimator and the scalability of the algorithm. In \Cref{sec:HCC}, we showcase HP-ACCORD through graph estimation in a liver cancer data set with 285,358 multi-omic variables. The paper is concluded in \Cref{sec:disc}.

\section{Challenges}\label{sec:challenges}

In this section, we first describe the computational hurdles that current techniques for high-dimensional precision matrix estimation grapple with when dealing with extremely large scales.

Graphical lasso solves the convex optimization problem
    $\min_{\bTheta\in\mathbb{S}^p}\{-\log\det \bTheta + \tr(\bS\bTheta) + \lambda \Vert \bTheta \Vert_{1}\}$, 
    $\lambda > 0$,
in order to estimate the precision matrix $\bTheta^*$ of a $p$-variate zero-mean distribution.
Here  $\mathbb{S}^p$ is the space of $p\times p$ symmetric matrices, 
$\bS = (1/n)\bX^T\bX$ is the sample covariance matrix, 
where 
$\bX = [X_1, X_2, \cdots, X_n]^T$ with $X_i \in \mathbb{R}^p$ is the centered data matrix, and
$\Vert\cdot\Vert_1$ is the vector $\ell_1$ norm. 
The smooth part of the objective is a simplified form of the negative log-likelihood of a zero-mean normal distribution; the $\ell_1$ penalty promotes the sparsity of the estimate.%
The estimate is fully characterized by the Karush-Kuhn-Tucker (KKT) optimality condition
\begin{equation}\label{eq:KKT}
    -\bTheta^{-1} + \bS + \lambda \bZ = \mathbf{0}, \quad \bZ \in \partial\Vert\bTheta\Vert_1,
\end{equation}
where $\partial\|\bTheta\|_1$ denotes the subdifferential of the convex function $\bX \mapsto \|\bX\|_1$ at $\bTheta$.
The vast literature on the algorithms for graphical lasso, e.g., 
\citet{d2008first,friedman2008sparse,li2010inexact,hsieh2011sparse,hsieh2014quic}, 
essentially reduces to how to solve the KKT equation \eqref{eq:KKT} iteratively.

The computational culprit in \eqref{eq:KKT} is the inverse of the $p \times p$ matrix variable $\bTheta$, which has to be computed every iteration.
Inverting a $p \times p$ matrix using direct methods (e.g., Cholesky decomposition) costs $O(p^3)$ arithmetic operations in general, which becomes prohibitive if $p$ is at the omics scale of a few hundred thousand. 
Storage requirements worsen the situation. The inverse $\bTheta^{-1}$ is in general not sparse, requiring $\Omega(p^2)$ memory space. For example, if $p=300,000$, then roughly 700GB of memory is needed, which calls for distributed computation.
Being inherently sequential, however, matrix inversion is difficult to parallelize or distribute, meaning that direct methods are not scalable even with high-performance computing (HPC) systems that have distributed memory.

Consequently, most existing algorithms resort to employing an (inner) iterative method to compute $\bTheta^{-1}$. 
For instance, \citet{friedman2008sparse} invokes a $p$-dimensional lasso regression solver $p$ times every (outer) iteration to compute one column of $\bTheta^{-1}$ at a time, which becomes prohibitively expensive in omics-scale problems.
%
BigQUIC  \citep{hsieh2013big} updates $\bTheta$ by block coordinate descent and then computes $\bTheta^{-1}$ column by column, by storing only a small fraction of the $p$ columns of $\bTheta^{-1}$ in a cache and recomputing the missing columns on demand;
the recomputation  solves a fraction of the $p$ linear equations defining the inverse ($\bTheta \bX = \bI_p$) by using conjugate gradient.
Since this strategy is only successful if the cache miss rate is low, BigQUIC partitions $\bTheta$ into a 
block matrix and permutes blocks to minimize the number of off-block-diagonal elements. 
Therefore, the scalability of BigQUIC is limited by the maximum degree 
of the underlying graph. For more complex graph structures of interest, e.g., those that arise in multi-omics studies, computable dimensions are practically less than $100,000$; see \S\ref{sec:exp}.

CLIME minimizes $\|\bTheta\|_1$ subject to the constraint $\|\bS\bTheta - \bI_p\|_{\infty} \leq \lambda$ over $\mathbb{R}^{p\times p}$, where $\|\cdot\|_{\infty}$ is the vector $\ell_{\infty}$ norm. 
This constraint is a margin-allowed version of the estimating equation $\bS\bTheta = \bI_p$ in case $\bS$ is singular.%
The optimization problem of CLIME can be decomposed into $p$ independent, $p$-dimensional linear programming (LP) problems, 
each solving a column of $\bTheta$. 
While LP is the most extensively studied convex optimization problem, commercial LP solvers like Gurobi or Cplex, which utilize interior-point or simplex methods,
struggle with scalability when handling hundreds of thousands of variables \citep{mazumder_computing_2019}.
Exacerbating the situation, for omics-scale CLIME we are required to solve hundreds of thousands of large-scale LP problems. This remains an almost unattainable task 
despite the independence. 
As a result, \texttt{fastclime} \citep{pang2014fastclime,pang_parametric_2017}, 
which could not handle more than 30,000 variables in our problem instance,
remains one of the most scalable implementations publicly available.

As the most scalable pseudolikelihood-based method to date,
the convex correlation selection method
\citep[CONCORD,][]{khare2015convex} solves the $\ell_1$-penalized minimization problem
\begin{equation}\label{eq:qcon}
    \min_{\bTheta\in\mathbb{S}^p} \{- \log\det \bTheta_D + ({1}/{2}) \tr (\bTheta^2\bS) + \lambda \Vert \bTheta_X \Vert_1\},
\end{equation}
where $\bTheta_D$ denotes the diagonal and $\bTheta_X$ denotes the off-diagonal parts of $\bTheta$.
It is a convex amendment of a non-convex pseudolikelihood-based method SPACE \citep{peng2009partial}, which 
in turn extends the node-wise regression approach by \citet{meinshausen2006high}
to estimate $\bTheta^*$.\footnote{CLIME can be understood as an extension of the latter in a different direction, with the aforementioned scalability bottleneck.}
The associated KKT condition is
\begin{equation}\label{eqn:concord_kkt}
    -\bTheta_D^{-1} + ({1}/{2})(\bTheta\bS + \bS\bTheta) + \lambda \bZ = \mathbf{0}, \quad \bZ \in \partial\Vert\bTheta_X\Vert_1.
\end{equation}
Compared with that for graphical lasso \eqref{eq:KKT}, KKT condition \eqref{eqn:concord_kkt} only involves the inverse of the diagonal matrix $\bTheta_D$, which is trivial to compute. 
While the second term in \eqref{eqn:concord_kkt} costs two $p \times p$ matrix-matrix multiplications with $O(p^3)$ operations,
they are much easier to distribute and parallelize than matrix inversion. 
The sparsity of the optimization variable $\bTheta$ imposed by the $\ell_1$ penalty further reduces the complexity of the multiplication. 
\citet{koanantakool2018communication} leverage these features to achieve massive scalability in distributed memory HPC systems. This HP-CONCORD implementation can handle 320,000-dimensional data drawn from a random graph structure resembling multi-omic networks.

The CONCORD method, while innovative, has significant room for improvement on both computational and statistical fronts. Statistically, the CONCORD approach is not consistent in estimating $\bTheta^*$, which provides key information on the strength of the edges in the underlying graph. 
The objective function in \eqref{eq:qcon}  defines a loss function $L(\bTheta, x) = -\log\det\bTheta_D + \frac{1}{2}\tr(x^T\bTheta^2x)$ for a sample $x\in\mathbb{R}^p$. 
The population risk $R(\boldsymbol{\Theta}) = \E L(\boldsymbol{\Theta}, X)$ is minimized by a $\boldsymbol{\Theta}$ satisfying equation 
\begin{equation}\label{eqn:concordestimating}
    \bTheta_D^{-1} - (1/2)(\bTheta\bSigma^* + \bSigma^*\bTheta)= \mathbf{0},
\end{equation}
where $\bSigma^* = {\bTheta^*}^{-1}$ is the covariance matrix of the random vector $X$.
However, the true precision matrix $\boldsymbol{\Theta}^*$ does \emph{not} minimize the risk unless $\boldsymbol{\Theta}_D^*$ is equal to the identity---a restrictive assumption that is unlikely to hold in practice. 
The currently available result on the consistency of the CONCORD estimator \citep[Theorem 2,][]{khare2015convex} 
requires an accurate estimator $\widehat{\boldsymbol{\Theta}_D}$ of $\bTheta^*_D$
with a rate $\max_{i=1,\dotsc,p} | \widehat{\boldsymbol{\Theta}_D}_{,ii} - \boldsymbol{\Theta}^*_{D,ii} | = O_P(\sqrt{\log n /n})$. 
Similarly, SPACE also requires a separate estimator of $\bTheta_D^*$ with the same rate.
Such a separate estimator is difficult to find in practice.
In this context, the precise connection between CONCORD parameter estimates and the partial correlation remains ambiguous.
Computationally, CONCORD-ISTA algorithm \citep{oh2014optimization} possesses a sublinear $O(1/t)$ rate of convergence, where $t$ is the number of iterations and the convergence is measured in terms of the objective function value. In addition to the sublinear convergence of the objective function, the variable iterate $\{\bTheta^{(t)}\}$ may converge to the minimizer $\hat{\bTheta}$ of \eqref{eq:qcon} in an arbitrarily slow rate \citep{Bauschke:ConvexAnalysisAndMonotoneOperatorTheoryIn:2011}.
Since the algorithm has to be terminated within a finite number of iterations, statistical error of $\bTheta^{(t)}$ (say $\|\bTheta^{(t)}-\bTheta^*\|$) may remain quite large compared to that of $\hat{\bTheta}$ (say $\|\hat{\bTheta}-\bTheta^*\|)$ even after a large number of iterations.

In the remainder of this paper, we propose in detail a novel approach to address these challenges and illustrate its usefulness through extensive numerical experiments and an analysis of high-dimensional multi-omics data.

\section{The ACCORD Estimator and HP-ACCORD}\label{sec:alg}
\subsection{ACCORD loss and estimator}

For a matrix $\bOmega \in \mathbb{R}^{p\times p}$, consider the following loss function
\begin{equation}\label{eqn:accordloss}
L(\bOmega, x) = -\log\det\bOmega_D + (1/2)\tr(\bOmega^T\bOmega x x^T)
.
\end{equation}
For the associated risk $R(\bOmega) := \E L(\bOmega, X) = -\log\det\bOmega_D
+ \frac{1}{2}\tr(\bOmega^T\bOmega\bSigma^*)$ with the essential domain 
$\dom R =  \{(m_{ij})\in\mathbb{R}^{p\times p}: m_{ii} > 0,~i=1,\dotsc, p\} \cap \mathbb{R}^{p\times p}$
on which $R$ is finite,
the following theorem holds (details are provided in the Supplemental Material).
\begin{theorem}\label{thm:p-risk}
    The ACCORD risk $R(\bOmega)$ is uniquely minimized by $\bOmega^* := {\bTheta^*_D}^{-1/2}\bTheta^*$.
\end{theorem}

In light of \Cref{thm:p-risk}, define the transformation
$$
    T: \bTheta \mapsto \bTheta_D^{-1/2}\bTheta
$$
from $\dom R$ to itself.
One important property of $T$ is that it is continuous and bijective with inverse $T^{-1}: \bOmega \mapsto \bOmega_D\bOmega$.
In other words, the population risk is uniquely minimized by a one-to-one transformation $T$ of the true precision matrix $\bTheta^*$.
In the sample version of $R$ where the sample covariance matrix $\bS$ replaces the $\bSigma^*$, the unique minimizer of the empirical risk $\hat{R}_n(\bOmega) = -\log\det \bOmega_D + \frac{1}{2}\tr(\bOmega^T\bOmega \bS)$ is given by 
$$
    \hat{\bOmega} = T(\tilde{\bTheta}) = \tilde{\bTheta}_D^{-1/2}\tilde{\bTheta},
    \quad
    \tilde{\bTheta} = \bS^{-1}
    .
$$
in the low-dimensional regime where $\bS$ is positive definite.
Then, $\hat{\bOmega} \to \bOmega^*$ as $\bS\to\bSigma^*$ by the continuous mapping theorem. Also, if we let $\hat{\bTheta} = T^{-1}(\hat{\bOmega}) =  \hat{\bOmega}_D\hat{\bOmega}$, then $\hat{\bTheta}\to\bTheta^*$.
Likewise, the partial correlations $(\rho^*_{ij})$ can be consistently estimated with
$\hat{\rho}_{ij} = -(1/2)(\hat{\omega}_{ij}/\hat{\omega}_{jj} + \hat{\omega}_{ji}/\hat{\omega}_{ii})$
when $\hat{\bOmega} = (\hat{\omega}_{ij})$, using the relation $\rho^*_{ij} = -\theta^*_{ij}/\sqrt{\theta^*_{ii}\theta^*_{jj}}$ and $\theta^*_{ij} = \omega^*_{ii}\omega^*_{ij}$.

In a high-dimensional setting,  
we can expect minimizing an $\ell_1$-penalized sample average of \eqref{eqn:accordloss} estimates $\bOmega^*=T(\bTheta^*)$ consistently under the usual sparsity assumption on $\bTheta^*$:
\begin{equation}\label{eqn:accord}
    \hat{\bOmega} 
    =
    \argmin\nolimits_{\bOmega\in\mathbb{R}^{p\times p}}\left\{
    -\log\det\bOmega_D
    + (1/2)\text{tr}(\bOmega^T\bOmega\bS) 
    + 
    \lambda\|\bOmega\|_1
    \right\}
    .
\end{equation}
The $\ell_1$ penalization on $\bOmega$ in \eqref{eqn:accord} is justified by another important property that $T$ and $T^{-1}$ preserve the support of their arguments; if $\omega_{ij} = [T(\bTheta)]_{ij}$ with $\bTheta=(\theta_{ij})$, then $\omega_{ij} = \theta_{ij}/\sqrt{\theta_{ii}}$ and $\theta_{ij} = \omega_{ii}\omega_{ij}$. Therefore,
$\omega_{ij} = 0$ if and only if $\theta_{ij}=0$,
and hence, $\bOmega^*$ and $\bTheta^*$ share the same sparsity pattern.

The loss \eqref{eqn:accordloss} and estimator \eqref{eqn:accord} resemble those of CONCORD in \S\ref{sec:challenges}.
The critical difference is that the argument $\bOmega$ is allowed to be asymmetric in our proposal. This simple change makes the estimator consistent 
and recover the support of $\bTheta^*$
(at least in the low-dimensional regime; high-dimensional results are presented in \S\ref{sec:meth}
), a feature absent in CONCORD. (The latter estimates a root of \eqref{eqn:concordestimating}. Let alone uniqueness, the existence of its closed form expression is unclear, and it is unlikely that it preserves the sparsity pattern of $\bTheta^*$.)
Due to the asymmetric nature of our approach, we name the loss function \eqref{eqn:accordloss} the ACCORD loss, and the estimator \eqref{eqn:accord} the ACCORD estimator, after Asymmetric ConCORD.

The KKT condition for the convex optimization problem for ACCORD \eqref{eqn:accord} is
\begin{equation}\label{eqn:accord_kkt}
-\bOmega_D^{-1}
+
\bOmega \bS 
+\lambda \bZ = \mathbf{0}, \quad \bZ \in \partial\left\|\bOmega\right\|_1,
\end{equation}
which conserves the computational attraction of CONCORD; cf. \eqref{eqn:concord_kkt}. Furthermore, there is only one matrix multiplication in \eqref{eqn:accord_kkt}, suggesting simpler computation.

For the ACCORD estimator \eqref{eqn:accord} to be well-defined, a solution to \eqref{eqn:accord_kkt} needs to be unique. 
A matrix $\bX \in \mathbb{R}^{n \times p}$ $(p \geq n)$ is said to have columns in general position if the affine span of any $n$ points $\{s_1X_{i_1}, \cdots, s_n X_{i_n}\}$, for arbitrary signs $s_1, \cdots, s_n \in \{-1, 1\}$, does not contain any element of $\{\pm X_i : i \neq i_{1},\cdots,i_{n}\}$. General position occurs almost surely if $\bX$ is drawn from a continuous distribution \citep{tibshirani2013lasso}.
\begin{theorem}\label{thm:uniqueness}
    Suppose that the data matrix $\bX \in \mathbb{R}^{n \times p}$ has columns in general position. Then, the ACCORD estimator \eqref{eqn:accord} with $\bS=(1/n)\bX^T\bX$ is unique.
\end{theorem}

\subsection{ACCORD-FBS algorithm}\label{sec:alg:accord-fbs}
For the computation of the ACCORD estimator, we use forward-backward splitting (FBS), also known as the proximal gradient method  \citep{combettes2007douglas, combettes2011proximal}. 
Let us split the ACCORD objective function $f$ into $f = g + h$ where
\begin{equation}\label{eqn:splitting}
	g(\bOmega) = ({1}/{2})\tr(\bOmega^T\bOmega\bS) 
	, \quad
	h(\bOmega) = -\log\det\bOmega_D + \lambda\Vert \bOmega \Vert_{1}
    .
\end{equation}
and apply a forward step 
for $g$:
$\bOmega^{(t+1/2)} = \bOmega^{(t)} - \tau_t\nabla g(\bOmega^{(t)})$,
and a backward (proximal) step 
for $h$:
$\bOmega^{(t)} = \argmin_{\bOmega\in\mathbb{R}^{p\times p}}\{h(\bOmega) + \frac{1}{2\tau_t}\|\bOmega - \bOmega^{(t+1/2)}\|_F^2\}$
to obtain a sequence of optimization variables $\{\bOmega^{(t)}=(\omega_{ij}^{(t)})\}$.
At the element level, we have the following closed-form iteration:
\begin{equation}\label{eqn:proxgrad}
	\begin{split}
	\omega^{(t+1)}_{ii} &= (y - \tau_t \lambda + \sqrt{(y - \tau_t\lambda)^2 + 4\tau_t}) / 2,
	\quad
	y = \omega^{(t)}_{ii} - \tau_t [\bOmega^{(t)}\bS]_{ii} 
	\\
	\omega^{(t+1)}_{ij} &= S_{\tau_t\lambda}(\omega^{(t)}_{ij} - \tau_t [\bOmega^{(t)} \bS]_{ij})
	, \quad
	i \neq j,
	\end{split}
\end{equation}
where $S_{a}(x) = (|x| - a)_+\operatorname{sign}(x)$ is the soft-thresholding operator.

Theory of FBS ensures that with the choice of the step size $\tau_t \in (0, 2/L)$, where $L = \sigma_{\max}(\bS)$ is the maximum singular value of $\bS$, the iterate sequence $\{\bOmega^{(t)}\}$ converges to the target $\hat{\bOmega}$ \citep{Bauschke:ConvexAnalysisAndMonotoneOperatorTheoryIn:2011}.
The $L$ coincides with the global Lipschitz constant of $\nabla g(\bOmega) = \bOmega\bS$ in the Euclidean (Frobenius) norm.
Backtracking line search adaptively finds a maximal step size $\tau_t$ for each iteration $t$ in such a way the convergence is guaranteed \citep{Beck:SiamJournalOnImagingSciences:2009}.
\Cref{alg:pgd}, named ACCORD-FBS, describes both fixed step size and backtracking FBS algorithms for ACCORD.
The step size satisfies $\tau_{t} \geq \min\{\tau_0, \beta/L\}$ for $\beta \in (0, 1)$ and the descent property $f(\bOmega^{(t+1)}) \leq f(\bOmega^{(t)})$ is guaranteed by the majorization-maximization (MM) principle \citep{lange2022nonconvex}.

\begin{algorithm}[t!]
    \caption{ACCORD-FBS}\label{alg:pgd}
    \begin{algorithmic}
    \State \textbf{Input}: sample covariance $\bS \in \mathbb{R}^{p\times p}$, minimum step size $1/L$, initial step size $\tau_0$, line search parameter $0< \beta < 1$, initial $\bOmega^{(0)}$
        \For{$t$ in $0, 1, 2,\cdots$}
            \For{$\tau_t$ in $\tau_0, \beta\tau_0, \beta^2\tau_0, \cdots$}
                \State $\nabla g(\bOmega^{(t)}) \gets \bOmega^{(t)}\bS$
                \Comment{Use \Cref{alg:1dmm}} for HP-ACCORD
                \State Update $\bOmega^{(t)}$ according to \eqref{eqn:proxgrad}
                \State $\bDelta \gets \bOmega^{(t+1)} - \bOmega^{(t)}$
            \EndFor{ if $g(\bOmega^{(t+1)}) \leq g(\bOmega^{(t)}) + \langle \bDelta, g(\bOmega^{(t)}) \rangle + \frac{1}{2\tau_t}\Vert \bDelta \Vert_F^2$ or $\tau_t \leq 1/L$}
        \EndFor{ until converge}
    \State \textbf{Output}: estimate $\hat{\bOmega} \gets \bOmega^{(t+1)}$
    \end{algorithmic}
\end{algorithm}

While the general convergence rate of FBS is $O(1/t)$ in the objective 
value unless the objective is strongly convex (which is not the case in ACCORD), we can nevertheless show that ACCORD-FBS converges linearly in both the objective and variable iterates.

\begin{theorem}\label{thm:linearconvergence}
Assume the condition for \Cref{thm:uniqueness},
and that the iterate sequence $\{\bOmega^{(t)}\}$ is generated by \Cref{alg:pgd} with the step size sequence $\{\tau_t\}$ satisfying either i) $\beta=1$, $\tau_0 \in [\underline{\tau}, \bar{\tau}]$, where $0 < \underline{\tau} \leq \bar{\tau} < 2/L$, or ii) $0 < \beta < 1$. 
Then, the objective value sequence $\{f(\bOmega^{(t)})\}$ converges to the minimum $f^{\star} = f(\hat{\bOmega})$ monotonically.
Furthermore, if the initial iterate $\bOmega^{(0)}$ is chosen such that 
$f(\bOmega^{(0)}) \leq f(\bI_p)= \tr(\bS)/2 + \lambda p$,
then 
the following holds.
\begin{equation}\label{eqn:complexitybound}
\begin{split}
	f(\bOmega^{(t)}) - f^{\star}
	&\leq 
	\left(\frac{1}{1 + 2a\sigma}\right)^t [f(\bOmega^{(0)}) - f^{\star}]
	, 
	\quad
	t \geq 0,
	\\
	\Vert \bOmega^{(t)} - \hat{\bOmega} \Vert_F
	&\leq
	\left(\frac{1}{\sqrt{1 + 2a\sigma}}\right)^{t-1}
	\left(1 + \frac{1}{a\sigma\sqrt{1 + {1}/{(2a\sigma)}}}\right)
	\sqrt{{[f(\bOmega^{(0)}) - f^{\star}]}/{a}}
	,
	\quad
	t \geq 1
 ,
\end{split}
\end{equation}
where
	$a = {1}/{\bar{\tau}} - {L}/{2} > 0$,
	$\sigma = [4\kappa(1/\underline{\tau} + L)^2]^{-1}$
in case i, and
	$a = [2\tau_0]^{-1}$,
	$\sigma = [4\kappa(1/\tau_{\min} + L)^2]^{-1}$,
	$\tau_{\min} = \min\{\tau_0, \beta/L\}$
in case ii.
The constant $\kappa$ is explicit and  depends only on 
$\bS$. 
\end{theorem}

Splitting \eqref{eqn:splitting}
is crucial for establishing linear convergence in that the global Lipschitz constant $L < \infty$ of $\nabla g$ exists, and $1/L > 0$ serves as the lower bound of the step size that guarantees the descent property.
Following CONCORD-ISTA \citep{oh2014optimization}, we may also split $f$ into $\tilde{g}(\bOmega) =  -\log\det\bOmega_D + ({1}/{2})\tr(\bOmega^T\bOmega\bS)$ 
and
$\tilde{h}(\bOmega) = \lambda\Vert \bOmega \Vert_{1}$ 
and call it ACCORD-ISTA. 
The problem with this splitting is that $\nabla\tilde{g}$ is not globally Lipschitz. 
The consequence is that the resulting step sizes satisfying the descent condition can be arbitrarily small, contributing to slow convergence of the iterates; 
see \S\ref{sec:challenges} and \S\ref{sec:exp}. 
\begin{remark}
    In \Cref{alg:pgd}, the computational complexity of each iteration is $O(np^2)$, with the bottleneck being the gradient computation $\nabla g(\bOmega^{(t)}) \gets (1/n)(\bOmega^{(t)}\bX^T)\bX$;
    see Supplementary Material C for comparison of the complexity with other methods.
    In fact, the required number of arithmetic operations can be further reduced by exploiting the sparsity of the iterate $\bOmega^{(t)}$.
    Moreover, matrix multiplication of the type $\bOmega^{(t)}\bX^T$ can be distributed among multiple computational nodes, allowing the algorithm to be more efficient in HPC environments.
\end{remark}

\subsection{HP-ACCORD: HPC implementation of ACCORD}\label{sec:alg:hp-accord}
\begin{algorithm}[t!]
    \caption{One-dimensionally distributed matrix multiplication (1DMM)}\label{alg:1dmm}
    \begin{algorithmic}
    \State \textbf{Input}: Partition $\bOmega = [\bOmega_1^T, \cdots, \bOmega_P^T]^T$, $\bS = [\bS_1, \cdots, \bS_P]$, where node $k$ holds $\bOmega_k$ and $\bS_k$
        \For{$k=1,\cdots,P$ simultaneously}
            \For{$j$ in $1\cdots,P$}
                \State \texttt{send} $\bOmega_{k}$ to node $k - j$ and \texttt{recv} $\bOmega_{k + j}$ from node $k + j$
                \State Compute $\bOmega_{k+j}\bS_{k}$
            \EndFor{}
            \State $\bG_k \gets [(\bOmega_1\bS_k)^T, \cdots, (\bOmega_P\bS_k)^T]^T$
        \EndFor{}
    \State \textbf{Output}: $\bG = [\bG_1, \cdots, \bG_P]$, where node $k$ holds $\bG_k$
    \end{algorithmic}
\end{algorithm}

For its simplicity and linear convergence rate, the ACCORD-FBS algorithm (\Cref{alg:pgd}) has an advantage in scaling up to handle massive-scale data.
Note that the main computational components of ACCORD-FBS are: 1) sparse-dense matrix multiplication in computing the gradient $\nabla g(\bOmega^{(t)}) = \bOmega^{(t)} \bS$ and 2) element-wise operations in computing  \eqref{eqn:proxgrad}; note that $\bOmega^{(t)}$ is sparse by construction. 
The latter is ``embarassingly parallel.''
The former can also be easily parallelized in shared-memory systems, such as those employing graphical processing units (GPUs). However, if the size of the data becomes massive so that it does not fit into the system memory, 
employment of a distributed memory system becomes necessary.
In distributed computation, communication cost becomes a significant factor of the performance.

Our distributed-memory HPC implementation of ACCORD-FBS, termed HP-ACCORD,
iteratively conducts the sparse-dense matrix multiplication for the gradient step 
as a special case of the SpDM$^3$ algorithm \citep{koanantakool2018communication}. 
This algorithm, named one-dimensionally distributed matrix multiplication (1DMM), is summarized in \Cref{alg:1dmm}.
In HP-ACCORD, $\bOmega$ and $\bS$ are separated by columns (row-wise separation is also possible for $\bOmega$) and store each partition in different computational nodes. Then, the gradient $\nabla g(\bOmega)$ is computed with 1DMM.
While the standard method for distributed matrix-matrix multiplication on HPC systems is the scalable universal matrix multiplication algorithm \citep[SUMMA,][]{vandegeijn1997SUMMA}, this method partitions involved matrices by both rows and columns and allocates them across computational nodes, and completing each block of the product requires multiple rounds of broadcasting submatrices of both operands, which results in a significant performance bottleneck.
On the other hand, 1DMM sends and receives one-dimensional blocks of only one operand, i.e., $\bOmega$ in \Cref{alg:1dmm}. Because of this difference, the communication cost of \Cref{alg:1dmm} is smaller than SUMMA since $\bOmega$ is sparse while $\bS$ is dense \citep{koanantakool2016communication}.
Computation of the gradient can alternatively be conducted in two steps: $\bY = \bOmega\bX^T$ and $\nabla g(\bOmega) = (1/n)\bY\bX$, each of which can be computed with 1DMM.
This strategy is advantageous when $n$ is much smaller than $p$ \citep{koanantakool2018communication}.

\subsection{Tuning}

The choice of the regularization parameter $\lambda$ impacts the practical performance of the ACCORD estimator. 
At the omics scale, sample reuse methods such as cross-validation or neighborhood selection \citep{meinshausen2010stability} are ruled out, as they incur a nontrivial number of expensive passes to compute the estimator.
Following CONCORD and SPACE, which are also pseudolikelihood-based methods, we adopt a Bayesian information criterion (BIC)-type approach for tuning.
Specifically, using the loss function part of \eqref{eqn:accord}, we choose $\lambda$ minimizing 
\begin{equation}\label{eqn:epBIC}
    (2n)\left\{
    -\log\det\hat{\bOmega}_D
    + (1/2)\text{tr}(\hat{\bOmega}^T\hat{\bOmega}\bS) \right\}
    + \|\hat{\bOmega}\|_0  \log n
    + 4\gamma\|\hat{\bOmega}\|_0 \log p
    ,
\end{equation}
where $\|\hat{\bOmega}\|_0$ is the number of nonzero off-diagonal elements of the estimate $\hat{\bOmega}=\hat{\bOmega}(\lambda)$. 
The last term is taken from the extended BIC for graphical lasso  \citep{foygel2010extended}, proposed to promote further sparsity;
$\gamma\in(0, 1]$ is a user-specified parameter. 
We may call quantity \eqref{eqn:epBIC} an extended pseudo-BIC (epBIC).

\subsection{Bias correction}

Correcting the biases introduced by the $\ell_1$ penalization by refitting is a common practice \citep{MEINSHAUSEN2007374,candes2007,cai2011constrained}. 
Following these approaches,
we propose the following second-stage refitting procedure for ACCORD.  
If we let the support of the ACCORD estimate $\hat{\bOmega}$ computed by solving \eqref{eqn:accord} for an appropriate $\lambda$ (e.g., epBIC of \eqref{eqn:epBIC}) be $\hat{S}_{\lambda}$, then we refit by computing
\begin{equation}\label{eqn:d-accord}
    \breve{\bOmega} 
    =
    \argmin_{\bOmega:\; \bOmega_{\hat{S}_{\lambda}^c}=0}\left\{
    -\log\det\bOmega_D
    + (1/2)\text{tr}(\bOmega^T\bOmega\bS) 
    + 
    \phi\lambda\|\bOmega\|_1
    \right\}
    ,
\end{equation}
where $0 \leq \phi \leq 1$.
The refitted estimator $\breve{\bOmega}=(\breve{\omega}_{ij})$ can be computed efficiently using \Cref{alg:pgd}, by replacing the $\lambda$ with $\phi\lambda$ for $(i,j) \in \hat{S}_{\lambda}$ and with $\infty$ for $(i,j) \not\in \hat{S}_{\lambda}$.

\section{Statistical Properties}\label{sec:meth}
In this section, we show that the ACCORD estimator can consistently estimate $\bOmega^\ast$, a one-to-one reparameterization of the true precision matrix $\bTheta^*$ in various measures, under appropriate conditions. All the results provided here are non-asymptotic.  
In addition to the vector $\ell_1$ and $\ell_{\infty}$ norms introduced in \S\ref{sec:intro}, 
we use $\vvvert \bM \vvvert$ to denote an  operator norm of matrix $\bM=(m_{ij})$ induced by the underlying vector norm. In particular, $\vvvert \bM \vvvert_{\infty}= \max_{\bx \neq 0} \frac{\|\bM\bx\|_{\infty}}{\|\bx\|_{\infty}} = \max_i \sum_{j=1}^p |m_{ij}|$. The Frobenius norm $\|\bM\|_F$ of $\bM$ is its vector $\ell_2$ norm.
For a finite set $A$, we denote by $|A|$ the number of elements in $A$.
For an $l \times m$ matrix $\bM$, $A\subset[l]:=\{1,\dotsc,l\}$, and $B\subset[m]$, we denote by $\bM_{AB}$ the $|A|\times|B|$ submatrix of $\bM$ taking the rows and columns of $\bM$ with indices in $A$ and $B$, respectively. 

\subsection{Estimation error bounds}
We first provide finite-sample estimation error bounds in vector $\ell_1$ and $\ell_2$ norms.
Recall that a zero-mean random vector $Z$ is sub-Gaussian with parameter $\sigma$ if $\mathbb{E}[\exp(tZ)] \leq \exp({\sigma^2 t^2}/{2})$ for all $t \in \mathbb{R}$.
Let $\kappa_{\Omega^\ast} = \vvvert \bOmega^\ast \vvvert_{\infty}$.
\begin{theorem}\label{thm:l2consistency}
    Suppose the data matrix $\bX \in \mathbb{R}^{n \times p}$ is composed of $n$ i.i.d. copies of zero-mean continuous random vector $X=(X_1,\dotsc, X_p)\in\mathbb{R}^p$ with covariance matrix $\bSigma^*= (\Sigma^*_{ij}) \allowbreak = {\bTheta^*}^{-1}$ and each $X_j/\sqrt{\Sigma^*_{jj}}$ being sub-Gaussian with parameter $\sigma$.
    Also suppose that there exists $\alpha, \beta, \eta > 0$ such that 
    $\E_{X}|\langle X, y\rangle|^{2} \geq \alpha$ and
    $\E_{X}|\langle X, y\rangle|^{2+\eta} \leq \beta^{2+\eta}$ for any $y \in \mathbb{R}^p$ with $\Vert y \Vert _2 = 1$.
    If we let $S = \{(i,j)\in[p]\times[p]: \theta^*_{ij} \neq 0\}$ be the support of $\bTheta^*$ (hence of $\bOmega^*)$,
    then there exist positive constants $\kappa,\allowbreak c_0, \allowbreak c_1,$ and $c_2$ that explicitly depends on
    $\alpha, \beta, \eta, \sigma,$ and $\max_{i\in[p]}\Sigma_{ii}^*$ such that
    $\lambda \!= 64(1+4\sigma^2)\kappa_{\Omega^\ast}(\max_{i\in[p]}\Sigma_{ii}^*)\sqrt{n^{-1}\log p}$
    in \eqref{eqn:accord}  yields
    $$
        \Vert \hat{\bOmega} - \bOmega^\ast \Vert_1 \leq 16\kappa^{-1}\lambda|S|
        \text{~and~~}
        \Vert \hat{\bOmega} - \bOmega^\ast \Vert_F \leq 4\kappa^{-1}\lambda\sqrt{|S|}
    $$
    with a probability at least
    $1 - c_1e^{-c_2n} - 4p^{-2}$,
    provided that
    $n > \max\{(64c_0^2/\kappa^2)|S|\log p, \allowbreak (1/16)\log p\}$.
\end{theorem}
\begin{remark} 
    The consistency of $\hat{\bOmega}$ in terms of $\|\cdot\|_1$ and $\|\cdot\|_F$ is obtained if $\sigma^2$ and $\kappa_{\Omega^\ast}$ are bounded, at the rate of $O_P(\sqrt{n^{-1}|S|\log p})$. 
    The latter can be bounded, e.g., when the graph implied by $\bTheta^*$ has 
    a bounded number of non-zero entries per row
    $d=\max_{i\in [p]}|\{j \in [p]:\allowbreak \theta^\ast_{ij} \neq 0 \}|$.
    Then $|\theta^*_{ij}|\leq\sqrt{\theta^*_{ii}\theta^*_{jj}}$
    and $\kappa_{\Omega^\ast}=\max_{i\in[p]}\sum_{j=1}^p|\theta^*_{ij}|/\sqrt{\theta^*_{ii}} \leq d \max_{i\in[p]}\sqrt{\theta^*_{ii}} \leq d / \sqrt{\lambda_{\min}(\bSigma^*)}$,
    where $\lambda_{\min}(\bSigma^*) \geq \alpha$ is the minimum eigenvalue of $\bSigma^*$.
    This rate and the sample complexity of $n \gtrsim |S|\log p$ match those for the graphical lasso \citep[Proposition 11.9]{wainwright2019high}.
\end{remark}

\subsection{Edge selection and sign consistency}
With further assumptions, the element-wise ($\ell_{\infty}$) error bound of $\hat{\bOmega}$ can be controlled, with which edge selection and sign consistency can be proved. 
The Hessian matrix of the population risk $R(\bOmega)$ (see \S\ref{sec:alg}) with respect to the usual vectorization of $\bOmega$ at $\bOmega^*$ is
\begin{equation}\label{eqn:grad}
    \bGamma^{\ast} 
    = \bSigma^* \otimes \bI_p + (\bOmega_D^{\ast -1}\otimes \bOmega_D^{\ast -1})\bUpsilon
    .
\end{equation}
where $\otimes$ is a Kronecker product and $\bUpsilon = \sum_{i=1}^p e_i e_i^T\otimes e_i e_i^T$.
Let $S$ be the support of $\bTheta^*$ as stated in \Cref{thm:l2consistency}. 
We assume the following property of the data distribution.

\begin{assumption}[Irrepresentability condition]\label{assm:irrepresentability}
    There exists $\alpha \in [0, 1)$ that satisfies 
    \begin{equation}\label{eqn:irrepresentability}
    \vvvert\bGamma^{\ast}_{S^cS}\bGamma^{\ast -1}_{SS}\vvvert_{\infty} 
    \leq 
    1 - \alpha
    .
    \end{equation}
\end{assumption}
Let us define the following associated quantities:
    $\kappa_{\Gamma^\ast} = \vvvert (\bGamma^\ast_{SS})^{-1} \vvvert_{\infty}$,
    $\gamma_1 = \vvvert \bOmega^{\ast -1}_D \vvvert_{\infty}$%
    .
Also, recall that $\kappa_{\Omega^\ast} = \vvvert \bOmega^\ast \vvvert_{\infty}$.
These quantities are defined to quantitatively measure and track the model complexity and are allowed to grow along with $(n,p,d)$,
where $d$ denotes the maximum number of non-zero entries per row in $\bTheta^*$.
\begin{theorem}\label{cor:exptail}
Suppose the data matrix $\bX \in \mathbb{R}^{n \times p}$ is composed of $n$ i.i.d. copies of zero-mean continuous random vector $X=(X_1,\dotsc, X_p)\in\mathbb{R}^p$ with covariance matrix $\bSigma^*= (\Sigma^*_{ij}) \allowbreak = {\bTheta^*}^{-1}$ and each $X_j/\sqrt{\Sigma^*_{jj}}$ being sub-Gaussian with parameter $\sigma$.
If further \Cref{assm:irrepresentability} holds, then 
for 
$\lambda = 80\sqrt{2}(1+4\sigma^2)(\max_{i\in[p]}\Sigma_{ii}^*)\kappa_{\Omega^\ast}\alpha^{-1}\sqrt{(\tau\log p + \log 4)/n}$, 
$\tau > 2$,
\begin{enumerate}[label=(\alph*)]
    \item there holds
    $\|\hat{\bOmega} - \bOmega^\ast\|_{\infty}%
			\leq%
    24\sqrt{2}(1+4\sigma^2)(\max_{i\in[p]}\Sigma_{ii}^*)\kappa_{\Gamma^\ast}\kappa_{\Omega^\ast}(1 + 10/\alpha) \sqrt{(\tau\log{p}+\log{4})/n}%
    $\\
    and
    $\|\hat{\bTheta} - \bTheta^\ast\|_{\infty}%
			\leq%
    (7/3)\kappa_{\Omega^*}\|\hat{\bOmega} - \bOmega^\ast\|_{\infty}
    $,
    where
    $\hat{\bTheta} = \hat{\bOmega}_D\hat{\bOmega}$
    ;
    \item the estimated support  $\hat{S}=\{(i,j)\in [p]\times [p]: \hat{\omega}_{ij} \neq 0\}$ is contained in the true support $S$ and includes all edges $(i,j)$ with 
		$|\omega^\ast_{ij}| > %
			24\sqrt{2}(1+4\sigma^2)(\max_{i\in[p]}\Sigma_{ii}^*)\kappa_{\Gamma^\ast}\kappa_{\Omega^\ast}(1 + 10/\alpha) \sqrt{(\tau\log{p}+\log{4})/n}$%
		,
\end{enumerate}
with a probability no smaller than $1 - p^{-(\tau-2)}$,
provided 
$n > 128(1+\sigma^2)^2(\max_{i\in[p]} \Sigma_{ii}^*)^2 \delta^{-2} (\allowbreak\tau\log{p} + \log{4})$ where
$$
    \delta = \min\left\{ \frac{\min\left\{\frac{1}{3\gamma_1}, \frac{1}{3\gamma_1^3\kappa_{\Gamma^\ast}}, \frac{\kappa_{\Omega^\ast}}{3d}\right\}}{3\kappa_{\Gamma^\ast}\kappa_{\Omega^*}\left(1 + {10}/{\alpha}\right)}
        , 
        \frac{2}{%
            27\gamma_1^3 
            \kappa_{\Gamma^\ast}^2\kappa_{\Omega^\ast}\left(1 + {10}/{\alpha}\right)^2%
        }
        ,
        8(1+4\sigma^2)(\max_{i\in[p]}\Sigma_{ii}^*)
    \right\} 
    .
$$
\end{theorem}
\begin{remark}
    With other quantities held fixed, the sample size required to achieve the rate 
    $\Vert \hat{\bTheta} - \bTheta^*\Vert_\infty = O_P(\sqrt{(\tau \log p)/n})$
    is $n \gtrsim d^2 \tau \log p$. 
    It can be also shown that for data distributions with a bounded $4m$-th moment, the error rate is $O_P(\sqrt{p^{\tau/m}/n})$ for $n \gtrsim d^2 p^{\tau/m}$; see 
    Supplementary Material A. 6.
    These sample complexities and rates of convergence match those of graphical lasso, obtained by \citet{ravikumar2011high} under 
    similar conditions.
\end{remark}

Part (b) of \Cref{cor:exptail} only states that the ACCORD estimator can exclude all false edges and find true edges with large enough $\omega_{ij}^*$'s. Exploiting it further, the sign consistency on \textit{all} edges can be declared. 
For $\hat{S}=S$,
let $\omega_{\min} = \min_{(i,j) \in S}|\omega_{ij}^*|$:
\begin{theorem}\label{thm:sign_consistency}
Assume the same conditions as \Cref{cor:exptail}. If the sample size satisfies
$$
    n > 128(1+4\sigma^2)^2(\max_{i\in[p]}\Sigma_{ii}^*)^2(\tau\log p + \log 4)/\min\{\omega_{\min}/[6\kappa_{\Gamma^\ast}\kappa_{\Omega^\ast}(1+10/\alpha)],\delta\}^2
$$
where $\sigma$ and $\delta$ are as defined in \Cref{cor:exptail},
then the perfect sign recovery event
    $\{\operatorname{sign}(\omega_{ij}^*) = \operatorname{sign}(\hat{\omega}_{ij})$ for all $(i,j)\}$
occurs with a probability no smaller than $1-p^{-(\tau - 2)}$.
\end{theorem}
It follows that the bias-corrected estimator in \eqref{eqn:d-accord} 
is also consistent: 
\begin{corollary}\label{cor:debiased_error}
    Assume the same conditions as \Cref{thm:sign_consistency}. Then,
    for the bias corrected estimator $\breve{\bOmega}$ in \eqref{eqn:d-accord},
    it holds that $\{\operatorname{sign}(\omega_{ij}^*) = \operatorname{sign}(\breve{\omega}_{ij}) \text{ for all } (i,j)\}$ and
    $$
        \Vert \breve{\bOmega} - \bOmega^*\Vert_\infty 
        \leq 
        24\sqrt{2}(1+4\sigma^2)(\max_{i\in[p]}\Sigma_{ii}^*)\kappa_{\Omega^\ast}(1 + 10\alpha^{-1})\sqrt{(\tau\log p + \log 4)/n}
    $$
    with a probability no smaller than $1-p^{-(\tau - 2)}$
\end{corollary}

\section{Numerical Experiments}\label{sec:exp}
\subsection{Linear convergence of ACCORD-FBS}\label{sec:exp:convergence}

We 
provide empirical evidences of the guaranteed descent property and linear convergence exhibited by  ACCORD-FBS (\Cref{alg:pgd}). 
The merit of the novel operator splitting utilized by the latter is demonstrated by comparing it with 
ACCORD-ISTA (see the last paragraph of \S\ref{sec:alg:accord-fbs}).
Firstly, we generated two Erdos-Renyi graphs of size $p\!=\! 1000$ 
with a sparsity level of $15\%$. 
To construct a ground truth precision matrix, we employed the following procedure. The edge weights were selected from a uniform distribution on $[0.5, 1]$ and their signs were flipped with a probability of $0.5$. To ensure symmetry and positive definiteness, the resulting matrix was added to its transpose and its diagonal entries are set to be 1.5 times the 
absolute sum 
of the off-diagonal entries of the corresponding rows. This matrix was scaled by pre- and post-multiplying a diagonal matrix such that all the diagonal entries are equal to one. Finally, variation among the diagonal entries was introduced by performing another pre- and post-multiplications with a diagonal matrix 
with uniformly distributed entries
on $[1, \sqrt{3}]$. 
Using this precision matrix, multivariate Gaussian data with a sample size of $n = 500$ were generated.

ACCORD-FBS was investigated with two variants: one with a constant step size of $\tau = {1}/{L}$ and the other employing backtracking line search with $\tau = {1}/{L}$ as a lower bound. In contrast, a grid of constant step sizes, $\tau \in \{ 0.25, 0.6\}$, along with backtracking, were chosen for ACCORD-ISTA to demonstrate its convergence behavior.

\begin{figure}[t]
\centering

\includegraphics[width=0.99\textwidth]{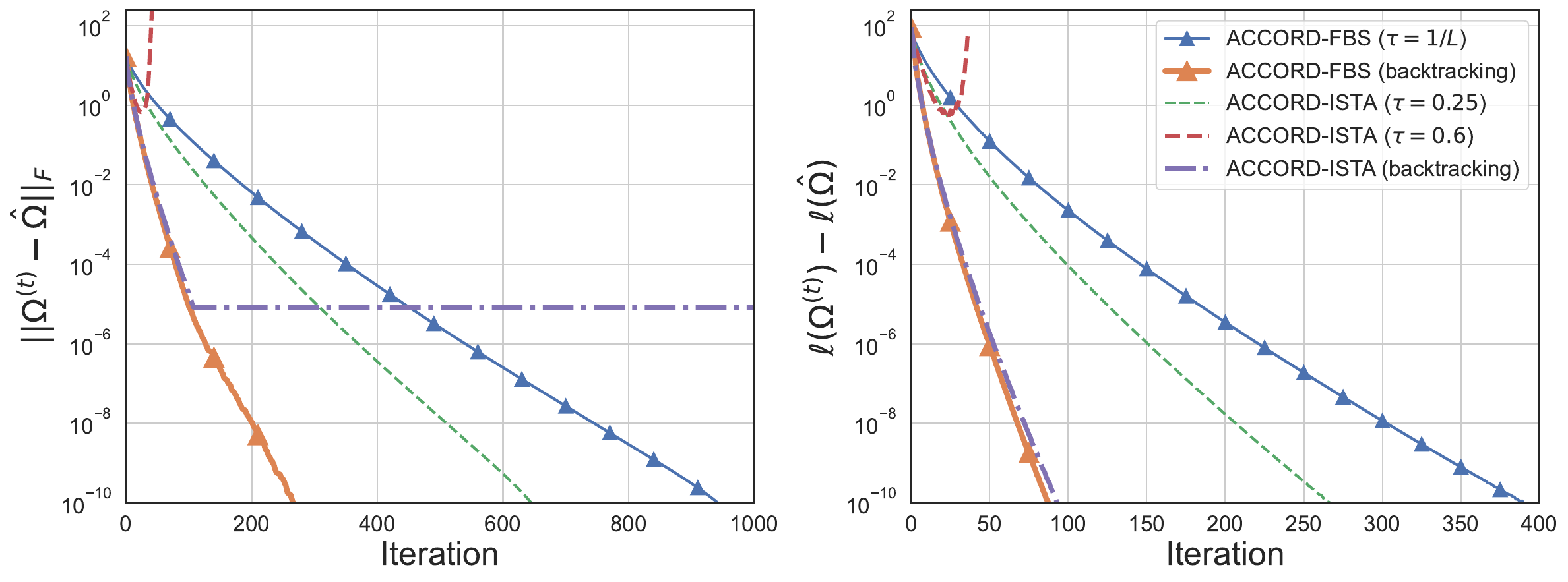}%
\caption{Convergence comparison between ACCORD-FBS and ACCORD-ISTA.}
\label{fig:linear-conv}
\end{figure}

The convergence behavior of the iterate $\{\bOmega^{(t)}\}$ and the objective value $\{\ell(\bOmega^{(t)})\}$ 
is illustrated in \Cref{fig:linear-conv}.
Here, $\hat{\bOmega}$ denotes the final iterate obtained by executing ACCORD-FBS until termination, with the criterion of $\Vert \bOmega^{(t+1)} - \bOmega^{(t)}\Vert < 10^{-15}$. 
In \Cref{fig:linear-conv}, ACCORD-FBS exhibits linear convergence for both variants, while backtracking shows a faster rate. On the other hand, the convergence behavior of ACCORD-ISTA varies significantly across different constant step sizes, ranging from slow convergence to divergence. Moreover, the iterates from ACCORD-ISTA with backtracking  encounters a plateau at an early stage. These observations underscore the challenges associated with the selection of an appropriate step size in ACCORD-ISTA. 

\subsection{Scalability of HP-ACCORD}\label{sec:hp-scalability}
We next investigate the scalability of HP-ACCORD (\S\ref{sec:alg:hp-accord}) using simulated data of dimension $p$ up to one million.
At this scale, even simulating multivariate Gaussian samples becomes a nontrivial task \citep{vono2022high}. We took a similar approach to the numerical experiments in \citet{li2010inexact}: generate a sparse $p\times p$ unit lower triangular matrix $\bL$ taking values in $[-1,1]$ and compute $\by = \bL^{-T}\bx$, $\bx \sim  N(\mathbf{0},\bI_p)$, by backsubstitution so that $\by$ follows $N(0, (\bL\bL^T)^{-1})$. The location of off-diagonal non-zero entries of $\bL$ were uniformly selected so that the graph implied by the precision matrix $\bL\bL^T$ had an average degree of $10.3$ for each $p$, and the maximum degree ranged from 42 to 66.

We employed the Nurion supercomputer at the Korea Institute of Science and Technology Information (KISTI) National Supercomputing Center (KSC) for the scalability experiment. Nurion
is a Cray CS500 system with $8,305$ Intel Xeon Phi 7250 1.4GHz manycore (KNL) computational nodes with 96GB of memory per node, featuring 
25.3 petaflops of peak performance.
As a benchmark, we compared HP-ACCORD with BigQUIC.\footnote{Available at \url{https://bigdata.oden.utexas.edu/software/1035/}.} and \texttt{fastclime}.\footnote{Available at \url{https://cran.r-project.org/src/contrib/Archive/fastclime/}.} 
Note that \texttt{fastclime} is an R package with a single-core-oriented implementation of CLIME 
in C at its heart.
%
Both BigQUIC and HP-ACCORD perform multi-core computations written in C++. However, BigQUIC's scalability is limited to a single node as a shared-memory algorithm. In contrast, HP-ACCORD can run on multiple nodes simultaneously utilizing distributed memory in supercomputing environments (\S\ref{sec:alg:hp-accord}).

\begin{figure}[t]
    \centering
    \begin{subfigure}[t]{0.48\textwidth}
        \centering
        \includegraphics[width=\linewidth]{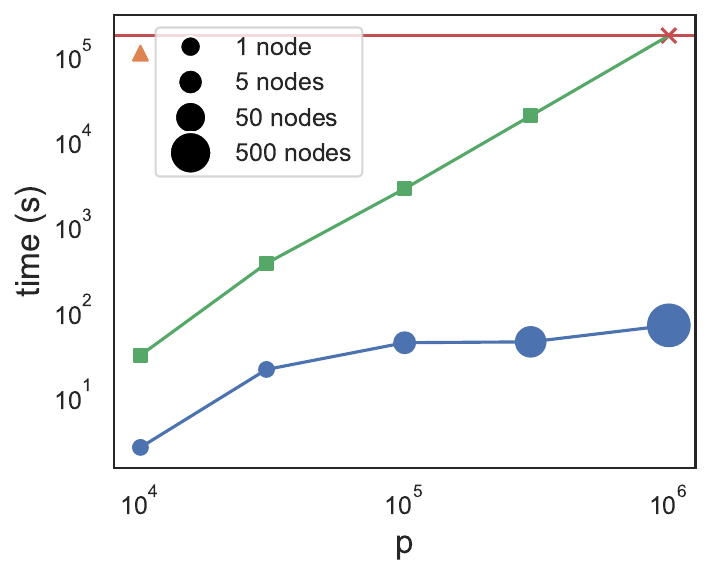}
        \caption{Large $\lambda$ (diagonal estimates)}
    \end{subfigure}
    \quad
    \begin{subfigure}[t]{0.48\textwidth}
        \centering
        \includegraphics[width=\linewidth]{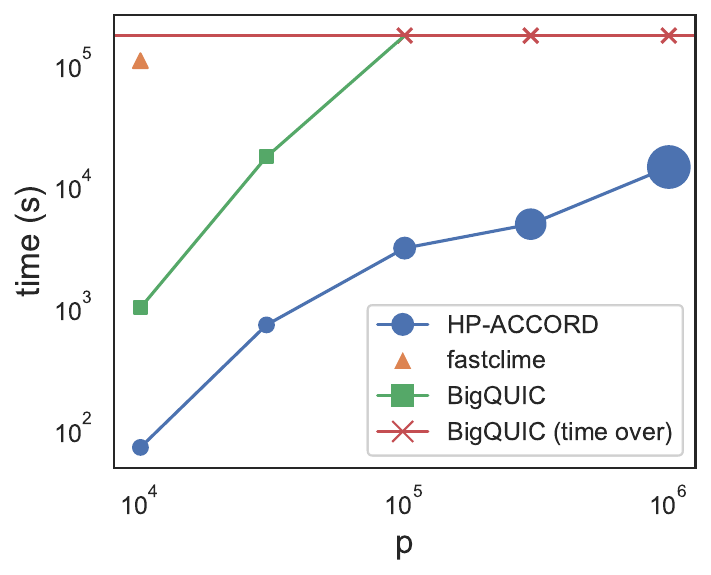}
        \caption{Moderate $\lambda$ (spase estimates)}
    \end{subfigure}
    \caption{Scalability comparison between  HP-ACCORD and other methods. The horizontal red lines indicate the 48-hour limit of the 
    supercomputer
    system.}
    \label{fig:scc}
\end{figure}

We report the results in \Cref{fig:scc}.
Since the memory of a single KNL node (96GB) was not enough for the computation with data dimensions greater than or equal to $100,000$, an appropriate number of nodes that can handle the input size data was used for HP-ACCORD, showcasing the scalability of the algorithm. 
The supercomputing center imposed a 48-hour restriction on the running time of a single job. Thus we only report accurate timing for the processes finished within this limit.
In panel (a), the regularization parameter $\lambda$ was set large enough so that the resulting precision matrix estimate $\hat{\bTheta}$ becomes diagonal. In panel (b), $\lambda$ was adjusted so that the numbers of nonzero entries of $\hat{\bTheta}$ have a similar scale to their true precision matrices. Note that \texttt{fastclime} had to use the time near the limit to estimate a precision matrix to compute estimate for two $\lambda$'s for $p=10,000$, and it could not run on data with $p=30,000$ or higher due to time and memory limitations. 
Also, for the both panel, we could observe that HP-ACCORD outperforms BigQUIC in terms of computation time, even when only a single node is employed. 
By adopting multiple computational nodes, HP-ACCORD was able to deliver the estimate within the time budget in all scenarios. On the contrary, BigQUIC failed to finish the process on time when the dimension exceeded 100,000 and the $\lambda$ was chosen to yield a non-diagonal estimator. 
Internally, 
BigQUIC tries to find a permuted partition $\{B_1, \cdots, B_k\}$ of the $p$ coordinates using a graph clustering algorithm \citep{karypis1998fast, dhillon2007weighted} so that most of the coordinate updates are performed in the diagonal blocks $(B_1, B_1), \cdots, (B_k, B_k)$. However, the size of each partition is limited due to the memory limit of a single computational node (about 20,000 in the Nurion environment in which each node has 96GB of memory), and clustering with such a limited block size bares many off-block diagonal edges in our simulated data. In consequence, the conjugate gradient method employed to compute the off-diagonal blocks are hardly skipped, causing the algorithm to run extremely slowly. %
Even when $\lambda$ is so large that the estimate should be diagonal, BigQUIC failed to complete the computation if the dimension reached one million.

\subsection{Edge detection and estimation error}
We examine the edge detection and parameter estimation performance of ACCORD on simulated data, comparing it with CONCORD, graphical lasso (GLASSO), SPACE, CLIME.
Following the hub graph setting in \citet{peng2009partial}, 
we generated a graph with 10 clusters each having 100 nodes and 90 edges and connected 100 randomly chosen node pairs from adjacent clusters.
The clusters had either a hub network or scale-free graph structure. 
A hub network was generated by
(1) constructing an Erd\"os-R\'enyi graph with 97 nodes and 45 edges, and (2) consecutively adding 3 hub nodes, each connected to random 15 nodes.
A scale-free graph was generated so that the degree $k$ of the nodes has a probability $P(k) \sim k^{-2.3}$.
We additionally considered a simple Erd\"os-R\'enyi graph with 1,000 nodes and 1,000 edges without the clustered structure.
The partial correlation matrix corresponding to the graph was chosen so that the matrix has minimum eigenvalue of at least 0.2, and each nonzero entries have absolute value of 0.1 at least.
Model selection was carried out using epBIC \eqref{eqn:epBIC} for ACCORD and CONCORD
for an evenly spaced grid of $\lambda$'s in logarithmic scale, while extended Gaussian BIC was used for GLASSO and CLIME. 
Although cross validation was used to select the $\lambda$ in the original paper of CLIME \citep{cai2011constrained}, we used BIC instead to avoid the repetitive parameter fitting.
For SPACE, we used ``BIC-type criterion'' defined in \citet{peng2009partial}.
For CONCORD, the matrix $\hat{\bTheta}$ minimizing \eqref{eq:qcon} was treated as a precision matrix despite the possible inconsistency (see \S\ref{sec:challenges}).
This procedure was repeated 50 times. 
For ACCORD, we also considered the debiasing procedure \eqref{eqn:d-accord}.

\begin{table}[t]
\small
\centering
\footnotesize
\begin{tabular}{ccccc}
\toprule
Graph & Method & AUPRC & \# TP edges & \# FP edges 
\\
\midrule
\multirow{4}{*}{Hub Network} & ACCORD & \textbf{0.843} (0.011) & 732 (18.6) & 55 (17.0)   \\
& CONCORD & 0.837 (0.011) & 715  (18.1) & 50  (16.5) \\
& GLASSO & 0.835 (0.010) & 706  (20.5) & 63  (11.3) \\
& SPACE & 0.813 (0.011) & 768  (13.8)   & 321  (33.0) \\
& CLIME & 0.835 (0.010) & 734  (13.8) & 64   (17.5) \\
\midrule
\multirow{4}{*}{Scale-free} & ACCORD & \textbf{0.882} (0.008) &  810 (13.9)  &   77 (18.7)\\
& CONCORD & \textbf{0.882} (0.008) & 797 (17.4)  &   67 (16.2)\\
& GLASSO & 0.874 (0.008) &  808 (14.4)  &  110 (16.9)\\
& SPACE & 0.864 (0.008) & 831 (13.8)  &   321 (33.0)\\
& CLIME & 0.879 (0.009) & 813 (15.8)  &   85 (22.4)\\
\midrule
\multirow{4}{*}{Erd\"os-R\'enyi} & ACCORD & \textbf{0.884} (0.009) &  811 (14.5)  &   74 (16.9) \\
& CONCORD & \textbf{0.885} (0.008) & 802 (14.7)  &   70 (14.5) \\
& GLASSO & 0.874 (0.009) &  809 (16.0)  &  114 (12.7) \\
& SPACE & 0.868 (0.009) &   836 (10.9)  &  369 (28.3) \\
& CLIME & 0.882 (0.009) & 817 (13.5)  &   88 (19.3)\\
\bottomrule
\end{tabular}
\caption{Edge detection performance, mean (standard deviation) over 50 replications.}
\label{tab:estimated_edges}
\end{table}

\Cref{tab:estimated_edges} 
reports the area under the precision-recall curve (AUPRC) and number of edges selected.
In terms of AUPRC (and Matthews correlation coefficient; see Supplementary Material B), ACCORD performed slightly better than or similarly to other methods. 
The trend of the precision-recall curves did not particularly vary among replications.
In all cases, the penalty selected from each selection method yielded reasonable number of selected edges.
We could observe that BIC-type methods with the extended term drastically reduces false positive (FP) edges in the expense of few true positive (TP) edges.

\Cref{table:debiasing-table1} reports the total squared error of the estimated precision matrix $\bTheta$ and its reparameterization $\bOmega$ by ACCORD and CONCORD, along with their debiased refit. 
Compared to CONCORD, ACCORD estimates clearly showed better results for both $\hat{\bTheta}$ and $\hat{\bOmega}$ in terms of the estimation error. The debiased refit also improves the estimation performance.
For more details about the impact of the debiased refit, see Supplementary Material B.

\begin{table}[t]
\footnotesize
\centering
\begin{tabular*}{\textwidth}{@{\extracolsep{\fill}}ccccc}
\toprule
\multicolumn{2}{c}{Estimation} & Graph & Total Squared Error ($\hat{\bTheta}$) & Total Squared Error ($\hat{\bOmega}$)\\
\midrule
\multirow{6}{*}{ACCORD} 
&
\multirow{3}{*}{Biased} 
&       Hub Network & 194.6 (10.7) & 73.7 (3.3) \\
& &      Scale-free & 202.6 (8.5) & 68.1 (2.2) \\
& & Erd\"os-R\'enyi & 201.0 (8.1) & 69.7 (2.1) \\
\cmidrule[0.1\cmidrulewidth](lr){2-5}
&
\multirow{3}{*}{Debiased} 
&      Hub Network & \textbf{ 37.7} (1.4) &  \textbf{19.7} (0.6) \\
& &      Scale-free & \textbf{ 36.2} (2.2) &  \textbf{16.5} (1.0) \\
& & Erd\"os-R\'enyi & \textbf{ 35.4} (1.6) &  \textbf{16.3} (0.8) \\
\midrule
\multirow{6}{*}{CONCORD} 
&
\multirow{3}{*}{Biased} 
&       Hub Network & 392.3 (18.0) & 108.0 (3.7) \\
& &      Scale-free & 433.6 (21.5) & 107.9 (3.8) \\
& & Erd\"os-R\'enyi & 437.5 (19.7) & 110.8 (3.0) \\
\cmidrule[0.1\cmidrulewidth](lr){2-5}
&
\multirow{3}{*}{Debiased} 
&       Hub Network & 206.6 (11.0) & 44.2 (1.4)\\
& &      Scale-free & 226.6 (13.9) & 43.6 (1.8)\\
& & Erd\"os-R\'enyi & 231.0 (13.5) & 44.8 (1.8)\\
\bottomrule
\end{tabular*}
\caption{MSE of biased and debiased ACCORD and CONCORD in partial correlation, mean (standard deviation) over 50 replications.
}
\label{table:debiasing-table1}
\end{table}

\begin{figure}[t]
    \centering
    \begin{subfigure}[b]{0.48\textwidth}
        \includegraphics[width=\linewidth]{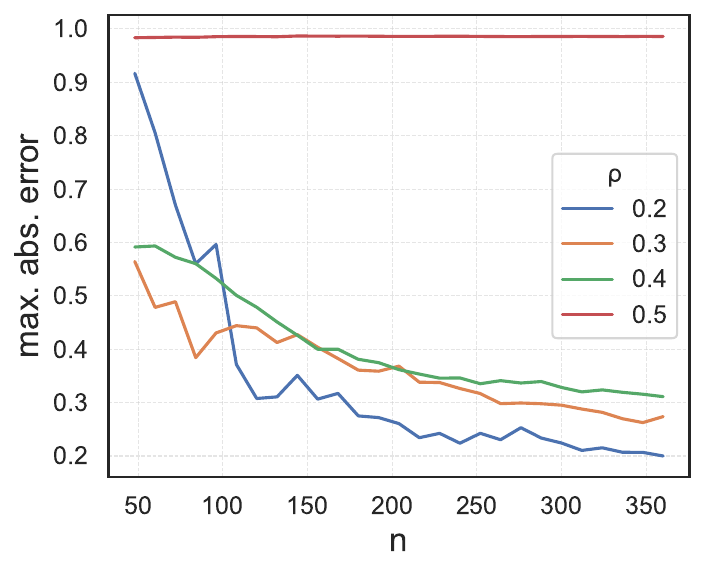}
        \caption{Chain graph}
        \label{fig:cg}
    \end{subfigure}
    \hfill
    \begin{subfigure}[b]{0.48\textwidth}
        \includegraphics[width=\linewidth]{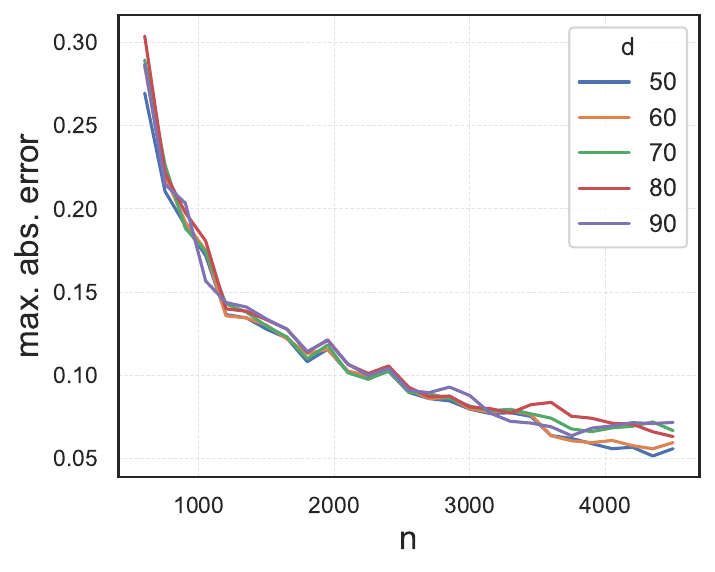}
        \caption{Star graph}
        \label{fig:sg}
    \end{subfigure}
    \caption{Maximum error of the ACCORD estimator by sample size $n$}
    \label{fig:simulation}
\end{figure}

\subsection{Estimation error dependency on precision matrix}
We further provide numerical results that illustrate how the structures and values of the partial correlation matrix can affect the estimation error of ACCORD. 
Motivated by experiments in \citet{ravikumar2011high}, we conducted simulation studies under two types of $\bTheta^*$: chain and star graphs. In both types, we let $\theta_{ii} = 1,$ implying that $\bOmega^* = \bTheta^*$.
In the chain graph setting, we fix $p = 120$ and let $\theta_{ij} = \rho > 0$ for $|i-j| = 1$ and $\theta_{ij} = 0$ for $|i-j| \geq 2.$ Thus, the maximum number of non-zero entries per row $d$ is fixed to $2$. With $\rho$ increasing, the $\kappa_{\Omega^*}$ and $\kappa_{\Gamma^*}$ also increase and the maximum irrepresentability constant $\alpha$ in \Cref{assm:irrepresentability} decreases. Note that for $\rho \geq 0.5$, \Cref{assm:irrepresentability} holds for any $\alpha \in [0,1)$.
In the star graph setting, $d-1$ nodes among $p = 200$ nodes are connected only to a single hub node, and there are no other connections. For all the connected edges, partial correlation entries are set as $2.5/(d-1)$ so that $\kappa_{\Omega^*}$ remains invariant with $d$. For sufficiently large $d$, both $\kappa_{\Gamma^*}$ and $\alpha$ remain nearly constant.
In both settings, an ACCORD estimate is computed with $n$ Gaussian samples drawn from the given $\bTheta^*$, and the penalty coefficient $\lambda$ is set to be proportional to $\sqrt{1/n}$ as suggested by the theory. 

In \Cref{fig:simulation} we provide the maximum estimation error observed with various sample sizes $n$ for the two types of graphs.
In the chain graph setting (panel (a)), the rates of the estimation error were similar 
for $\rho \leq 0.4$, while its magnitude increased with $\rho$. For $\rho = 0.5$, the estimator could not recover the chain graph structure, and increasing the sample size did not further improve the estimation error, indicating the necessity of the irrepresentability condition.
In the star graph setting (panel (b)), changing $d$ did not significantly affect the estimation error. 
This observation is consistent with \Cref{cor:exptail}.

\section{Case Study: Integrative Analysis of Multi-Omics data in Liver Cancer} \label{sec:HCC} 

\subsection{HP-ACCORD to delineate complex mechanisms in gene expression regulation}

In this section, we show that HP-ACCORD enables sophisticated biological inference using ultrahigh-dimensional multi-omic data through the example of the LIver Hepatocellular Carcinoma (LIHC) cohort ($n$=365) of TCGA. We used the LIHC dataset consisting of the expression data of 15,598 protein-coding genes (mRNAs), the expression data of 364 microRNAs (miRNAs) and DNA methylation levels ($\beta$-values) at 269,396 CpG islands located within the upstream regulatory regions of genes. The methylation data contained a subset of variables provided in the original HumanMethylation450k BeadChip array data from the TCGA data portal \citep{ally2017comprehensive}, where we included the probes in genomic regulatory regions or gene bodies only.

In this analysis, the main goal is to identify gene modules whose expression levels are co-regulated by common transcription regulators such as transcription factors (TF) and other co-activators. The graph identified in this analysis can, in turn, facilitate the prioritization of potential expression regulators for downstream biological inference in the context of liver cancer. Epigenetic modulation is an important confounder for this type of analysis. For instance, two genes targeted by a common TF may not show consistent changes in mRNA expression if their epigenetic states were not equally favorable for active transcription. Although DNA methylation does not capture all aspects of the complex epigenetic mechanisms, it is one of the main contributors to this process. Here we aim to estimate partial correlation network from an integrated data set consisting of DNA methylation and mRNA expression. By considering DNA methylation profiles in the same analysis, it is possible to identify gene-to-gene co-regulation network at the mRNA level independent of DNA methylation-driven effects.  

Since altered DNA methylation levels tend to be correlated locally in genomic neighborhoods and the probes for DNA methylation represent most variables in the data, the underlying graph is expected to reflect a sparse precision matrix with a large number of small block diagonals when the variables were ordered by genomic coordinates in each chromosome. The most causally implicated correlations between methylation probes and mRNAs are also expected to be from genomic neighborhoods with the exception of DNA methyltransferase genes and others regulating methylation and demethylation processes. By contrast, two genes showing high correlation can be located distantly, or even in different chromosomes. However, these latter correlations often form block diagonal patterns and we expect the target precision matrix to remain within a structure amenable to consistent estimation. 

We performed this analysis on two HPC systems: 
Nurion at KSC (see \S\ref{sec:hp-scalability}) and the Cori system at the National Energy Research Scientific Computing Center (NERSC), which is a Cray XC40 system with more than 2,000 computational nodes that are equipped with dual-socket 16-core Intel Xeon Processor E5-2698 (Haswell) and 128GB of memory for each node.
BigQUIC was not able to complete the computation of the precision matrix for a wide range of the regularization parameter $\lambda$, except for those that yielded diagonal matrices; this observation is consistent with the results from the numerical studies in \S\ref{sec:hp-scalability}.
As our analysis uses partial correlation as the metric for the relative contribution of epigenetic factors onto gene expression, we decided to forgo CONCORD in this analysis, given the ambiguity in the relationship between its estimate and the precision matrix 
as discussed in \S\ref{sec:challenges}.

In the following sections, we describe the partial correlation patterns within and across the two omics data. Using the precision matrix estimate and partial correlations, we aim to tease apart the impact of active TF-driven co-regulation of target genes from that of DNA methylation-mediated gene expression repression in the promoter regions. To the best of our knowledge, the effort to deconvolute the effects of two or more types of gene expression regulation using graphical models has not been attempted due to the computational bottleneck we address in this work. There is a caveat that the effects of other epigenetic regulators such as histone modifications and chromatin states are not accounted for in this analysis. Nonetheless, the analysis clearly shows the advantage of estimation in the ultrahigh-dimensional space when we compare the mRNA-mRNA co-regulation network reported from the two analyses in terms of the percentage of gene pairs with at least one or more shared TFs.  

\subsection{Network structure of epigenomic and transcriptomic data}

\begin{figure}[t]%
    \centering
    \subfloat{{\includegraphics[width=1\textwidth]{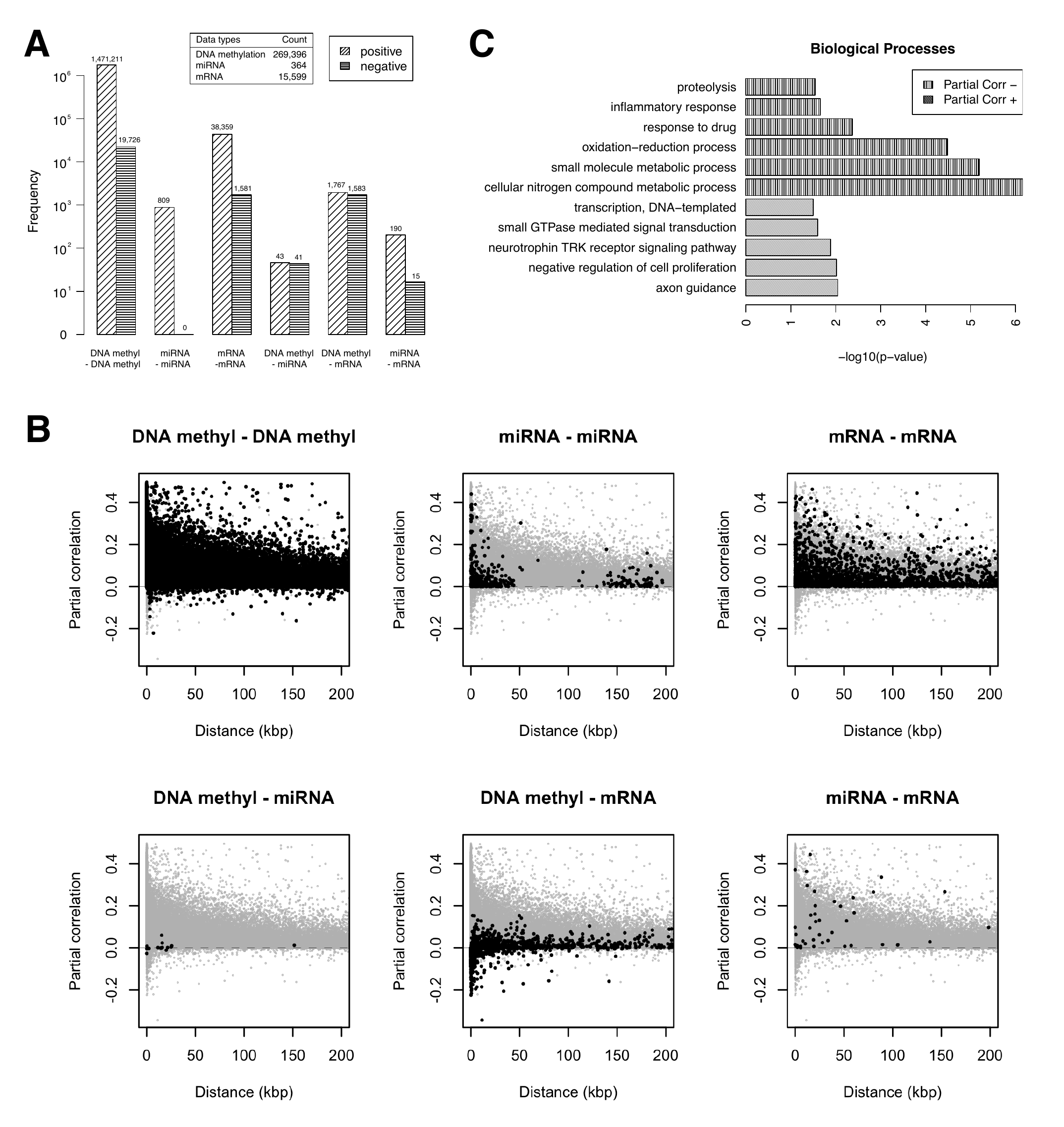}}}%
    \caption{Summary of intra-modality and inter-modality correlations. (A) Barplots show the total number of edges with non-zero correlations in each of the six edge categories. (B) Estimated partial correlations against genomic distances between connected nodes. Gray dots represent all edges and they are shown in panels. Black dots show the data pertaining to the edge category in each panel. (C) Gene ontology terms enriched in the nodes (genes) with positive and negative methylation-mRNA partial correlations. }%
    \label{fig:GD}%
\end{figure}

We estimated partial correlations by solving \eqref{eqn:accord} at $\lambda=0.45$ based on epBIC \eqref{eqn:epBIC} 
and then applying the debiasing procedure \eqref{eqn:d-accord}.
We extracted non-zero partial correlations to form edges and calculated the frequency of intra-modality and inter-modality pairs. \Cref{fig:GD}A shows the frequency of each type of inter- and intra-modality edges, 
with separate positive and negative edge counts.

Since the DNA methylation probes outnumbered both mRNA- and miRNA expression variables in the input data, the majority of the selected edges were intra-modality correlations between methylation probes ($>$99\%), which showed predominantly positive partial correlations (98\%) at all ranges of genomic distances. The second most frequent edges were between mRNAs and between miRNAs, where the vast majority showed positive partial correlations (96\%), and the correlations were close to zero for the pairs associated with negative partial correlations. This observation clearly reaffirms that physically proximal protein coding genes and non-coding RNAs are often co-transcribed \citep{ribeiro2022shared, shine2024co}. Among the small number of non-zero inter-modality correlations, the most pronounced category was that of edges between DNA methylation probes and mRNA expression (total transcript level per gene); the signs of the partial correlations were evenly split between positive and negative (\Cref{fig:GD}A).

When we examined the genomic distances between connected nodes in the six categories, positive inter-modality correlations were more likely to come from the data feature pairs located on the same chromosomes and within 200 kilobase distance of one another (\Cref{fig:GD}B). The edges connecting mRNAs and miRNAs showed exclusively positive partial correlations within a 100 kilobase distance only, hinting at high local specificity of co-transcription. Last but not least, the edges connecting mRNAs and DNA methylation probes showed both positive and negative partial correlations. A closer examination revealed that the largest negative partial correlations were between mRNAs and DNA methylation probes within a 1 kbp distance (from TSS), consistent with the established role of DNA methylation in the regulatory regions for gene expression repression. 
We further investigated biological functions enriched in the genes with negative correlation between DNA methylation and mRNA expression. The results showed that the DNA methylation-mediated gene expression repression were mostly observed in the genes encoding subunits of enzymes involved in small molecule metabolism and redox reaction regulation (\Cref{fig:GD}C), suggesting that epigenetic regulation is an active repressor for the gene expression of metabolic enzymes in liver tumors.%

\subsection{Dissecting active regulatory contribution of TFs and repressive epigenetic regulation}

\begin{figure}[ht!]%
    \centering
    \subfloat{{\includegraphics[width=1\textwidth]{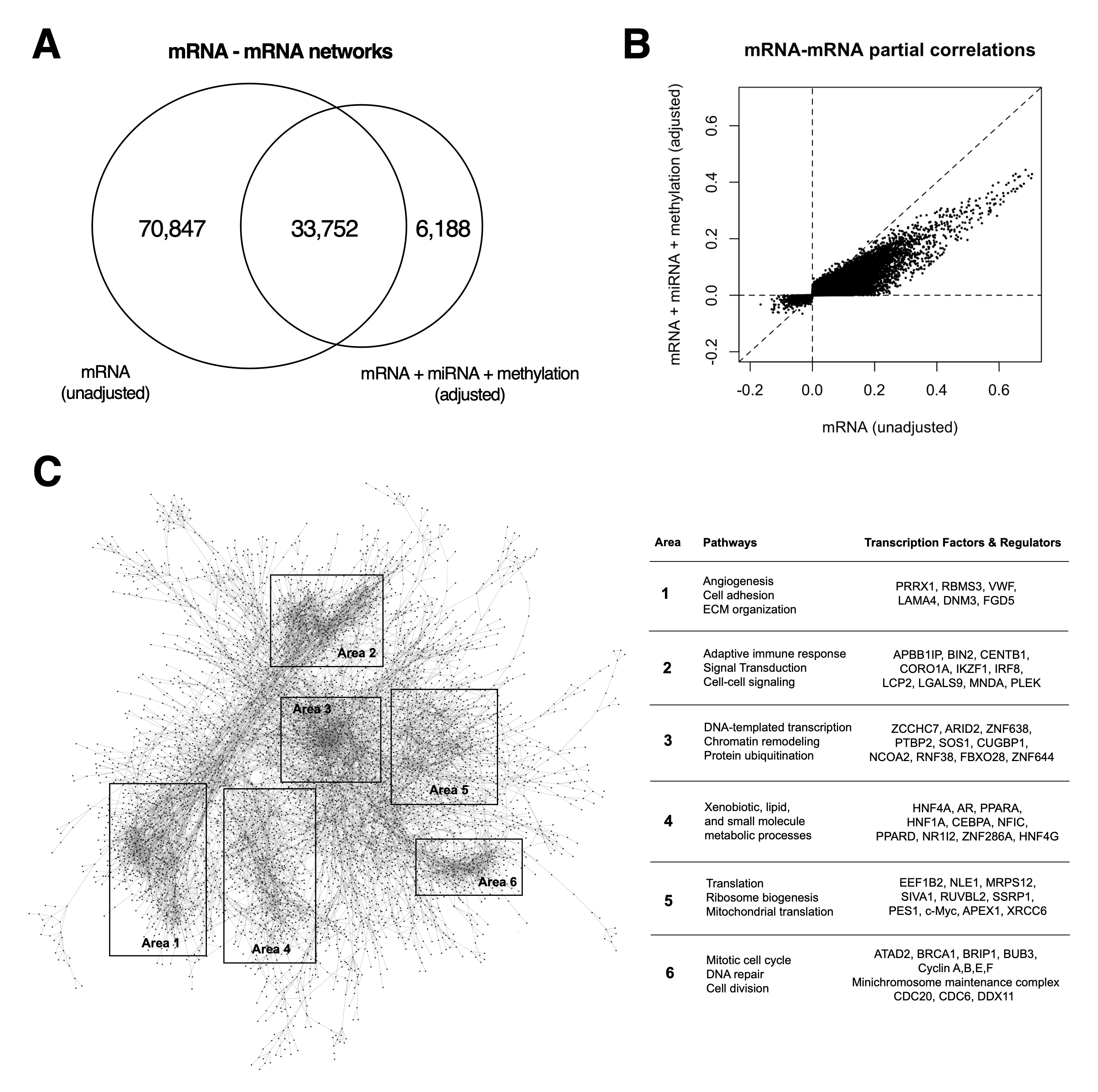}}} 
    \caption{ (A) Comparison of the two partial correlation networks of mRNA expression data estimated with and without miRNA and DNA methylation data. (B) Comparison of partial correlation estimates for 110,787 edges appearing in the two mRNA-mRNA networks. (C) The network estimated with DNA methylation data. The able on the right side shows the biological processes and TFs enriched in each area. }%
    \label{fig:NC}%
\end{figure}

Upon understanding the overall structure of the graph, we next investigated the regulatory impact of DNA methylation on gene expression levels and dissected the regulatory contribution of TFs on their target gene expression from the repressive effects of DNA methylation in the regulatory regions. To this end, we estimated another graph of mRNA expression variables using only the mRNA expression data as input, thereby establishing a reference network which does not exclude the impact of epigenetic regulation. We then compared the resulting mRNA-mRNA network to the one estimated earlier. For the comparison, we removed edges with absolute partial correlation smaller than 0.02 in the network visualization, an arbitrary threshold determined from the histogram of non-zero partial correlation estimates, in order to minimize the impact of false positive edges as mentioned in \S\ref{sec:exp}. 

The new mRNA-mRNA network contained 104,599 edges at $\lambda=0.30$ based on epBIC. We observed that the graph estimated without the DNA methylation data, called the unadjusted network hereafter, contained 84.5\% of the edges from the graph estimated with the DNA methylation data, called the adjusted network (\Cref{fig:NC}A). In the edges appearing in both networks, partial correlation estimates in the unadjusted network were greater in magnitude than those in the adjusted network, as expected (\Cref{fig:NC}B). 
Overall, the analysis clearly show that a significant portion of the conditional dependence relationship between two mRNA nodes can be explained away by DNA methylation levels, highlighting the contribution of epigenetic elements in the co-expression patterns. 

Since the impact of DNA methylation-driven repression of mRNA expression was accounted for, it is reasonable to hypothesize that the adjusted network data allows the analyst to infer active co-regulation driven by TFs and co-activators without the potential confounding by epigenetic regulation. We thus visualized the adjusted network using Cytoscape software \citep{shannon2003cytoscape} and selected six sub-networks of high connectivity (\Cref{fig:NC}C). We performed hypergeometric probability-based test for the enrichment of biological functions and TFs in each subnetwork. The subnetworks showed specific enrichment of cancer-associated biological processes, ranging from extracellular matrix remodeling in the parenchymal environment, immune response and signal transduction, cell cycle and protein translation, and small molecular metabolism with previously well-characterized TFs in various cellular processes. By contrast, the unadjusted network showed a higher degree of connectivity between the nodes (visualization not shown due to lack of legibility), and the overall network lacked clear separation of subnetworks. 

\begin{figure}[h]%
    \centering
    \subfloat{{\includegraphics[width=1\textwidth]{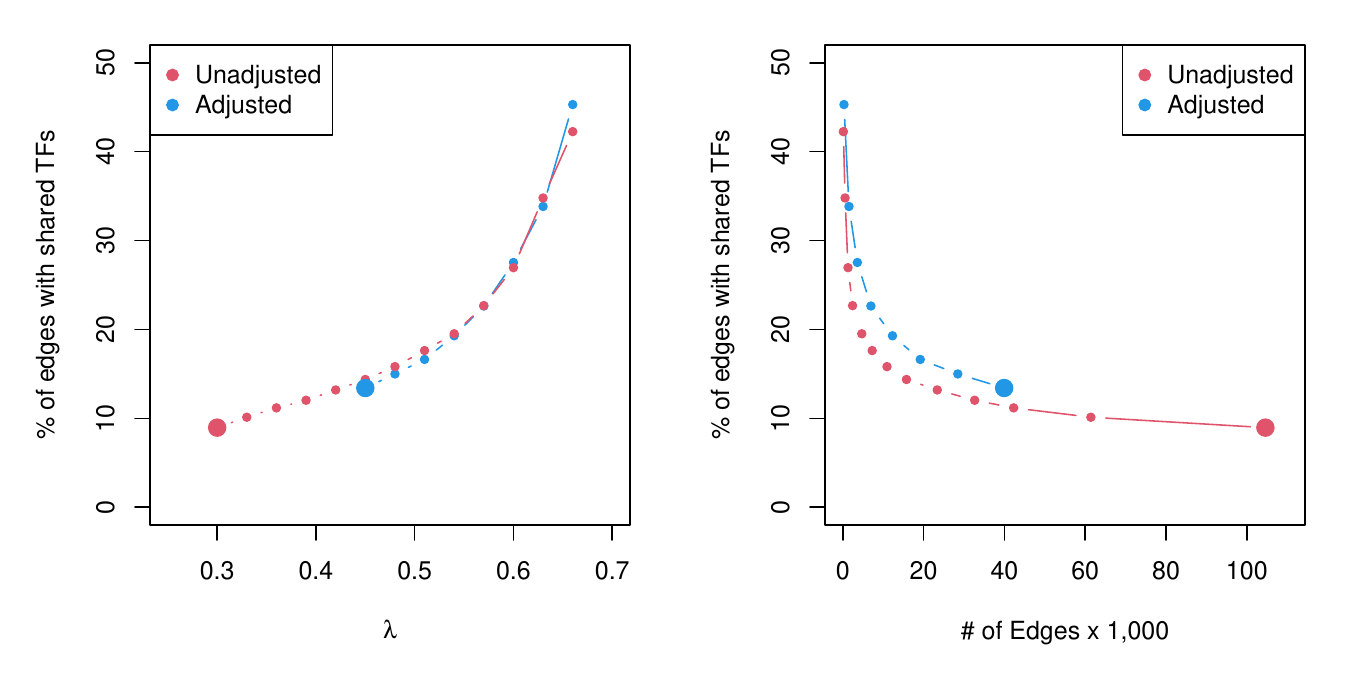}}}%
    \caption{
    Recovery of potentially co-regulated gene pairs sharing one or more common TFs in human cells at a range of $\lambda$ values and at the corresponding network sizes. The curves show the percentages of TF-sharing gene pairs in the selected edges in the DNA methylation adjusted network at $\lambda = 0.45, 0.48, \ldots, 0.66$ (blue) and in the unadjusted network at $\lambda = 0.30, 0.33, \ldots, 0.51$ (red). The large circles correspond to the $\lambda$ values optimized using epBIC in the two networks.  }%
    \label{fig:CR}%
\end{figure}

To investigate the quality of the two networks in terms of the recovery of co-regulated genes, we compared how often the reported mRNA-mRNA edges have previously been validated using a compendium of \textit{bona fide} TF-target regulatory relationships compiled from credible sources. Specifically, we calculated the proportion of edges connecting genes with at least one shared TFs in the validated TF-target relationships \citep{zhao2005tred, zheng2008itfp, rouillard2016harmonizome, han2015trrust}. The comparison showed that the adjusted network has greater enrichment of gene pairs sharing common TFs and co-activators (13.6\%) than the unadjusted network (9.4\%) at the respective optimal $\lambda$ values selected using epBIC (\Cref{fig:CR}, left panel). Considering that the latter network is three times as large as the former, as indicated by large circles in the diagrams (right panel), the results reaffirm that the network derived from the analysis accounting for the impact of DNA methylation confers greater specificity in the reported gene-to-gene connections. When we compared the networks with more stringent selection of edges at equivalent network sizes, the pattern also remained consistent (right panel): the adjusted network captures gene pairs sharing common TFs better than the unadjusted network. 

In summary, the evaluation shows that the network derived from the dual-omic data generates more robust relational hypotheses than the network derived from the transcriptomic data alone. 
By directly estimating conditional dependence structure from ultrahigh-dimensional data, spurious associations can be screened out and the filtered data improves the quality of biological inference of regulatory relationships. 
It goes without saying that the superior performance in the present case study comes from the scalable computation enabled by the HP-ACCORD framework.

\section{Discussion}\label{sec:disc}
In discovery-oriented clinical and molecular biology research, technological advances have steadily increased the number of variables that can be analyzed. Typical data sets have more than tens or hundreds of thousands of variables. The latest technological advances also push sample sizes beyond thousands, best evidenced by the surge in single-cell resolution profiles in the literature. When making biological inferences from these truly big data, 
striking a balance between statistical performance and computational scalability is therefore essential.%

The analysis in \S\ref{sec:HCC} is a testament to this trade-off. 
We acknowledge that the validity of individual edges reported in the analysis should be subjected to experimental confirmation. However, we verified that the mRNA co-expression network estimated from the ultra high-dimensional dual-omic data was largely nested within the network obtained from mRNA expression variables alone, not deviating to a completely distinct result. Moreover, the analysis incorporating epigenomic profiles allowed us to recover potentially co-regulated gene pairs in higher percentage with a smaller network than the latter, through the evaluation for the recovery of \textit{bona fide} TF-target relationships validated in human cells. 

The analysis presented in this paper would have been impossible without the scalable computation. In an era where new omics modalities are constantly being added to the multi-omics repertoire, graphical model estimation in ultra-high-dimensional spaces will become increasingly necessary for data integration, and computational scalability will remain as essential as ever. We believe HP-ACCORD paves the way to meeting this unmet demand and encourages consideration of computational scalability in the development of other statistical frameworks in their backbone.%

\clearpage

\bibliographystyle{abbrvnat} 
\bibliography{ref}       

\begin{thebibliography}{47}
\providecommand{\natexlab}[1]{#1}
\providecommand{\url}[1]{\texttt{#1}}
\expandafter\ifx\csname urlstyle\endcsname\relax
  \providecommand{\doi}[1]{doi: #1}\else
  \providecommand{\doi}{doi: \begingroup \urlstyle{rm}\Url}\fi

\bibitem[Ally et~al.(2017)Ally, Balasundaram, Carlsen, Chuah, Clarke, Dhalla, Holt, Jones, Lee, Ma, et~al.]{ally2017comprehensive}
A.~Ally, M.~Balasundaram, R.~Carlsen, E.~Chuah, A.~Clarke, N.~Dhalla, R.~A. Holt, S.~J. Jones, D.~Lee, Y.~Ma, et~al.
\newblock Comprehensive and integrative genomic characterization of hepatocellular carcinoma.
\newblock \emph{Cell}, 169\penalty0 (7):\penalty0 1327--1341, 2017.

\bibitem[Bauschke and Combettes(2011)]{Bauschke:ConvexAnalysisAndMonotoneOperatorTheoryIn:2011}
H.~H. Bauschke and P.~L. Combettes.
\newblock \emph{Convex analysis and monotone operator theory in {H}ilbert spaces}.
\newblock Springer Science \& Business Media, New York, NY, USA, 2011.

\bibitem[Beck and Teboulle(2009)]{Beck:SiamJournalOnImagingSciences:2009}
A.~Beck and M.~Teboulle.
\newblock A fast iterative shrinkage-thresholding algorithm for linear inverse problems.
\newblock \emph{SIAM J. Imaging Sci.}, 2\penalty0 (1):\penalty0 183--202, 2009.

\bibitem[Cai et~al.(2011)Cai, Liu, and Luo]{cai2011constrained}
T.~Cai, W.~Liu, and X.~Luo.
\newblock A constrained $\ell_1$ minimization approach to sparse precision matrix estimation.
\newblock \emph{J. Amer. Statist. Assoc.}, 106\penalty0 (494):\penalty0 594--607, 2011.

\bibitem[Candes and Tao(2007)]{candes2007}
E.~Candes and T.~Tao.
\newblock {The Dantzig selector: Statistical estimation when p is much larger than n}.
\newblock \emph{Ann. Statist.}, 35\penalty0 (6):\penalty0 2313 -- 2351, 2007.
\newblock URL \url{https://doi.org/10.1214/009053606000001523}.

\bibitem[Combettes and Pesquet(2007)]{combettes2007douglas}
P.~L. Combettes and J.-C. Pesquet.
\newblock A {D}ouglas--{R}achford splitting approach to nonsmooth convex variational signal recovery.
\newblock \emph{IEEE J. Sel. Top. Signal Process.}, 1\penalty0 (4):\penalty0 564--574, 2007.

\bibitem[Combettes and Pesquet(2011)]{combettes2011proximal}
P.~L. Combettes and J.-C. Pesquet.
\newblock Proximal splitting methods in signal processing.
\newblock In \emph{Fixed-point algorithms for inverse problems in science and engineering}, pages 185--212. Springer, 2011.

\bibitem[d'Aspremont et~al.(2008)d'Aspremont, Banerjee, and El~Ghaoui]{d2008first}
A.~d'Aspremont, O.~Banerjee, and L.~El~Ghaoui.
\newblock First-order methods for sparse covariance selection.
\newblock \emph{SIAM J. Matrix Anal. Appl.}, 30\penalty0 (1):\penalty0 56--66, 2008.

\bibitem[Dhillon et~al.(2007)Dhillon, Guan, and Kulis]{dhillon2007weighted}
I.~S. Dhillon, Y.~Guan, and B.~Kulis.
\newblock Weighted graph cuts without eigenvectors a multilevel approach.
\newblock \emph{IEEE transactions on pattern analysis and machine intelligence}, 29\penalty0 (11):\penalty0 1944--1957, 2007.

\bibitem[Foygel and Drton(2010)]{foygel2010extended}
R.~Foygel and M.~Drton.
\newblock Extended bayesian information criteria for gaussian graphical models.
\newblock In \emph{{A}dvances in {N}eural {I}nformation {P}rocessing {S}ystems}, volume~23. Curran Associates, 2010.

\bibitem[Friedman et~al.(2008)Friedman, Hastie, and Tibshirani]{friedman2008sparse}
J.~Friedman, T.~Hastie, and R.~Tibshirani.
\newblock Sparse inverse covariance estimation with the graphical lasso.
\newblock \emph{Biostatistics}, 9\penalty0 (3):\penalty0 432--441, 2008.
\newblock ISSN 1465-4644, 1468-4357.

\bibitem[Han et~al.(2015)Han, Shim, Shin, Shim, Ko, Shin, Kim, Cho, Kim, Lee, et~al.]{han2015trrust}
H.~Han, H.~Shim, D.~Shin, J.~E. Shim, Y.~Ko, J.~Shin, H.~Kim, A.~Cho, E.~Kim, T.~Lee, et~al.
\newblock Trrust: a reference database of human transcriptional regulatory interactions.
\newblock \emph{Scientific reports}, 5\penalty0 (1):\penalty0 11432, 2015.

\bibitem[Hasin et~al.(2017)Hasin, Seldin, and Lusis]{hasin2017}
Y.~Hasin, M.~Seldin, and A.~Lusis.
\newblock Multi-omics approaches to disease.
\newblock \emph{Genome Biol.}, 18\penalty0 (1):\penalty0 83, 2017.

\bibitem[Hsieh et~al.(2011)Hsieh, Dhillon, Ravikumar, and Sustik]{hsieh2011sparse}
C.-J. Hsieh, I.~Dhillon, P.~Ravikumar, and M.~Sustik.
\newblock Sparse inverse covariance matrix estimation using quadratic approximation.
\newblock In \emph{Advances in Neural Information Processing Systems}, volume~24. Curran Associates, 2011.

\bibitem[Hsieh et~al.(2013)Hsieh, Sustik, Dhillon, Ravikumar, and Poldrack]{hsieh2013big}
C.-J. Hsieh, M.~A. Sustik, I.~S. Dhillon, P.~K. Ravikumar, and R.~Poldrack.
\newblock {{BIG}} \& {{QUIC}}: Sparse inverse covariance estimation for a million variables.
\newblock In \emph{Advances in Neural Information Processing Systems}, volume~26. Curran Associates, 2013.

\bibitem[Hsieh et~al.(2014)Hsieh, Sustik, Dhillon, Ravikumar, et~al.]{hsieh2014quic}
C.-J. Hsieh, M.~A. Sustik, I.~S. Dhillon, P.~Ravikumar, et~al.
\newblock {QUIC}: quadratic approximation for sparse inverse covariance estimation.
\newblock \emph{J. Mach. Learn. Res.}, 15\penalty0 (1):\penalty0 2911--2947, 2014.

\bibitem[Huang et~al.(2017)Huang, Chaudhary, and Garmire]{huang2017}
S.~Huang, K.~Chaudhary, and L.~Garmire.
\newblock More is better: Recent progress in multi-omics data integration methods.
\newblock \emph{Front. Genet.}, 8:\penalty0 84, 2017.

\bibitem[Karypis and Kumar(1998)]{karypis1998fast}
G.~Karypis and V.~Kumar.
\newblock A fast and high quality multilevel scheme for partitioning irregular graphs.
\newblock \emph{SIAM Journal on scientific Computing}, 20\penalty0 (1):\penalty0 359--392, 1998.

\bibitem[Khare et~al.(2015)Khare, Oh, and Rajaratnam]{khare2015convex}
K.~Khare, S.~Oh, and B.~Rajaratnam.
\newblock A convex pseudolikelihood framework for high dimensional partial correlation estimation with convergence guarantees.
\newblock \emph{J. R. Stat. Soc., B: Stat. Methodol.}, 77\penalty0 (4):\penalty0 803--825, 2015.
\newblock ISSN 1369-7412.

\bibitem[Koanantakool et~al.(2016)Koanantakool, Azad, Bulu{\c{c}}, Morozov, Oh, Oliker, and Yelick]{koanantakool2016communication}
P.~Koanantakool, A.~Azad, A.~Bulu{\c{c}}, D.~Morozov, S.~Oh, L.~Oliker, and K.~Yelick.
\newblock Communication-avoiding parallel sparse-dense matrix-matrix multiplication.
\newblock In \emph{2016 {IEEE} International Parallel and Distributed Processing Symposium}, pages 842--853. IEEE, 2016.

\bibitem[Koanantakool et~al.(2018)Koanantakool, Ali, Azad, Buluc, Morozov, Oliker, Yelick, and Oh]{koanantakool2018communication}
P.~Koanantakool, A.~Ali, A.~Azad, A.~Buluc, D.~Morozov, L.~Oliker, K.~Yelick, and S.~Oh.
\newblock Communication-avoiding optimization methods for distributed massive-scale sparse inverse covariance estimation.
\newblock In \emph{International Conference on Artificial Intelligence and Statistics}, volume~84, pages 1376--1386. PMLR, 2018.

\bibitem[Lange et~al.(2022)Lange, Won, Landeros, and Zhou]{lange2022nonconvex}
K.~Lange, J.-H. Won, A.~Landeros, and H.~Zhou.
\newblock Nonconvex optimization via {MM} algorithms: Convergence theory.
\newblock In W.~Piegorsch, R.~Levine, H.~Zhang, and T.~C.~M. Lee, editors, \emph{Computational Statistics in Data Science}, pages 509--534. John Wiley \& Sons, 2022.

\bibitem[Li et~al.(2010)Li, Toh, et~al.]{li2010inexact}
L.~Li, K.-C. Toh, et~al.
\newblock An inexact interior point method for {$L_1$}-regularized sparse covariance selection.
\newblock \emph{Math. Program. Comput.}, 2\penalty0 (3-4):\penalty0 291--315, 2010.

\bibitem[Luo et~al.(2014)Luo, Song, and Witten]{luo2014sure}
S.~Luo, R.~Song, and D.~Witten.
\newblock Sure screening for gaussian graphical models.
\newblock \emph{arXiv preprint arXiv:1407.7819}, 2014.

\bibitem[Mazumder et~al.(2019)Mazumder, Wright, and Zheng]{mazumder_computing_2019}
R.~Mazumder, S.~Wright, and A.~Zheng.
\newblock Computing estimators of {D}antzig selector type via column and constraint generation, aug 2019.
\newblock URL \url{http://arxiv.org/abs/1908.06515}.
\newblock arXiv:1908.06515 [math, stat].

\bibitem[Meinshausen(2007)]{MEINSHAUSEN2007374}
N.~Meinshausen.
\newblock Relaxed lasso.
\newblock \emph{Comput. Statist. Data Anal.}, 52\penalty0 (1):\penalty0 374--393, 2007.
\newblock ISSN 0167-9473.
\newblock URL \url{https://www.sciencedirect.com/science/article/pii/S0167947306004956}.

\bibitem[Meinshausen and B{\"u}hlmann(2006)]{meinshausen2006high}
N.~Meinshausen and P.~B{\"u}hlmann.
\newblock High-dimensional graphs and variable selection with the {{Lasso}}.
\newblock \emph{Ann. Statist.}, 34\penalty0 (3):\penalty0 1436--1462, 2006.
\newblock ISSN 0090-5364, 2168-8966.

\bibitem[Meinshausen and B{\"u}hlmann(2010)]{meinshausen2010stability}
N.~Meinshausen and P.~B{\"u}hlmann.
\newblock Stability selection.
\newblock \emph{J. R. Stat. Soc., B: Stat. Methodol.}, 72\penalty0 (4):\penalty0 417--473, 2010.

\bibitem[Oh et~al.(2014)Oh, Dalal, Khare, and Rajaratnam]{oh2014optimization}
S.-Y. Oh, O.~Dalal, K.~Khare, and B.~Rajaratnam.
\newblock Optimization methods for sparse pseudo-likelihood graphical model selection.
\newblock In \emph{Advances in Neural Information Processing Systems}, volume~27, pages 667--675. Curran Associates, 2014.

\bibitem[Pang et~al.(2014)Pang, Liu, and Vanderbei]{pang2014fastclime}
H.~Pang, H.~Liu, and R.~J. Vanderbei.
\newblock The fastclime package for linear programming and large-scale precision matrix estimation in {R}.
\newblock \emph{J. Mach. Learn. Res.}, 15:\penalty0 489--493, 2014.

\bibitem[Pang et~al.(2017)Pang, Liu, Vanderbei, and Zhao]{pang_parametric_2017}
H.~Pang, H.~Liu, R.~J. Vanderbei, and T.~Zhao.
\newblock Parametric simplex method for sparse learning.
\newblock In \emph{{A}dvances in {N}eural {I}nformation {P}rocessing {S}ystems}, volume~30. Curran Associates, 2017.

\bibitem[Peng et~al.(2009)Peng, Wang, Zhou, and Zhu]{peng2009partial}
J.~Peng, P.~Wang, N.~Zhou, and J.~Zhu.
\newblock Partial correlation estimation by joint sparse regression models.
\newblock \emph{J. Amer. Statist. Assoc.}, 104\penalty0 (486):\penalty0 735--746, 2009.

\bibitem[Ravikumar et~al.(2011)Ravikumar, Wainwright, Raskutti, and Yu]{ravikumar2011high}
P.~Ravikumar, M.~J. Wainwright, G.~Raskutti, and B.~Yu.
\newblock {High-dimensional covariance estimation by minimizing $\ell_1$-penalized log-determinant divergence}.
\newblock \emph{Electron. J. Stat.}, 5:\penalty0 935 -- 980, 2011.
\newblock URL \url{https://doi.org/10.1214/11-EJS631}.

\bibitem[Ribeiro et~al.(2022)Ribeiro, Ziyani, and Delaneau]{ribeiro2022shared}
D.~M. Ribeiro, C.~Ziyani, and O.~Delaneau.
\newblock Shared regulation and functional relevance of local gene co-expression revealed by single cell analysis.
\newblock \emph{Communications biology}, 5\penalty0 (1):\penalty0 876, 2022.

\bibitem[Rouillard et~al.(2016)Rouillard, Gundersen, Fernandez, Wang, Monteiro, McDermott, and Ma’ayan]{rouillard2016harmonizome}
A.~D. Rouillard, G.~W. Gundersen, N.~F. Fernandez, Z.~Wang, C.~D. Monteiro, M.~G. McDermott, and A.~Ma’ayan.
\newblock The harmonizome: a collection of processed datasets gathered to serve and mine knowledge about genes and proteins.
\newblock \emph{Database}, 2016:\penalty0 baw100, 2016.

\bibitem[Shannon et~al.(2003)Shannon, Markiel, Ozier, Baliga, Wang, Ramage, Amin, Schwikowski, and Ideker]{shannon2003cytoscape}
P.~Shannon, A.~Markiel, O.~Ozier, N.~S. Baliga, J.~T. Wang, D.~Ramage, N.~Amin, B.~Schwikowski, and T.~Ideker.
\newblock Cytoscape: a software environment for integrated models of biomolecular interaction networks.
\newblock \emph{Genome Res.}, 13\penalty0 (11):\penalty0 2498--2504, 2003.

\bibitem[Shine et~al.(2024)Shine, Gordon, Sch{\"a}rfen, Zigackova, Herzel, and Neugebauer]{shine2024co}
M.~Shine, J.~Gordon, L.~Sch{\"a}rfen, D.~Zigackova, L.~Herzel, and K.~M. Neugebauer.
\newblock Co-transcriptional gene regulation in eukaryotes and prokaryotes.
\newblock \emph{Nature Reviews Molecular Cell Biology}, 25\penalty0 (7):\penalty0 534--554, 2024.

\bibitem[Tibshirani(2013)]{tibshirani2013lasso}
R.~J. Tibshirani.
\newblock The lasso problem and uniqueness.
\newblock \emph{Electron. J. Stat.}, 7:\penalty0 1456--1490, 2013.

\bibitem[van~de Geijn and Watts(1997)]{vandegeijn1997SUMMA}
R.~A. van~de Geijn and J.~Watts.
\newblock {{SUMMA}}: Scalable universal matrix multiplication algorithm.
\newblock \emph{Concurr. Comput. Pract. Exp.}, 9\penalty0 (4):\penalty0 255--274, 1997.
\newblock ISSN 1096-9128.

\bibitem[Vandereyken et~al.(2023)Vandereyken, Sifrim, Thienpont, and Voet]{vandereyken_2023}
K.~Vandereyken, A.~Sifrim, B.~Thienpont, and T.~Voet.
\newblock Methods and applications for single-cell and spatial multi-omics.
\newblock \emph{Nat. Rev. Genet.}, 24\penalty0 (8):\penalty0 494--515, mar 2023.

\bibitem[Vono et~al.(2022)Vono, Dobigeon, and Chainais]{vono2022high}
M.~Vono, N.~Dobigeon, and P.~Chainais.
\newblock High-dimensional {G}aussian sampling: a review and a unifying approach based on a stochastic proximal point algorithm.
\newblock \emph{SIAM Rev.}, 64\penalty0 (1):\penalty0 3--56, 2022.

\bibitem[Wainwright(2019)]{wainwright2019high}
M.~J. Wainwright.
\newblock \emph{High-dimensional statistics: A non-asymptotic viewpoint}.
\newblock Cambridge University Press, New York, NY, USA, 2019.

\bibitem[Yuan and Lin(2007)]{yuan2007model}
M.~Yuan and Y.~Lin.
\newblock Model selection and estimation in the {{Gaussian}} graphical model.
\newblock \emph{Biometrika}, 94\penalty0 (1):\penalty0 19--35, 2007.
\newblock ISSN 0006-3444.

\bibitem[Zhang and Horvath(2005)]{zhang2005}
B.~Zhang and S.~Horvath.
\newblock A general framework for weighted gene co-expression network analysis.
\newblock \emph{Statistical Applications in Genetics and Molecular Biology}, 4\penalty0 (1):\penalty0 1544, 2005.

\bibitem[Zhao et~al.(2005)Zhao, Xuan, Liu, and Zhang]{zhao2005tred}
F.~Zhao, Z.~Xuan, L.~Liu, and M.~Q. Zhang.
\newblock Tred: a transcriptional regulatory element database and a platform for in silico gene regulation studies.
\newblock \emph{Nucleic acids research}, 33\penalty0 (suppl\_1):\penalty0 D103--D107, 2005.

\bibitem[Zheng et~al.(2008)Zheng, Tu, Yang, Xiong, Wei, Xie, Zhu, and Li]{zheng2008itfp}
G.~Zheng, K.~Tu, Q.~Yang, Y.~Xiong, C.~Wei, L.~Xie, Y.~Zhu, and Y.~Li.
\newblock Itfp: an integrated platform of mammalian transcription factors.
\newblock \emph{Bioinformatics}, 24\penalty0 (20):\penalty0 2416--2417, 2008.

\bibitem[Zheng et~al.(2020)Zheng, Shi, Li, and Yuan]{ZHENG2020104645}
Z.~Zheng, H.~Shi, Y.~Li, and H.~Yuan.
\newblock Uniform joint screening for ultra-high dimensional graphical models.
\newblock \emph{Journal of Multivariate Analysis}, 179:\penalty0 104645, 2020.
\newblock ISSN 0047-259X.
\newblock URL \url{https://www.sciencedirect.com/science/article/pii/S0047259X20302268}.

\end{thebibliography}


\begin{thebibliography}{10}
\providecommand{\natexlab}[1]{#1}
\providecommand{\url}[1]{\texttt{#1}}
\expandafter\ifx\csname urlstyle\endcsname\relax
  \providecommand{\doi}[1]{doi: #1}\else
  \providecommand{\doi}{doi: \begingroup \urlstyle{rm}\Url}\fi

\bibitem[Bauschke and Combettes(2011)]{Bauschke:ConvexAnalysisAndMonotoneOperatorTheoryIn:2011_a}
H.~H. Bauschke and P.~L. Combettes.
\newblock \emph{Convex analysis and monotone operator theory in {H}ilbert spaces}.
\newblock Springer Science \& Business Media, New York, NY, USA, 2011.

\bibitem[Beck and Shtern(2017)]{beck_linearly_2017}
A.~Beck and S.~Shtern.
\newblock Linearly convergent away-step conditional gradient for non-strongly convex functions.
\newblock \emph{Math. Program.}, 164\penalty0 (1-2):\penalty0 1--27, 2017.
\newblock ISSN 0025-5610, 1436-4646.
\newblock URL \url{http://link.springer.com/10.1007/s10107-016-1069-4}.

\bibitem[Bolte et~al.(2017)Bolte, Nguyen, Peypouquet, and Suter]{bolte_error_2017}
J.~Bolte, T.~P. Nguyen, J.~Peypouquet, and B.~W. Suter.
\newblock From error bounds to the complexity of first-order descent methods for convex functions.
\newblock \emph{Math. Program.}, 165\penalty0 (2):\penalty0 471--507, 2017.
\newblock ISSN 0025-5610, 1436-4646.
\newblock URL \url{http://link.springer.com/10.1007/s10107-016-1091-6}.

\bibitem[Hoffman(1952)]{hoffman_approximate_1952}
A.~J. Hoffman.
\newblock On approximate solutions of systems of linear inequalities.
\newblock \emph{Journal of Research of the National Bureau of Standards}, 49\penalty0 (4):\penalty0 263--265, 1952.

\bibitem[Ko and Won(2019)]{ko2019optimal}
S.~Ko and J.-H. Won.
\newblock Optimal minimization of the sum of three convex functions with a linear operator.
\newblock In \emph{The 22nd International Conference on Artificial Intelligence and Statistics}, pages 1185--1194. PMLR, 2019.

\bibitem[Ko et~al.(2019)Ko, Yu, and Won]{ko2019easily}
S.~Ko, D.~Yu, and J.-H. Won.
\newblock Easily parallelizable and distributable class of algorithms for structured sparsity, with optimal acceleration.
\newblock \emph{Journal of Computational and Graphical Statistics}, 28\penalty0 (4):\penalty0 821--833, 2019.

\bibitem[{\L}ojasiewicz(1958)]{lojasiewicz1958division}
S.~{\L}ojasiewicz.
\newblock Division d'une distribution par une fonction analytique de variables r\'{e}elles.
\newblock \emph{Comptes Rendus Hebdomadaires Des Seances de l'Academie Des Sciences}, 246\penalty0 (5):\penalty0 683--686, 1958.

\bibitem[Ravikumar et~al.(2011)Ravikumar, Wainwright, Raskutti, and Yu]{ravikumar2011high_a}
P.~Ravikumar, M.~J. Wainwright, G.~Raskutti, and B.~Yu.
\newblock {High-dimensional covariance estimation by minimizing $\ell_1$-penalized log-determinant divergence}.
\newblock \emph{Electron. J. Stat.}, 5:\penalty0 935 -- 980, 2011.
\newblock URL \url{https://doi.org/10.1214/11-EJS631}.

\bibitem[Wainwright(2019)]{wainwright2019high_a}
M.~J. Wainwright.
\newblock \emph{High-dimensional statistics: A non-asymptotic viewpoint}.
\newblock Cambridge University Press, New York, NY, USA, 2019.

\bibitem[Zualinescu(2003)]{zualinescu_sharp_2003}
C.~Zualinescu.
\newblock Sharp estimates for {Hoffman}'s constant for systems of linear inequalities and equalities.
\newblock \emph{SIAM J. Optim.}, 14\penalty0 (2):\penalty0 517--533, 2003.
\newblock ISSN 1052-6234, 1095-7189.
\newblock URL \url{http://epubs.siam.org/doi/10.1137/S1052623402403505}.

\end{thebibliography}

\end{document}


\appendix
\title{Supplementary Materials for Learning Massive-scale Partial Correlation Networks in Clinical Multi-omics Studies with HP-ACCORD}
\date{}
\maketitle

\section{Proofs}\label{sec:proofs}
\subsection{Proof of Theorem 3.1}
The gradient of the risk $R$ is given by
$
    \nabla R(\bOmega) = -\bOmega_D^{-1} + \bOmega \bSigma^*
$
and the Hessian matrix (with respect to the vectorization of $\bOmega$) is
\begin{equation}\label{eq:hessian}
    \nabla^2R(\bOmega) = \bSigma^* \otimes \bI_p + (\bOmega_D^{-1}\otimes \bOmega_D^{-1})\bUpsilon,
\end{equation}
where $\otimes$ is a Kronecker product and $\bUpsilon = \sum_{i=1}^p e_i e_i^T\otimes e_i e_i^T$.

Since $\bOmega^*$ is positive definite, $\bSigma^{*-1} = \bTheta^* = \bOmega_D^*\bOmega^*$ and, therefore, $\bTheta_D^* = \bOmega_D^{*2}$. As a result, 
$$
    \bOmega^* = \bTheta_D^{*-1/2}\bTheta^*
$$
resides in $\dom{R}$ and makes $\nabla R(\bOmega^*) = 0$. So $\bOmega^*$ minimizes $R$ in $\mathbb{R}^{p\times p}$.

Now since the covariance $\bSigma^\ast$ is positive definite, so is the Hessian $\nabla^2 R(\bOmega)$ for $\bOmega \in \dom{R}$.
That is, $R$ strictly convex on $\dom{R}$.
Therefore, $\bOmega^*$ is the unique minimizer of the population risk $R$.

\subsection{Proof of Theorem 3.2}
Recall that $\bX \in \mathbb{R}^{n \times p}$ $(p \geq n)$ has columns in \emph{general position} if the affine span of any $n$ points $\{\sigma_1X_{i_1}, \cdots, \sigma_nX_{i_n}\}$, for arbitrary signs $\sigma_1, \cdots, \sigma_n \in \{-1, 1\}$, does not contain any element of $\{\pm X_i : i \neq i_{1},\cdots,i_{n}\}$, where $X_i$ is the $i$-th column of $\bX$.

The objective function of the ACCORD estimation problem (3.2) can be expressed as
\begin{equation}\label{eqn:vectf}
   \tilde{f}(\tilde{\bX}\vect(\bOmega^T))
   := 
   \sum_{i=1}^p \{ \tilde{g}(\tilde{\bX}^{i}\Omega^T_{i}) + \lambda \Vert\Omega^T_{i}\Vert_1 \}
   ,
\end{equation}
where $\Omega^T_{i}$ is $i$-th column of $\bOmega^T$,
$\tilde{\bX}^{i} := [e_i, \bX^T]^T$, $\tilde{g}([a, b^T]^T) := -\log(a) + \frac{1}{2n} \Vert b \Vert_2^2$,
and $\tilde{\bX} = \text{diag}(\tilde{\bX}^{1}, \cdots, \tilde{\bX}^{p})$.
Here, we see that $\tilde{g}$ is strictly convex on its natural domain.

\begin{lemma}\label{lem:accordsolprops}
For any $\bS = \frac{1}{n}\bX^T\bX$ and $\lambda \geq 0$, solutions to problem (3.2) have the following properties:
    \begin{enumerate}
        \item there is either a unique solution or an (uncountably) infinite number of solutions;
        \item every solution $\hat{\bOmega}$ has the same values of $\hat{\bOmega}_D$ and $\tilde{\bX}\vect(\hat{\bOmega}^T)$;
        \item if $\lambda > 0$, then every solution $\hat{\bOmega}$ has the same $\ell_1$ norm $\Vert \hat{\bOmega} \Vert_1$.
    \end{enumerate}
\end{lemma}

\begin{proof}
    Since the objective function \eqref{eqn:vectf}
    is a convex coercive function, a solution always exists. Denote the optimal value of $\tilde{f}$ as $f^\star$. If there exists two distinct solutions $\bOmega^{(1)}$ and $\bOmega^{(2)}$, then $\alpha \bOmega^{(1)} + (1-\alpha)\bOmega^{(2)}$ is also a solution for any $0 < \alpha < 1$ since all the level set of $f$ is convex. Then, we have
\begin{align*}
    &f^\star =\tilde{f}\left(\tilde{\bX}\vect\left(\alpha{\bOmega^{(1)}}^T + (1-\alpha){\bOmega^{(2)}}^T \right)\right)\\
    &=\sum_{i=1}^p\left\{\tilde{g}\left(\tilde{\bX}^{i}\left(\alpha{\Omega^{(1)}_i}^T + (1-\alpha){\Omega^{(2)}_i}^T \right)\right)\right\} + \lambda \Vert \alpha \bOmega^{(1)} + (1-\alpha)\bOmega^{(2)} \Vert_1\\
    &\leq \alpha\sum_{i=1}^p \left\{ \tilde{g}(\tilde{\bX}^{i}{\Omega^{(1)}_i}^T) + \lambda \Vert{\Omega^{(1)}_i}^T\Vert_1 \right\} + 
    (1-\alpha)\sum_{i=1}^p \{ \tilde{g}(\tilde{\bX}^{i}{\Omega^{(2)}_i}^T) + \lambda \Vert{\Omega^{(2)}_i}^T\Vert_1 \}\\
    &= \alpha \tilde{f}(\tilde{\bX}\vect({\bOmega^{(1)}}^T)) + (1-\alpha)\tilde{f}(\tilde{\bX}\vect({\bOmega^{(2)}}^T)) = \alpha f^\star + (1-\alpha) f^\star = f^\star.
\end{align*}
Thus the inequality must hold with equality.
Note that this inequality arises from the strict convexity of $\tilde{g}$, thus equality holds if and only if $\tilde{\bX}\vect({\bOmega^{(1)}}^T) = \tilde{\bX}\vect({\bOmega^{(2)}}^T)$, which implies $\bOmega^{(1)}_D = \bOmega^{(2)}_D$ and $\bX{\Omega^{(1)}_i}^T = \bX{\Omega^{(2)}_i}^T$, $i=1,\cdots, p$.
It then follows that $\|\bOmega^{(1)}\|_1 = \|\bOmega^{(2)}\|_1$.
\end{proof} 

The Karush-Kuhn-Tucker (KKT) optimality condition for (3.2) can be written as
\begin{equation}\label{eqn:KKT1}
    \hat{\bOmega}_D^{-1} - n^{-1}\bX^T\bX\hat{\bOmega}^T = \lambda \bZ,
\end{equation}
where
\begin{equation}\label{eqn:KKT2}
    z_{ij} \in
    \begin{cases}
        \{\text{sign}(\hat{\omega}_{ij})\} & \text{ if } \hat{\omega}_{ij} \neq 0, \\
        [-1, 1] & \text{ if } \hat{\omega}_{ij} = 0,
    \end{cases}
\end{equation}
for $\hat{\bOmega} = (\hat{\omega}_{ij})$ and $\bZ = (z_{ij})$.
Note that $z_{ii}$ is always $1$ as $\hat{\omega}_{ii}$ is always positive for all $i=1,\dotsc,p$. Let $r_i$ be the $i$-th column of $n^{-1}\bX\hat{\bOmega}^T $. We define the equicorrelation set $\mathcal{E}$ by
$$
\mathcal{E} = \bigcup_{i=1}^p\{(j, i): j \in \mathcal{E}_i\}, \quad 
\mathcal{E}_i = \{j \in (1,\cdots, p): |X_j^Tr_i| = \lambda\} \cup \{i\}
.
$$
The KKT condition \eqref{eqn:KKT1}--\eqref{eqn:KKT2} implies that $\hat{\omega}_{ji} = 0$ if $(j,i) \notin \mathcal{E}$, and thus 
problem (3.2) is equivalent to finding a minimizer of
$\sum_{i=1}^p \left\{ \tilde{g}(\tilde{\bX}^{i}_{\mathcal{E}_i}\Omega^T_{i,\mathcal{E}_i}) + \lambda\Vert\Omega^T_{i,\mathcal{E}_i}\Vert_1 \right\}$,
where $\tilde{\bX}^{i}_{\mathcal{E}_i}$ is a submatrix consisting of columns of $\tilde{\bX}^i$ with column indices in $\mathcal{E}_i$ and $\Omega^T_{i,\mathcal{E}_i}$ is a vector consisting of $\omega_{ij}$ for $j \in \mathcal{E}_i$. 
In light of the proof of \Cref{lem:accordsolprops},
we see that if $\textbf{null}(\tilde{\bX}^i_{\mathcal{E}_i}) = \{0\}$ for every $i$, then the minimizer is unique.
In order to find a sufficient condition for this to hold, suppose the case where $\textbf{null}(\tilde{X}_{\mathcal{E}_i}) \neq \{0\}$ for some $i$. Then, 
there exists $\mathcal{D} \subset \mathcal{E}_i$ with at most $n$ elements such that 
$$
X_j = \sum_{k \in \mathcal{D}\setminus\{j\}}c_kX_k
$$
for some $j \in \mathcal{D}$ and $c_k \in \mathbb{R}\setminus\{0\}$. Note that $i \notin \mathcal{D}$ as $\tilde{\bX}^i_i = [1, X_i^T]^T$ cannot be spanned by the columns of $\tilde{\bX}^i_{-i} = [0, \bX_{-i}^T]^T$. 
Taking the inner product with $r_i$ on both sides of the equation above yields
$$
s_{ij}\lambda = \sum_{k \in \mathcal{D}\setminus\{j\}} s_{ik}c_k\lambda = \sum_{k \in \mathcal{D}\setminus\{j\}} (s_{ik}s_{ij}c_k) (s_{ij}\lambda)
$$
where $s_{ik} \in \{-1, 1\}$ for $k \in \mathcal{D}$, since $|X_j^Tr_i| = \lambda$ for $j \in \mathcal{E}_i \setminus \{i\}$. So for $\lambda > 0$, we have
$\sum_{k \in \mathcal{D}\setminus\{j\}} (s_{ik}s_{ij}c_k) = 1$.
In other words, $X_j$ is an affine combination of $\pm X_k$'s, $k\in\mathcal{D}\setminus\{j\}$. Thus, we conclude that if there exist more than one solutions, the columns of $\bX$ are not in the general position.

\subsection{Proof of Theorem 3.2}
Recall (3.4) that the ACCORD objective function is $f(\bOmega) = g(\bOmega) + h(\bOmega)$ where
$g(\bOmega) = ({1}/{2})\tr(\bOmega^T\bOmega\bS)$
and
$h(\bOmega) = -\log\det\bOmega_D + \lambda\Vert \bOmega \Vert_{1}$.

Convergence of $\{\bOmega^{(t)}\}$ to $\hat{\bOmega}$  is a standard result in the operator splitting literature; see, e.g., \citet{Bauschke:ConvexAnalysisAndMonotoneOperatorTheoryIn:2011_a}, \citet{ko2019easily}, and \citet{ko2019optimal}. Monotone convergence of $\{f(\bOmega^{(t)})\}$ to $f^{\star}=\min_{\bOmega} f(\bOmega)$ follows from the continuity of $f$ and the descent property of the algorithm.

In order to establish the rate of convergence, we first prove a {\L}ojasiewicz error bound \citep{lojasiewicz1958division}.
Recall the expression \eqref{eqn:vectf}.
Let $y = \vect(\bOmega^T) \in \mathbb{R}^{p^2}$ and write 
\begin{align*}
	\bar{g}(\bZ) &= 
	\sum_{i=1}^p \tilde{g}(z_i)
	,
	\quad
	\bar{h}(\bOmega) = \bar{h}(y)
	= \lambda\Vert y \Vert_1
	,
	\\
	\tilde{f}(\bOmega) &= \tilde{f}(\tilde{\bX}\bOmega^T) = \bar{g}(\tilde{\bX}y) + \bar{h}(y),
\end{align*}
where $\bZ = [z_1^T, \cdots, z_p^T]^T$ for $z_i \in \mathbb{R}^{n+1}$; $\tilde{g}$ and $\tilde{\bX}$ are defined in \eqref{eqn:vectf}.

The smooth part $\bar{g}$ is a proper closed convex function 
and strongly convex on any compact convex subset in its natural domain $\{\bZ \in \mathbb{R}^{(n+1)p}: \bZ_{i,1} > 0\}$.
The nonsmooth part $\bar{h}$, with $\dom\bar{h}=\mathbb{R}^{p^2}$, is piece-wise linear, thus its epigraph is polyhedral.
Let $y^{\star} = \vect(\hat{\bOmega}^T)$ be the unique minimizer of $f$.
Then, $f(y^{\star}) \leq f(\vect(\bI)) = \text{tr}(\bS)/2 + \lambda p$ and
$L_0 = \{y \in \mathbb{R}^{p^2}: f(y) \leq \text{tr}(\bS)/2  + \lambda p \}$ is nonempty, compact, and convex. Obviously, $y^{\star} \in L_0$.
Now let
\[
	R = \max\{\Vert y \Vert_1: y \in L_0\} \in (0, \infty)
\]
so that $\Vert y \Vert_1 \leq R$ whenever $y \in L_0$. $\Vert y^{\star}\Vert_1 \leq R$ also holds. We also split $y$ into $y_D$ and $y_X$, where each corresponds to the elements of $\bOmega_D$ and $\bOmega_X$. Let $\bP$ a permutation matrix such that
$$
y = \bP [y_D^T, \; y_X^T]^T.
$$
Then,
\begin{align*}
    \min_{y\in\mathbb{R}^{p^2}} f(y)
    &=
    \min\{\bar{g}(\tilde{\bX}y) + \bar{h}(y): y \in L_0\}
    \\
    &= 
    \min\{\bar{g}(\tilde{\bX}y) + \lambda\Vert y \Vert_1 : \Vert y \Vert_1 \leq R \}
    \\
    &=
    \min\bigg\{\bar{g}\left(\tilde{\bX}\bP 
    [y_D^T, \; {y_X^{+}}^T - {y_X^{-}}^T]^T
    \right)
    + \lambda(\mathbf{1}^T y_D + \mathbf{1}^T y_X^{+} + \mathbf{1}^T y_X^{-})\\
    &\qquad: \mathbf{1}^T y_D + \mathbf{1}^T y_X^{+} + \mathbf{1}^T y_X^{-} \leq R, \by_D \geq 0, y^{+}_X \geq 0, y^{-}_X \geq 0 \bigg\} 
    \\
    &=
    \min\{\bar{g}(\bE\tilde{y}) + b^T\tilde{y}: \bA\tilde{y} \leq a \}
    ,
\end{align*}
where  
$b = \lambda 1$,
$\bE = \tilde{\bX}\bP\tilde{\bE}$, and
\begin{align*}
    \tilde{y}
    &= 
    \begin{bmatrix}
    y_D \\ y_X^{+} \\ y_X^{-} 
    \end{bmatrix}
    ,
    \quad
    \tilde{\bE} = 
    \begin{bmatrix}
        \bI & \mathbf{0} & \mathbf{0}  \\
        \mathbf{0} & \bI         & -\bI   \\
    \end{bmatrix}
    ,
    \quad
    \bA =
    \begin{bmatrix}
        1^T & 1^T & 1^T  \\
        -\bI         & \mathbf{0} & \mathbf{0}  \\
        \mathbf{0} & -\bI         & \mathbf{0}   \\
        \mathbf{0} & \mathbf{0}   & -\bI         \\
    \end{bmatrix}
    ,
    \quad
    a =
    \begin{bmatrix}
        R \\ 0
    \end{bmatrix}
    ,
    \end{align*}
and where $0$'s and $1$'s denote entries of zeros and ones with an appropriate size. Note 
the constraint set $\tilde{Y} = \{ \tilde{y} \in \mathbb{R}^{4p^2}: \bA \tilde{y} \leq a \}$ is a compact polyhedron 
in which $\bar{g}$ is strongly convex with parameter 
$$\nu(\tilde{Y}):= \min\left\{\frac{1}{\max_{\tilde{y} \in \tilde{Y}}\Vert \tilde{y}_D\Vert_\infty^2}, \frac{1}{2n}\right\}$$
Now, let us define the diameters and radii of the sets
\begin{align*}
	D &= \max\{ \Vert \tilde{y}_1 - \tilde{y}_2 \Vert_2 : \tilde{y}_1, \tilde{y}_2 \in \tilde{Y}\} < \infty
	,
	\\
	D_E &= \max\{ \Vert \bE\tilde{y}_1 - \bE\tilde{y}_2 \Vert_2 : \tilde{y}_1, \tilde{y}_2 \in \tilde{Y}\} \leq \vvvert \bE \vvvert_{2} D
	,
	\\
	G &= \max\{ \Vert \bE\tilde{y} \Vert_2: \tilde{y} \in \tilde{Y} \}
	\leq \vvvert \bE \vvvert_{2} D
	,
\end{align*}
where $\vvvert\bE\vvvert_{r} = \sup_{v \neq 0}\Vert \bE v \Vert_r/\Vert v \Vert_r$ is the operator $r$-norm.
We also let $\tilde{Y}^{\star} = \argmin_{\tilde{y} \in \tilde{Y}}\{\allowbreak\bar{g}(\bE\tilde{y}) + b^T\tilde{y}\}$, which is a nonempty and compact set. 
Then, by \citet[Lemma 2.5]{beck_linearly_2017},
for any $\tilde{y} \in \tilde{Y}$,
\begin{equation}\label{eqn:KL0}
	\dist^2(\tilde{y}, \tilde{Y}^{\star})
	\leq
	\kappa[ \bar{g}(\bE\tilde{y}) + b^T\tilde{y} - \min_{\bar{y} \in \tilde{Y}}\{\bar{g}(\bE\bar{y}) + b^T\bar{y}\} ]
	,
\end{equation}
where $\dist(p, S) = \inf_{s \in S}\Vert p - s \Vert_2$ and
\begin{equation}\label{eqn:EBconstant}
	\kappa =
	\theta^2(\Vert b \Vert_2 D + 3G D_E + 2(G^2 + 1)/\nu(\tilde{Y}))
\end{equation}
for the Hoffman constant $\theta$ that only depends on $\bA$ and $\bE$ \citep{hoffman_approximate_1952}.
This constant is characterized as
\[
	\theta = \max_{\bB \in \mathcal{B}} 1/\lambda_{\min}(\bB\bB^T)
	,
\]
where $\lambda_{\min}$ denotes the smallest eigenvalue and $\mathcal{B}$ is the set of matrices constructed by linearly independent rows of $[\bE^T, \bA^T]^T$, and can be estimated from $\bA$ and $\bE$ \citep{zualinescu_sharp_2003}.

Going back to the original variable, observe that for any $y \in \mathbb{R}^{p^2}$ such that $\Vert y \Vert_1 \leq R$ and $y_D \geq 0$, there exists $\tilde{y} \in \tilde{Y}$ such that 
\[
	y = \bP\tilde{\bE}\tilde{y}, \quad f(y) = \tilde{f}(\tilde{\bX}y) = \bar{g}(\tilde{\bX}\bP\tilde{\bE}\tilde{y}) + b^T\tilde{y} = \bar{g}(\bE\tilde{y}) + b^T\tilde{y},
\]
and vice versa.
Denote $\tilde{y}^{\star}$ as the projection of $\tilde{y}$ onto $\tilde{Y}^{\star}$.
Since $y^{\star}$, the minimizer of $f$, is unique, we also have $y^{\star} = \bP\tilde{\bE}\tilde{y}^{\star}$.
It also holds that 
\[
	f^{\star} := \min_{\by\in\mathbb{R}^{p^2}}f(y) = f(y^{\star}) = \bar{g}(\bE\tilde{y}^{\star}) + b^T\tilde{y}^{\star} = \min_{\bar{y}\in \tilde{Y}}\{\bar{g}(\bE\bar{y}) + b^T\bar{y}\}.
\]
Then,
\[
    \Vert y - y^{\star} \Vert_2
    \leq
    \vvvert \tilde{\bE} \vvvert_{2}
    \Vert \tilde{y} - \tilde{y}^{\star} \Vert_2\\
    =
    \sqrt{2}
    \dist(\tilde{y}, Y^{\star})
    .
\]
It follows from the inequality \eqref{eqn:KL0} that for any $\bOmega \in \mathbb{R}^{p\times p}$ such that $\Vert \bOmega \Vert_1 \leq R$,
\[
	\Vert \bOmega - \hat{\bOmega} \Vert_F^2
	\leq 
	2\kappa[ f(\bOmega) - f^{\star} ]
	,
\]
i.e., the {\L}ojasiewicz error bound inequality holds.
Then, from \citet[Theorem 5]{bolte_error_2017}, 
the KL inequality 
\begin{equation}\label{eqn:KL1}
	\varphi'(f(\bOmega) - f^{\star})\dist(0, \partial f(\bOmega)) \geq 1
\end{equation}
holds with $\varphi(s) = 2\sqrt{2\kappa s}$ for all $\bOmega$ such that $\|\bOmega\|_1 \leq R$ and $f(\bOmega) > f^{\star}$.
The inverse $\psi: [0,\infty) \ni y\mapsto y^2/(8\kappa)$ of $\varphi$ satisfies Assumption (\textbf{A}) of \citet{bolte_error_2017}.

Now consider the proposed splitting (3.4).
If $\tau_t \in [\underline{\tau}, \bar{\tau}]$ with $0 < \underline{\tau} \leq \bar{\tau} < 2/L$, then \citet[Proposition 13]{bolte_error_2017} asserts that Assumptions (\textbf{H1}) and (\textbf{H2}) of \citet{bolte_error_2017} are satisfied with $a=1/\bar{\tau} - L/2$ and $b=1/\underline{\tau}+L$.
The non-asymptotic complexity bound (3.6) follows immediately from \citet[Corollary 20]{bolte_error_2017}. 

If $\tau_t$ is chosen by the line search, then we have
$$
g(\bOmega^{(t+1)}) \leq g(\bOmega^{(t)}) + \langle \nabla g(\bOmega^{(t)}), \bOmega^{(t+1)} - \bOmega^{(t)} \rangle + \frac{1}{2\tau_t}\|\bOmega^{(t+1)} - \bOmega^{(t)}\|_F^2. 
$$
where
$$
\bOmega^{(t+1)} 
=\argmin_{\bOmega} \left\{\langle \nabla g(\bOmega^{(t)}), \bOmega - \bOmega^{(t)} \rangle + \frac{1}{2\tau_t}\|\bOmega - \bOmega^{(t)}\|_F^2 + h(\bOmega)\right\}
.
$$
From the convexity of $h$, the latter is equivalent to
\begin{equation}\label{eqn:subgradient}
	-\nabla g(\bOmega^{(t)}) - \tau_t^{-1}(\bOmega^{(t+1)}-\bOmega^{(t)}) \in \partial h(\bOmega^{(t+1)})
	,
\end{equation}
which implies
\begin{align*}
	h(\bOmega^{(t)}) 
	&\geq
	h(\bOmega^{(t+1)}) + \langle -\nabla g(\bOmega^{(t)}) - \tau_t^{-1}(\bOmega^{(t+1)} - \bOmega^{(t)}), \bOmega^{(t)} - \bOmega^{(t+1)} \rangle
	\\
	&=
	h(\bOmega^{(t+1)}) + \langle \nabla g(\bOmega^{(t)}), \bOmega^{(t+1)} - \bOmega^{(t)} \rangle + \tau_t^{-1}\|\bOmega^{(t+1)} - \bOmega^{(t)}\|_F^2
\end{align*}
Combining the above two inequalities yields
$$
	g(\bOmega^{(t+1)}) + h(\bOmega^{(t+1)})
	+ \frac{1}{2\tau_t}\|\bOmega^{(t+1)} - \bOmega^{(t)}\|_F^2
	\leq
	g(\bOmega^{(t)}) + h(\bOmega^{(t)})
	.
$$
Since $\tau_t \leq \tau_0$ for all $t$, Assumption (\textbf{H1}) of \citet{bolte_error_2017} is satisfied with $a=1/(2\tau_0)$.
Now from the condition \eqref{eqn:subgradient} there is $s^{(t+1)} \in \partial h(\bOmega^{(t+1)})$ such that 
$s^{(t+1)} + \nabla g(\bOmega^{(t)}) + \tau_t^{-1}(\bOmega^{(t+1)}-\bOmega^{(t)}) = 0$. This entails
\begin{align*}
	\|s^{(t+1)} + \nabla g(\bOmega^{(t+1)})\|_F
	&\leq
	\|s^{(t+1)} + \nabla g(\bOmega^{(t)})\|_F
	+ \|\nabla g(\bOmega^{(t)}) - \nabla g(\bOmega^{(t+1)}\|_F
	\\
	&\leq
	\tau_t^{-1}\|\bOmega^{(t+1)} - \bOmega^{(t)}\|_F
	+ L \|\bOmega^{(t)} - \bOmega^{(t+1)}\|_F
	\\
	&\leq
	(\tau_{\min}^{-1} + L)\|\bOmega^{(t+1)} - \bOmega^{(t)}\|_F
	,
\end{align*}
since $\tau_t \geq \tau_{\min}$.
Invoking that $s^{(t+1)} + \nabla g(\bOmega^{(t+1)}) \in \partial f(\bOmega^{(t+1)})$, we see that Assumption (\textbf{H2}) of \citet{bolte_error_2017} is satisfied with $b=\tau_{\min}^{-1}+L$.
The non-asymptotic complexity bound (3.6) follows again from \citet[Corollary 20]{bolte_error_2017}.

\subsection{Proof of Theorem 4.1}
Recall that the empirical ACCORD risk is $\ell(\bOmega) = -\log\det\bOmega_D
+ (1/2)\text{tr}(\bOmega^T\bOmega\bS)$ and the ACCORD estimator (3.2) is defined as a minimizer of $\ell(\bOmega) + \lambda\|\bOmega\|_1$ for $\lambda \geq 0$.
The Hessian matrix of the loss is
\begin{equation}\label{eqn:hess}
   \nabla^2\ell(\bOmega) = \bS \otimes \bI_p + \sum_{i=1}^p \omega_{ii}^{-2}(e_i e_i^T \otimes e_i e_i^T) = \bS \otimes \bI_p + (\bOmega_D^{-1} \otimes \bOmega_D^{-1})\bUpsilon,
\end{equation}
where $\bUpsilon = \sum_{i=1}^p e_i e_i^T\otimes e_i e_i^T$,
if the matrix variable $\bOmega$ is vectorized in column-major order, i.e., by $\vect(\bOmega)=(\omega_{11}, \dotsc, \omega_{p1}, \dotsc, \omega_{1p}, \dotsc, \omega_{pp})$; $\otimes$ is the Kronecker product.

To prove the claimed error bounds, we follow the proof of Theorem 9.36 in \citet{wainwright2019high_a}. First define the error function
\begin{equation}\label{eq:errorE}
    \mathcal{E}(\bDelta) 
    := \ell(\bOmega^\ast + \bDelta) - \ell(\bOmega^\ast) - \langle \nabla \ell(\bOmega^\ast), \bDelta \rangle,
\end{equation} 
and verify its restricted strong convexity 
\begin{equation}\label{eq:RSCcondition}
    \mathcal{E}(\bDelta) \geq \kappa \Vert\bDelta\Vert_F^2, \quad \kappa > 0,~ \bDelta \in \mathbb{C}(S), 
\end{equation}
where $S$ is the support of $\bOmega^*$ and $\mathbb{C}(S) := \{\bDelta : \Vert\bDelta_{S^c}\Vert_1 \leq 3 \Vert\bDelta_{S}\Vert_1\}$. 
Then, the claimed error bounds are achieved in the event of $G(\lambda)=\{\|\nabla\ell(\bOmega^*)\|_{\infty} \leq \lambda/2\}$, which is a sufficient condition for $\hat{\bDelta} = \hat{\bOmega} - \bOmega^* \in \mathbb{C}(S)$.

\subsubsection{Verifying restricted strong convexity}
\begin{theorem}[Restricted strong convexity of the ACCORD loss]\label{thm:rsc}
    Suppose the data matrix $\bX \in \mathbb{R}^{n \times p}$ is composed of $n$ i.i.d. copies of zero-mean continuous random vector $X=(X_1,\dotsc, X_p)\in\mathbb{R}^p$ with covariance matrix $\bSigma^*= (\Sigma^*_{ij}) \allowbreak = {\bTheta^*}^{-1}$ and each $X_j/\sqrt{\Sigma^*_{jj}}$ being sub-Gaussian with parameter $\sigma$.
    Also, suppose that there exist $\alpha, \beta, \eta > 0$ such that 
    $\E_{X}|\langle X, y\rangle|^{2} \geq \alpha$ and
    $\E_{X}|\langle X, y\rangle|^{2+\eta} \leq \beta^{2+\eta}$ for any $y \in \mathbb{R}^p$ with $\Vert y \Vert _2 = 1$.
    Let $S = \{(i,j)\in[p]\times[p]: \theta^*_{ij} \neq 0\}$ be the support of $\bTheta^*$ (hence of $\bOmega^*)$.
    and each $X_j/\sqrt{\Sigma^*_{jj}}$ is sub-Gaussian with parameter $\sigma$. 
    Then, for the error function $\mathcal{E}$ \eqref{eq:errorE}, 
    the inequality
    \begin{equation}\label{eq:cRSC}
        \mathcal{E}(\bDelta) \geq \kappa \Vert \bDelta \Vert_F^2  - c_0 \sqrt{\frac{\log p}{n}}\Vert\bDelta\Vert_F\Vert\bDelta\Vert_1, \quad   
    \end{equation}
    holds with a probability of at least $1 - c_1e^{-c_2 n},$ for some positive constants $\kappa, c_0, c_1$, and $c_2$ that explicitly depends on $\alpha, \beta, \eta, \sigma$, and $\max_{i \in [p]}\Sigma^*_{ii}$.
\end{theorem}
\begin{remark}
    If \eqref{eq:cRSC} holds, then the restricted strong convexity condition
    \begin{equation}\label{eq:rsc_notol}
        \mathcal{E}(\bDelta)
        \geq \left(\kappa - 4c_0 \sqrt{|S|\frac{\log p}{n}}\right) \Vert\bDelta\Vert_F^2 \geq \frac{\kappa}{2}\Vert\bDelta\Vert_F^2, 
    \end{equation}
    holds for  $\bDelta \in \mathbb{C}(S)$ as $\Vert\bDelta\Vert_1 = \Vert\bDelta_S\Vert_1 + \Vert\bDelta_{S^c}\Vert_1 \leq 4\Vert\bDelta_S\Vert_1 \leq 4\sqrt{|S|}\Vert\bDelta\Vert_F$, 
    provided that $\sqrt{n^{-1}|S|\log p} \leq \kappa/(8c_0)$.
\end{remark}
\begin{proof}

Let $X^i \in \mathbb{R}^p$ be the $i$-th observation of the data matrix $\bX$ and
$y_j \in \mathbb{R}^p$ be the $j$-th column of $\bDelta^T$.
Using Taylor's remainder theorem and the expression \eqref{eqn:hess} of the Hessian of $\ell$, we have
$$
    \mathcal{E}(\bDelta) = \vect(\bDelta)^T \nabla ^2 \ell(\bOmega^\ast + t \bDelta)\vect(\bDelta)
    \geq \vect(\bDelta)^T(\bS \otimes \bI_p)\vect(\bDelta)
    = \frac{1}{n} \sum_{i=1}^n\sum_{j=1}^p |\langle X^i, y_j \rangle|^2
$$
for some $0 < t < 1$. Define a truncating function 
$$
\varphi_K(x) = \begin{cases}
    |x|^2, & \text{ if } |x| \leq K, \\
    K^2, & \text{ if } |x| > K,
\end{cases}
$$
for some $K > 0.$ Here, it suffices to prove that for any $y \in \mathbb{R}^p$ with $\Vert y \Vert_2 =1$,
\begin{equation}\label{eq:scaledrsc}
    \frac{1}{n}\sum_{i=1}^n \varphi_K( \langle X^i, y \rangle)
    \geq c_3 - c_4 \sqrt{\frac{\log 2p}{n}}\Vert y \Vert_1
\end{equation}
holds for some $K, c_3,$ and $c_4$ with a probability at least $1 - c_1 e^{-c_2 n}$. To see this, let $\Vert y_j \Vert_2 = t_j$. Without loss of generality, assume that $t_j > 0$ for all $j$. 
If \eqref{eq:scaledrsc} holds for all $y$ with $\|y\|_2=1$, then since  $\|y_j/t_j\|_2=1$, it follows that
\begin{align*}
    \frac{1}{n}\sum_{i=1}^n & \varphi_{t_j K}( \langle X^i, y_j \rangle)  =
    \frac{1}{n}\sum_{i=1}^n t_j^2\varphi_{K}( \langle X^i, y_j/t_j \rangle)
    \overset{\eqref{eq:scaledrsc}}{\geq} c_3 t_j^2 - c_4 \sqrt{\frac{\log 2p}{n}}\Vert y_j \Vert_1 t_j 
    \\
    \Rightarrow\; \mathcal{E}(\bDelta) &\geq \frac{1}{n}\sum_{j=1}^p\sum_{i=1}^n \varphi_{t_j K}( \langle X^i, y_j \rangle)
    \geq c_3 \sum_{j=1}^p t_j^2 - 2c_4 \sqrt{\frac{\log p}{n}}\sum_{j=1}^p\Vert y_j \Vert_1 t_j \\
    & \geq c_3 \sum_{j=1}^p t_j^2 - 2c_4 \sqrt{\frac{\log p}{n}}\sqrt{\sum_{j=1}^p \Vert y_j \Vert_1^2} \sqrt{\sum_{j=1}^p t_j^2}
    \geq 
    c_3 \Vert \bDelta \Vert_F^2 - 2c_4 \sqrt{\frac{\log p}{n}} \Vert \bDelta \Vert_1 \Vert \bDelta \Vert_F,
\end{align*}
i.e., \eqref{eq:cRSC}
with $\kappa = c_3$ and $c_0 = 2c_4$, due to the Cauchy-Schwarz inequality and that $\sum_{j=1}^p\|y_j\|_1^2 \leq (\sum_{j=1}^p\|y_j\|_1)^2$.

To show \eqref{eq:scaledrsc},
define
\begin{align*}
    Z(r) &:= \inf_{\Vert y \Vert_2 = 1, \Vert y \Vert_1 \leq r}\left\{\frac{1}{n}\sum_{i=1}^n \varphi_K( \langle X^i, y \rangle) - \Expect_{X} [\varphi_K(\langle X, y \rangle)] \right\}\\
    &= -\sup_{\Vert y \Vert_2 = 1, \Vert y \Vert_1 \leq r}\left\{\frac{1}{n}\sum_{i=1}^n -\varphi_K( \langle X^i, y \rangle) - \Expect_{X} [-\varphi_K(\langle X, y \rangle)]\right\}.
\end{align*}
Note that if
\begin{equation}\label{eq:exp_bound}
    \Expect_X [\varphi_K(\langle X, y \rangle)] \geq \frac{3}{4} \alpha, 
\end{equation}
and
\begin{equation}\label{eq:tail_bound}
    \quad Z(r) \geq - \alpha / 2 - c_4' r\sqrt{\frac{\log 2p}{n}} 
\end{equation}
holds for some $c_4' > 0$, then
$$
\frac{1}{n}\sum_{i=1}^n \varphi_K( \langle X^i, y \rangle) \geq \Expect_X [\varphi_K(\langle X, y \rangle)] - \alpha/2 - c_4' r\sqrt{\frac{\log 2p}{n}} \geq \alpha/4 - c_4' r\sqrt{\frac{\log 2p}{n}}
$$
and thus
\eqref{eq:scaledrsc} with the $\|y\|_1$ replaced by $r$ holds with $c_3 = \alpha /4$ and $c_4 = c_4'$, provided that $\Vert y \Vert_1 \leq r$.

To show the expectation bound \eqref{eq:exp_bound}, observe that
\begin{align*}
    \Expect_X [\varphi_K(\langle X, y \rangle)] &\geq \Expect_X [|\langle X, y \rangle|^2 \cdot I(|\langle X, y \rangle| \leq K)] \\
    &\geq \alpha - \Expect_X [|\langle X, y \rangle|^2 \cdot I(|\langle X, y \rangle| > K)],
\end{align*}
so that it suffices to show that the last term is at most $\alpha/4$. 
From the given conditions, we have
$$
    \Pr(|\langle X, y\rangle| > K) \leq \Expect_X[|\langle X, y\rangle|^{2+\eta}]/K^{2+\eta} \leq \beta^{2+\eta}/K^{2+\eta}
$$
by Markov's inequality. Hence, the H\"older's inequality yields
$$
\Expect_X [|\langle X, y \rangle|^2 \cdot I(|\langle X, y \rangle| > K)] \leq (\Expect[|\langle X, y \rangle|^{2+\eta}])^{\frac{2}{2+\eta}}(\Pr(|\langle X, y\rangle| > K))^{\frac{\eta}{2+\eta}} \leq  {\beta^{2+\eta}}/{K^{\eta}}.
$$
Thus, $K^{\eta} = 4\beta^{2+\eta}/\alpha$ results in \eqref{eq:exp_bound}. In the sequel, we assume $K = (4\beta^{2+\eta}/\alpha)^{\frac{1}{\eta}}.$

To prove the tail bound \eqref{eq:tail_bound},
we need the following lemmas regarding a supremum of a functional on $\mathcal{F}$, a class of integrable real-valued functions with domain $\mathbb{R}^p$.
\begin{lemma}[Functional Hoeffding inequality; {\citealp[Theorem 3.26]{wainwright2019high_a}}]\label{lem:fhoeff}
    Suppose that every $f \in \mathcal{F}$ is uniformly bounded in $[a, b],$ i.e., $f(x) \in [a, b].$ Then, for $Z = -\sup_{f \in \mathcal{F}}\{\frac{1}{n}\sum_{i=1}^n f(X^i)\}$, where all $X^i$'s are independent random variables, we have
    $$
    \Pr(Z \leq \Expect[Z] - \delta) \leq \exp\left(-\frac{n\delta^2}{4(b-a)^2}\right).
    $$
    for any $\delta \geq 0.$
\end{lemma}
\begin{lemma}[{\citealp[Proposition 4.11]{wainwright2019high_a}}]\label{lem:rademacher}
    Let $(X^1, \cdots, X^n)$ be an i.i.d. sequence of some distribution, and let $(\epsilon_1, \cdots, \epsilon_n)$ be an i.i.d. sequence of Rademacher variables, i.e., random variables uniformly distributed in $\{-1, 1\}$. Then, we have
    $$
    \Expect_{X}\sup_{f \in \mathcal{F}}\left[\frac{1}{n}\sum_{i=1}^n f(X^i) - \Expect_{X}f(X)\right] \leq 2\Expect_{X,\epsilon} \left[\sup_{f \in \mathcal{F}}\frac{1}{n}\sum_{i=1}^n\epsilon_if(X^i)\right].
    $$
\end{lemma}
\begin{lemma}[Ledoux-Talagrand contraction inequality]\label{lem:cont}
    Suppose that $\phi_i : \mathbb{R} \to \mathbb{R}$ is a $L$-Lipschitz continuous function. Also, let $\epsilon_1, \cdots, \epsilon_n \in \{-1, 1\}$ be an i.i.d. sequence of Rademacher variables. Then, we have
    $$
    \Expect_{x, \epsilon}\left[\sup_{f \in \mathcal{F}}
    \sum_{i=1}^n \epsilon_i(\phi_i \circ f)(X^i)\right]
    \leq
    L \Expect_{x, \epsilon}\left[\sup_{f \in \mathcal{F}}
    \sum_{i=1}^n \epsilon_i f(X^i)\right]
    $$
\end{lemma}

\begin{lemma}[Expected $\ell_{\infty}$-norm of sub-Gaussian]\label{lem:inf-mom}
    Let $X = (X_1, \cdots, X_p) $ be a vector of sub-Gaussian random variables 
    with parameter $\nu$,
    where each random variable is not necessarily independent of each other. Then,
    $$
        \Expect_{X}[\max_i|X_i|] \leq \nu\sqrt{2\log 2p}.
    $$
\end{lemma}

Note that $\varphi_K$ is a 2K-Lipschitz continuous function, and all $X_1, \cdots, X_p$ are sub-Gaussian random variables with parameter $\nu := \sigma \sqrt{\max_{i\in[p]} \Sigma^\ast_{ii}}$. Then, we have the following inequality for the for the expectation of $Z(r)$.
\begin{equation}\label{eq:Zineq}
\begin{split}
    \Expect[Z(r)] 
    &= -\Expect_{X}\sup_{\Vert y \Vert_2 = 1, \Vert y \Vert_1 \leq r}\left\{\frac{1}{n}\sum_{i=1}^n -\varphi_K( \langle X^i, y \rangle) - \Expect_X [-\varphi_K(\langle X, y \rangle)] \right\}\\
    &\geq 
    -2\Expect_{X, \epsilon}\left[\sup_{\Vert y \Vert_2 = 1, \Vert y \Vert_1 \leq r}\frac{1}{n}\sum_{i=1}^n \epsilon_i\varphi_K( \langle X^i, y \rangle)\right] \quad \text{(\Cref{lem:rademacher})}\\ 
    &\geq
    -4K\Expect_{X, \epsilon}\left[\sup_{\Vert y \Vert_2 = 1, \Vert y \Vert_1 \leq r}\frac{1}{n}\sum_{i=1}^n \epsilon_i \langle X^i, y \rangle \right] \quad \text{(\Cref{lem:cont})}\\ 
    &\geq
    -4Kr\Expect_{X, \epsilon}\left\Vert\frac{1}{n} \sum_{i=1}^n\epsilon_i X^i\right\Vert_\infty  \quad\text{(H\"older's inequality)}\\
    &\geq -4\sqrt{2}K\nu r\sqrt{\frac{\log 2p}{n}} \quad\text{(\Cref{lem:inf-mom})}.
\end{split}
\end{equation}
The last inequality holds because each element of $n^{-1}\sum_{i=1}^{n}\epsilon_i X^i$ is sub-Gaussian with parameter $\nu/\sqrt{n}$.

On the other hand, since the value of $\varphi_K$ is nonnegative and bounded above by $K^2$, the range of $\varphi_K(\langle \cdot, y\rangle) - \E_X[\varphi_K(\langle X, y\rangle)]$ is bounded within $[-K^2, K^2].$ Hence, \Cref{lem:fhoeff} yields
\begin{equation}\label{eq:Zr}
\begin{split}
    & \Pr\left(Z(r) \leq - \alpha / 2 - (1 + 4\sqrt{2}K\nu)r\sqrt{\frac{\log 2p}{n}}\right) 
    \\
    & \leq 
    \Pr\left(Z(r) \leq \Expect[Z(r)] - \alpha / 2 - r\sqrt{\frac{\log 2p}{n}}\right) 
    \\
    &\leq \exp\left(-\frac{(r\sqrt{\frac{\log 2p}{n}} + \alpha / 2)^2}{16K^4}n\right)
    \leq \exp\left(-\frac{r^2\log 2p}{16K^4} - \frac{\alpha^2}{64K^4}n\right).    
\end{split}
\end{equation}
Thus
\eqref{eq:tail_bound} holds with a probability at least $1 - e^{-\frac{r^2\log 2p}{16K^4} - \frac{\alpha^2}{64K^4}n}$, for $c_4' = 1 + 4\sqrt{2}K\nu$.
Then, it follows that for all $y\in\mathbb{R}^p$ with $\Vert y \Vert_2 = 1$, $\Vert y \Vert_1 \leq r$, 
\eqref{eq:scaledrsc} with $\|y\|_1$ replaced by $r$ holds with a probability at least $1 - e^{-\frac{r^2\log 2p}{16K^4} - \frac{\alpha^2}{64K^4}n}$, for $c_3 = \alpha/4$, and $c_4 = c_4'$. In other words, 
\begin{equation}\label{eqn:rbound}
\begin{split}
& \Pr\left( \exists y ~\text{with}~ \|y\|_2=1, \|y\|_1\leq r ~\text{such that}~\frac{1}{n}\sum_{i=1}^n\varphi_K(\langle X^i, y \rangle) < \frac{\alpha}{4} - c_4'\sqrt{\frac{\log 2p}{n}} r \right)
\\
& =
\Pr\left( \inf_{y: \|y\|_2=1, \|y\|_1\leq r} \frac{1}{n}\sum_{i=1}^n\varphi_K(\langle X^i, y \rangle) -\frac{3}{4}\alpha < -\frac{\alpha}{2} - c_4'\sqrt{\frac{\log 2p}{n}} r \right)
\\
& \leq
\Pr\left( Z(r) < -\frac{\alpha}{2} - c_4'\sqrt{\frac{\log 2p}{n}} r \right)
\leq
\exp\left(-\frac{r^2\log 2p}{16K^4} - \frac{\alpha^2}{64K^4}n\right)
,
\quad
c_4' = 1 + 4\sqrt{2}K\nu
.
\end{split}
\end{equation}
To complete the proof, we need to show that \eqref{eq:scaledrsc} holds with a high probability independent of $\|y\|_1$ for some $c_3$ and $c_4$. To see this, let us choose $c_3=\alpha/4$ and $c_4 = 2c_4'$ in \eqref{eq:scaledrsc} and
$\mathcal{V}$ be the event that \eqref{eq:scaledrsc} with this choice is violated.
That is,
$$
    \mathcal{V} = \left\{\frac{1}{n}\sum_{i=1}^m\varphi_K(\langle X^i, y \rangle) < \frac{\alpha}{4} - 2c_4'\sqrt{\frac{\log 2p}{n}}\|y\|_1~\text{for some}~y\in\mathbb{R}^p~\text{with}~\|y\|_2 =1 \right\}
    .
$$
Suppose $\hat{y}\in\mathbb{R}^p$ violates \eqref{eq:scaledrsc}, with $\|\hat{y}\|_2 = 1$. If we define $A_m = \{y\in\mathbb{R}^p: \|y\|_2 = 1,~ 2^{m-1} \leq \|y\|_1 < 2^m\}$ for $m=1, 2, \dotsc$, then,
since $1 = \Vert y \Vert_2 \leq \Vert y \Vert_1 \leq \sqrt{p}$, we have
$\hat{y} \in A_m$ for some $m$. Thus
$$
    \inf_{y: \|y\|_2 = 1, \|y\|_1 \leq 2^{m}}\frac{1}{n}\sum_{i=1}^n\varphi_K(\langle X^i, y \rangle) 
    \leq \frac{1}{n}\sum_{i=1}^n\varphi_K(\langle X^i, \hat{y} \rangle)
    < \alpha/4 - c_4' 2^{m}\sqrt{(\log 2p)/{n}}
    .
$$
Therefore,
\begin{align*}
    \Pr(\mathcal{V}) &\leq 
    \sum_{m=1}^{\infty} \Pr\left(\inf_{y: \|y\|_2 = 1, \|y\|_1 \leq 2^{m}}\frac{1}{n}\sum_{i=1}^n\varphi_K(\langle X^i, y \rangle) 
    < \frac{\alpha}{4} - c_4' \sqrt{\frac{\log 2p}{n}} 2^{m}
    \right)
    \\
    &\stackrel{\eqref{eqn:rbound}}{\leq}
    \sum_{m=1}^{\infty} \exp\left(-\frac{4^{m}\log 2p}{16K^4} - \frac{\alpha^2}{64K^4}n\right)
    \\
    &\leq
    \sum_{m=1}^{\infty} \exp\left(-\frac{4^{m-3}}{K^4}\right) \exp\left( - \frac{\alpha^2}{64K^4}n\right)
    = c_1 \exp(-c_2 n)
    ,
\end{align*}
where $c_1 = \sum_{m=1}^\infty \exp\left(-\frac{4^{m-3}}{K^4}\right) = \sum_{m=1}^\infty \exp\left(-\frac{4^{m-3}}{(4\beta^{2+\eta}/\alpha)^{4/\eta}}\right) < \infty$ 
and $c_2 = \frac{\alpha^2}{64K^4} = \alpha^{2 + 4/\eta}/[64(4\beta^{2+\eta})^{4/\eta}]$.
Thus, \eqref{eq:scaledrsc} holds with probability $\Pr(\mathcal{V}^c) \geq 1 - c_1 e^{-c_2n}$ for $c_3=\alpha/4$ and $c_4=2(1+4\sqrt{2}(4\beta^{2+\eta}/\alpha)^{1/\eta}\nu)$, as desired.
It follows that \eqref{eq:cRSC} holds with $\kappa = \alpha/4$ and $c_0 = 4(1+4\sqrt{2}(4\beta^{2+\eta}/\alpha)^{1/\eta}\nu)$ with the same probability.
\end{proof}

\subsubsection{Verifying the event $G(\lambda)$}
We need an upper probability bound for the event
    $$
        G(\lambda)^c = \left\{\Vert \nabla \ell(\bOmega^*) \Vert_{\infty}  > \lambda/2 \right\} = \left\{\Vert -{\bOmega_D^*}^{-1} + \bOmega^*\bS \Vert_{\infty} > \lambda / 2 \right\}.
    $$
Recall that $\bTheta^* = \bOmega^*_D\bOmega^*$, which implies ${\bOmega^*_D}^{-1}=\bOmega^*\bTheta^{* -1}=\bOmega^*\bSigma^*$. Therefore,
    $$
        \Pr(G(\lambda)^c) \leq \Pr(\Vert \bOmega^* (-\bSigma^* + \bS) \Vert_{\infty}  \geq \lambda / 2) 
        \leq
        \Pr(\vvvert\bOmega^*\vvvert_{\infty}\Vert \bS - \bSigma^* \Vert_\infty \geq \lambda / 2).
    $$

From \citet[Lemma 1]{ravikumar2011high_a}, we know that
$$
    \Pr(\| \bS - \bSigma^* \|_{\infty} \geq t) 
    \leq
    4 \exp\left\{-\frac{nt^2}{128(1+4\sigma^2)^2(\max_{i\in[p]}\Sigma^*_{ii})^2} + 2\log p\right\}
$$
for $0 < t < 8(1+4\sigma^2)\max_{i\in[p]} \Sigma^*_{ii}.$ Thus, letting $$\lambda=(2\vvvert\bOmega^*\vvvert_{\infty}) \cdot 16(1+4\sigma^2)(\max_{i\in[p]}\Sigma^*_{ii})(\sqrt{n^{-1}\log p} + \delta)$$ 
upper-bounds 
$P(G(\lambda)^c) = 1 - P(G(\lambda))$
by $4\exp(-2n\delta^2)$ for the sample size $n$ such that $\sqrt{n^{-1}\log p} + \delta < 1/2$.

\subsubsection{Putting things together} 
We conclude that, by \citet[Theorem 9.19]{wainwright2019high_a}, if the event $\mathcal{V}^c \cap G(\lambda)$ occurs, then there holds
$$
    \Vert \hat{\bOmega} - \bOmega^\ast \Vert_F \leq 4\kappa^{-1}\lambda\sqrt{|S|} 
    \quad \text{and}\quad
    \Vert \hat{\bOmega} - \bOmega^\ast \Vert_1 \leq 4\sqrt{|S|}\Vert \hat{\bOmega} - \bOmega^\ast \Vert_F \leq 16\kappa^{-1}\lambda|S|
    .
$$
The probability of this event is bounded by
$$
    1 - \Pr(\mathcal{V} \cup G(\lambda)^c) \geq 1 - c_1e^{-c_2n} - 4e^{-2n\delta^2},
$$
provided that $\sqrt{n^{-1}|S|\log p} \leq \kappa/(8c_0)$ and $\sqrt{n^{-1}\log p} + \delta < 1/2$.
Choosing $\delta = \sqrt{n^{-1}\log p}$ yields the desired result.%

\subsection{Proof of Theorem 4.2}
We prove the theorem by proving an umbrella theorem under a more general tail condition.
\begin{definition}[Tail condition; {\citealp[Definition 1]{ravikumar2011high_a}}]\label{def:tail}
    Random vector $X$ with covariance matrix $\bSigma^\ast=(\Sigma^\ast_{ij})$ satisfies tail condition $\mathcal{T}(f,v_\ast)$ if there exists a constant $v_\ast \in (0,\infty]$ and a function $f : \mathbb{N} \times (0,\infty) \to (0,\infty)$, which is monotonically increasing in either argument, such that for any $(i,j) \in [p]\times [p]$ and sample estimate $\bS=(s_{ij})$ of $\bSigma^*$:
    \begin{equation}\label{eqn:tailcondition}
    P[|s_{ij}-\Sigma^\ast_{ij}| \geq \delta] \leq {1/}{f(n,\delta)}, 
    \quad \forall \delta \in (0,{1}/{v_\ast}]
    .
    \end{equation}
\end{definition}
We adopt the convention $1/0 = +\infty$, so that the value $v^* = 0$ indicates that inequality \eqref{eqn:tailcondition} holds for any $\delta \in (0, \infty)$.
Define inverses of $f$
$$
    \bar{n}_f(\delta, r) = \argmax\{n : f(n,\delta) \leq r\}
    \quad
    \text{and}
    \quad
    \bar{\delta}_f(n, r) = \argmax\{\delta : f(n,\delta) \leq r\}
    .
$$
It follows that $n > \bar{n}_f(\delta,r)$ for some $\delta > 0$ implies $\bar{\delta}_f(n,r) \leq \delta$.
\begin{theorem}\label{thm:consistency}
Suppose that the data matrix $\bX$ consists of $n$ i.i.d. copies of zero-mean continuous random vector $X\in\mathbb{R}^p$ satisfying the tail condition $\mathcal{T}(f, v_{\ast})$ of \Cref{def:tail}. Also suppose that the covariance matrix $\bSigma^*$ of $X$ satisfies Assumption 4.1.
Let $\bS=\frac{1}{n}\bX^T\bX$ and
$\hat{\bOmega}$ be the unique solution to the ACCORD problem in (3.2) with $\lambda = \lambda_n = (10\kappa_{\Omega^\ast}/\alpha)\bar\delta_f(n, p^\tau)$ for some $\tau > 2$.
If the sample size $n$ is such that
\begin{equation}\label{eqn:samplecomplexity}
	n > \bar{n}_f\left(
        {
	\min\left\{ \frac{\min\left\{\frac{1}{3\gamma_1}, \frac{1}{3\gamma_1^3\kappa_{\Gamma^\ast}}, \frac{\kappa_{\Omega^\ast}}{3d}\right\}}{3\kappa_{\Gamma^\ast}\kappa_{\Omega^*}\left(1 + \frac{10}{\alpha}\right)}
        , 
        \frac{2}{%
            27\gamma_1^3 
            \kappa_{\Gamma^\ast}^2\kappa_{\Omega^\ast}\left(1 + \frac{10}{\alpha}\right)^2%
        }
        ,
        \frac{1}{v_\ast}
    \right\}
    }
	,
	p^\tau
	\right)
 ,
\end{equation}
then for $C = 3\kappa_{\Gamma^\ast}\kappa_{\Omega^\ast}\left(1 + {10}/{\alpha}\right)$ and with a probability no smaller than $1 - p^{-(\tau-2)}$,
\begin{enumerate}[label=(\alph*)]
	\item there holds
		\begin{equation}\label{eqn:Linftybound}
			\|\hat{\bOmega} - \bOmega^\ast\|_{\infty}
			\leq
			C\:\bar\delta_f(n, p^\tau)
			,
		\end{equation}
        \item
            \begin{equation}\label{eq:theta_bound}
                \quad
                \|\hat{\bOmega}_D\hat{\bOmega} - \bTheta^*\|_\infty \leq (7/3)\kappa_{\Omega^*}\|\hat{\bOmega} - \bOmega^\ast\|_{\infty};
            \end{equation}
	\item the estimated support set $S(\hat{\bOmega})=\{(i,j)\in V\times V : \hat{\omega}_{ij} \neq 0,~i\neq j\}$ is contained in the true support $S$ and includes all coordinates $(i,j)$ with 
		$|\omega^\ast_{ij}| >%
			C\:\bar\delta_f(n, p^\tau)$%
		.
\end{enumerate}
\end{theorem}
\begin{proof}
Let
\begin{equation}\label{eqn:oracle}
    \tilde{\bOmega} = \argmin_{\bOmega_{S^c}=0}\left\{ -\log\det\bOmega_D + \frac{1}{2}\text{tr}(\bOmega^T\bOmega\bS) + \lambda\Vert \bOmega \Vert_{1} \right\}
\end{equation}
be the oracle solution, where $S$ denotes the support of $\bOmega^*$ (hence of $\bTheta^*={\bSigma^*}^{-1})$. This solution is also unique under the general position condition.
We want to find a condition such that $\hat{\bOmega} = \tilde{\bOmega}$ with a high probability.

The KKT optimality condition for 
(3.2) is
\begin{equation}\label{eqn:optimality}
    -\bOmega_D^{-1} + \bOmega\bS + \lambda\bZ = 0,
    \quad
    \exists \bZ \in \partial \|\bOmega\|_{1}
    ,
\end{equation}
where
    $\partial\|\bOmega\|_{1} = \{(z_{ij}): z_{ij} = \operatorname{sign}(\omega_{ij})~\text{if}~\omega_{ij} \neq 0,~z_{ij} \in [-1, 1]~\text{if}~\omega_{ij}=0\}$.
Thus if we let $\tilde{\bZ} = (\tilde{z}_{ij}) = \frac{1}{\lambda}(\tilde{\bOmega}_D^{-1} - \tilde{\bOmega}\bS)$ and verify that
\begin{equation}\label{eqn:pdw}
    \|\tilde{\bZ}_{S^c}\|_{\infty} = \max_{(i,j)\in S^c} |\tilde{z}_{ij}| < 1
    ,
\end{equation}
then the pair $(\tilde{\bOmega}, \tilde{\bZ})$ satisfies condition \eqref{eqn:optimality} and $\tilde{\bOmega}$ 
minimizes the objective function in (3.2).
By the uniqueness of the solution, we have $\hat{\bOmega} = \tilde{\bOmega}$.
The rest of the proof is to find a condition under which \eqref{eqn:pdw} holds. 
In order to do this, we need the following lemmas, proved in \Cref{sec:proofs:technical}.
\begin{lemma}[Control of the oracle estimation error]\label{lem:oraclecontrol}
Suppose
$$
    r := 3\kappa_{\Gamma^{\ast}}(\kappa_{\Omega^{\ast}}\|\bW\|_{\infty} + \lambda)
    \leq
    \min\left\{\frac{1}{3\gamma_1}, \frac{1}{3\gamma_1^3\kappa_{\Gamma^{\ast}}}, \frac{\kappa_{\Omega^{\ast}}}{3d} \right\}
    ,
$$
where $\bW = \bS - \bSigma^*$.
Then,
    $\|\tilde{\bOmega} - \bOmega^{\ast}\|_{\infty}%
    \leq%
    r$%
    .
\end{lemma}

\begin{lemma}[Control of remainder]\label{lem:remainder}
For $\bDelta = \tilde{\bOmega} - \bOmega^{\ast}$, let
$$
    R(\bDelta_D)
    = -\bOmega_D^{\ast -1}\left[\sum_{k=2}^{\infty}(-\bDelta_D\bOmega_D^{\ast -1})^k\right]
    .
$$
If $\|\bDelta\|_{\infty} \leq \frac{1}{3\gamma_1}$, then
    $\|R(\bDelta_D)\|_{\infty}%
    \leq%
    \frac{3}{2}\gamma_1^3\|\bDelta\|_{\infty}^2$%
    .
\end{lemma}

\begin{lemma}[Strict dual feasibility]\label{lem:strictdual}
For the primal solution $\tilde{\bOmega}$ to the oracle problem \eqref{eqn:oracle},
let $\bDelta = \tilde{\bOmega} - \bOmega^{\ast}$.
Suppose
$$
    \| R(\bDelta_D) \|_{\infty} + \kappa_{\Omega^{\ast}}\|\bW\|_{\infty} + d\|\bW\|_{\infty} \|\bDelta\|_{\infty}
    \leq
    {\alpha\lambda}/{4}
    .
$$
Then under the irrepresentability assumption (4.2), the dual optimum $\tilde{\bZ}$ for the oracle problem \eqref{eqn:oracle} satisfies \eqref{eqn:pdw}, i.e.,
$\|\bZ_{S^c}\|_{\infty} < 1$, and therefore, $\hat{\bOmega} = \tilde{\bOmega}$.
\end{lemma}

Putting all things together, we now complete the main proof. The lower bound \eqref{eqn:samplecomplexity} of the sample size implies
\begin{equation}\label{eqn:coverrorlowerbound}
    \bar\delta_f(n, p^\tau) 
    \leq
    \min\left\{ \frac{\min\left\{\frac{1}{3\gamma_1}, \frac{1}{3\gamma_1^3\kappa_{\Gamma^\ast}}, \frac{\kappa_{\Omega^\ast}}{3d}\right\}}{3\kappa_{\Gamma^\ast}\kappa_{\Omega^*}\left(1 + \frac{10}{\alpha}\right)}
        , 
        \frac{2}{%
            27\gamma_1^3 
            \kappa_{\Gamma^\ast}^2\kappa_{\Omega^\ast}\left(1 + \frac{10}{\alpha}\right)^2%
        }
        ,
        \frac{1}{v_\ast}
    \right\}.
\end{equation}
The inequality \eqref{eqn:coverrorlowerbound} implies $\bar\delta_f(n, p^\tau) \leq 1/v_\ast$. Then the event
\begin{equation}\label{eqn:noisetail}
    \{\|\bW\|_{\infty} \leq \bar\delta_f(n, p^\tau)\}
\end{equation}
occurs with a probability of at least $1-p^{-(\tau-2)}$ \citep[Lemma 8]{ravikumar2011high_a}.
In the sequel, we condition on the event \eqref{eqn:noisetail}. 

Recall that we have chosen $\lambda = \frac{10\kappa_{\Omega^\ast}}{\alpha}\bar\delta_f(n, p^\tau)$. 
Then,
\begin{equation}\label{eqn:noiselambdabound}
    \kappa_{\Omega^\ast}\|\bW\|_{\infty}
    \leq
    \frac{\alpha}{10}\lambda
    ,
\end{equation}
and
    $r = 3\kappa_{\Gamma^\ast}(\kappa_{\Omega^\ast}\|\bW\|_{\infty} + \lambda)%
    \leq 3\kappa_{\Gamma^\ast}\left({\alpha}/{10} + 1\right)\lambda%
    =%
    3\kappa_{\Gamma^\ast}\kappa_{\Omega^\ast}\left(1 + {10}/{\alpha}\right)\bar\delta_f(n, p^\tau)$.
Now \Cref{lem:oraclecontrol} combined with the inequality \eqref{eqn:coverrorlowerbound}  yields
\begin{equation}\label{eqn:deltabound}
    \|\bDelta\|_{\infty} \leq r 
    \leq 
    3\kappa_{\Gamma^\ast}\kappa_{\Omega^\ast}\left(1 + {10}/{\alpha} \right)\bar\delta_f(n, p^\tau)
    \leq
    \min\left\{\frac{1}{3\gamma_1}, \frac{1}{3\gamma_1^3\kappa_{\Gamma^\ast}}, \frac{\kappa_{\Omega^\ast}}{3d}\right\}
    .
\end{equation}
It then follows from \Cref{lem:remainder} that 
\begin{equation}\label{eqn:Rdeltabound}
\begin{split}
    \|R(\bDelta_D)\|_{\infty}
    & \leq
    \frac{3}{2}\gamma_1^3\|\bDelta\|_{\infty}^2 \leq \frac{3}{2}\gamma_1^3 r^2
    \\
    & \leq 
    \frac{3}{2}\gamma_1^3 
    \cdot 9\kappa_{\Gamma^\ast}^2\kappa_{\Omega^\ast}^2\left(1 + {10}/{\alpha} \right)^2\bar\delta_f^2(n, p^\tau)
    \\
    &=
    \frac{27}{2}\gamma_1^3 
    \kappa_{\Gamma^\ast}^2\kappa_{\Omega^\ast}^2\left(1 + {10}/{\alpha} \right)^2\bar\delta_f(n, p^\tau) 
    \frac{\alpha}{10\kappa_{\Omega^\ast}}\lambda
    \\
    &\leq
    \frac{\alpha}{10}\lambda,
\end{split}
\end{equation}
since from \eqref{eqn:coverrorlowerbound}, $n$ satisfies
    $\frac{27}{2}\gamma_1^3%
    \kappa_{\Gamma^\ast}^2\kappa_{\Omega^\ast}^2\left(1 + {10}/{\alpha}\right)^2\bar\delta_f(n, p^\tau)%
    \leq%
    \kappa_{\Omega^\ast}$%
    .
Then, from \eqref{eqn:noiselambdabound},
\begin{equation}\label{eqn:ddeltaWbound}
    d\|\bDelta\|_{\infty}\|\bW\|_{\infty}
    \leq
    dr\|\bW\|_{\infty}
    \leq
    d\frac{\kappa_{\Omega^\ast}}{3d}\|\bW\|_{\infty}
    \leq
    \frac{1}{3}\frac{\alpha}{10}\lambda
    =
    \frac{\alpha}{30}\lambda.
\end{equation}
Combining \eqref{eqn:noiselambdabound}, \eqref{eqn:Rdeltabound}, and \eqref{eqn:ddeltaWbound},
$$
    \|R(\bDelta_D)\|_{\infty} + \kappa_{\Omega^\ast}\|\bW\|_{\infty} + d\|\bDelta\|_{\infty}\|\bW\|_{\infty}
    \leq
    \frac{\alpha}{10}\lambda
    +
    \frac{\alpha}{10}\lambda
    +
    \frac{\alpha}{30}\lambda
    <
    \frac{\alpha}{4}\lambda,
$$
and the condition for \Cref{lem:strictdual} is satisfied.
Consequently, $\hat{\bOmega} = \tilde{\bOmega}$ and $\bDelta = \hat{\bOmega} - \bOmega^\ast$. Then, conclusion \eqref{eqn:Linftybound} holds by \eqref{eqn:deltabound}. That
$S(\hat{\bOmega}) \subset S$
also follows since $\hat{\bOmega}_{S^c} = \tilde{\bOmega}_{S^c} = \bOmega^\ast_{S^c} = 0$. 
Furthermore, if 
$$
    |\omega^\ast_{ij}| > 3\kappa_{\Gamma^\ast}\kappa_{\Omega^\ast}\left(1 + {10}/{\alpha} \right)\bar\delta_f(n, p^\tau),
$$ 
then $\hat{\omega}_{ij} \neq 0$ since otherwise
$|\hat{\omega}_{ij} - \omega^\ast_{ij}| >  3\kappa_{\Gamma^\ast}\kappa_{\Omega^\ast}\left(1 + {10}/{\alpha} \right)\bar\delta_f(n, p^\tau)$, contradicting \eqref{eqn:Linftybound}. 

Finally, note that
    \begin{align*}
        \| \hat{\bOmega}_D \hat{\bOmega} - \bOmega_D^*\bOmega^*\|_{\infty}
        &\leq 
        \vvvert \hat{\bOmega}_D \vvvert_\infty \|\hat{\bOmega} - \bOmega^\ast\|_{\infty} + \|\hat{\bOmega}_D - \bOmega_D^\ast\|_{\infty} \vvvert \bOmega^* \vvvert_{\infty}
        \\
        &\leq
        (\vvvert \hat{\bOmega}_D \vvvert_{\infty} + \kappa_{\Omega^*})
        \|\hat{\bOmega} - \bOmega^\ast\|_{\infty}
        .
    \end{align*}
Under the conditions of \Cref{thm:consistency}, conditioned the event \eqref{eqn:noisetail}, we have $\hat{\bOmega} = \tilde{\bOmega}$. Then,
$$
    \vvvert \hat{\bOmega}_D \vvvert_{\infty} 
    = \vvvert \bDelta_D + \bOmega_D^* \vvvert_{\infty} \leq \vvvert \bOmega_D^* \vvvert_{\infty} \vvvert \bI_p + \bDelta_D \bOmega_D^{*-1} \vvvert_{\infty}
    \leq \kappa_{\Omega^*} \Vert \bI_p + \bDelta_D \bOmega_D^{*-1} \Vert_{\infty}
    ,
$$ 
because $\bI_p + \bDelta_D \bOmega_D^{*-1}$ is diagonal.
Recall that on the conditioned event, $\Vert \bDelta_D\Vert_{\infty} \leq \Vert \bDelta \Vert_{\infty} \leq r \leq 1/(3\gamma_1)$. Therefore, $\Vert \bI + \bDelta_D \bOmega_D^{*-1} \Vert_{\infty} \leq 4/3$ and we conclude that
$$
        \| \hat{\bTheta} - \bTheta^*\|_{\infty} \leq (7/3)\kappa_{\bOmega^*}\|\hat{\bOmega} - \bOmega^\ast\|_{\infty}.
$$
\end{proof}

\begin{proof}[Proof of Theorem 4.2]
    The conclusion of Theorem 4.2 follows by noting that each normalized coordinate variable $X_i/\sqrt{\Sigma^\ast_{ii}}$ is sub-Gaussian with parameter $\sigma$ and satisfies the tail condition $\mathcal{T}(f, v_\ast)$ with $v_\ast = [(\max_{i\in[p]} \Sigma^\ast_{ii})8(1+4\sigma^2)]^{-1}$ and
\begin{align*}
    f(n, \delta) &= \frac{1}{4}\exp(c_\ast n\delta^2),
    \quad
    c_\ast = [128(1+4\sigma^2)^2 \max_{i\in[p]}(\Sigma^\ast_{ii})^2]^{-1}
    ,
    \\
    \bar\delta_f(n, r) &= \sqrt{\frac{\log(4r)}{c_\ast n}},
    \qquad
    \bar{n}_f(\delta, r) = \frac{\log(4r)}{c_\ast \delta^2}
\end{align*}
\citep[\S2.3]{ravikumar2011high_a} and applying \Cref{thm:consistency}.
\end{proof}

\subsection{Proof of Theorem 4.3 and Corollary 4.4}
Again we prove a more general result:
\begin{theorem}[Model selection and sign consistency]\label{thm:sign_consistency_gen}
Under the conditions of \Cref{thm:consistency}, suppose that the sample size $n$ is such that
$$
    n > \bar{n}_f\left(\min\left\{C^{-1}\omega, \delta	\right\}, p^\tau \right)
$$
for some $0 < \omega < \omega_{\min}$. Then, the event
$$
    \{\operatorname{sign}(\omega_{ij}^*) = \operatorname{sign}(\hat{\omega}_{ij}) \text{ for all } (i,j) \in S\}
$$
occurs with a probability no smaller than $1-p^{-(\tau - 2)}$.
\end{theorem}
\begin{proof}
Following the procedures of proving \Cref{thm:consistency}, we can show that with probability not smaller than $1-p^{-(\tau - 2)}$, $\hat{\bOmega}_{S^c} = \tilde{\bOmega}_{S^c} = 0$, which indicates that the ACCORD solution is the same as the oracle solution. Thus, all edges outside the true edge set $S$ are excluded. Then, using \Cref{lem:oraclecontrol}, we have
\begin{align*}
    \Vert\hat{\bOmega}_S - \bOmega_S^*\Vert_\infty \leq C\bar{\delta}_f(n, p^\tau) \leq \omega < \omega_{\min}
\end{align*}
since $n > \bar{n}_f\left(\min\left\{C^{-1}\omega, \delta	\right\}, p^\tau \right)$ implies $\bar\delta_f(n, p^\tau) \leq C^{-1}\omega$. Then, we can conclude that for every $(i,j) \in S$, $\hat{\omega}_{ij}$ has the same sign as $\omega_{ij}^*$, because otherwise, $\omega_{\min} > |\hat{\omega}_{ij} - \omega_{ij}^*| = |\hat{\omega}_{ij}| + |\omega_{ij}^*| \geq |\omega_{ij}^*|$, which violates the definition of $\omega_{\min}$.
\end{proof}
\begin{proof}[Proof of Theorem 4.3]
    The conclusion of Theorem 4.3 follows from \Cref{thm:sign_consistency_gen} by specifying $f(n, \delta)$, $v_\ast$, $c_\ast$, $\bar\delta_f(n, r)$, and $\bar{n}_f(\delta, r)$ as in the proof of Theorem 4.2 and setting $\omega=\omega_{\min}/2$.
\end{proof}
\begin{proof}[Proof of Corollary 4.4]
Denote the estimator $\hat{\bOmega}$ obtained under penalty coefficient $\lambda$ by $\hat{\bOmega}^{\lambda}$ to emphasize the role of the coefficient.
Recall that in the proof of \Cref{thm:sign_consistency_gen}, 
$\lambda = \frac{10\kappa_{\Omega^\ast}}{\alpha}\bar\delta_f(n, p^\tau)$,
and
    both the error bound $\Vert\hat{\bOmega}_S^{\lambda} - \bOmega_S^*\Vert_\infty \leq C\bar{\delta}_f(n, p^\tau)$ and the perfect support recovery of $\hat{\bOmega}^{\lambda}$
    are derived on the event that $\kappa_{\Omega^\ast}\|\bW\|_{\infty}
    \leq \frac{\alpha}{10}\lambda$,
    which occurs with a probability at least $1-p^{-(\tau-2)}$ under the conditions of Theorem 4.3.
In the sequel, we condition on this event.
Also recall that $r = 3\kappa_{\Gamma^*}(\kappa_{\Omega^*}\|\bW\|_{\infty} + \lambda )$
and let $r_\phi = 3\kappa_{\Gamma^*}(\kappa_{\Omega^*}\|\bW\|_{\infty} + \phi\lambda)$.
Then, 
by \Cref{lem:oraclecontrol}, 
it holds that
$$
        \Vert \tilde{\bOmega}^{\phi\lambda} - \bOmega^*\Vert_\infty \leq 
        r_{\phi} 
        \leq
        r
        \leq
        3\kappa_{\Gamma^{\ast}}(1 + \alpha/10) \lambda =  3\kappa_{\Gamma^{\ast}}\kappa_{\Omega^*}(1 + 10/\alpha) \bar\delta_f(n,p^\tau)
$$
for the oracle estimator $\tilde{\bOmega}^{\phi\lambda}$ in \eqref{eqn:oracle} with penalty $\phi\lambda\; (0 \leq \phi < 1)$. 
    Since the bias-corrected estimator $\breve{\bOmega}=\breve{\bOmega}^{\phi\lambda}$ is equal to $\tilde{\bOmega}^{\phi\lambda}$ when $\hat{\bOmega}^{\lambda}$ recovers the support perfectly, we complete the proof.
\end{proof}
\subsection{Results for polynomial tails}\label{sec:proofs:polynomial}
It can be shown that a zero-mean random vector $X\in\mathbb{R}^p$ with covariance matrix $\bSigma^*$ where each coordinate variable $X_i$ has a bounded $4m$-th moment satisfies the tail condition with $v_\ast=0$ and has a polynomial-type tail. Specifically, if
$$
    \mathbb{E}\left[\left({X_i}/{\sqrt{\Sigma^\ast_{ii}}}\right)^{4m}\right]
    \leq
    K_m < \infty
$$
for some $K_m > 0$, then
$$
    f(n,\delta) = c_\ast n^m \delta^{2m},
    \quad
    c_\ast = [m^{2m+1}2^{2m}(\max_i\Sigma^\ast_{ii})^{2m}(K_m+1)]^{-1}
$$
and
$$
    \bar{\delta}_f(n, r) = \frac{(r/c_\ast)^{1/2m}}{\sqrt{n}}
    ,
    \quad
    \bar{n}_f(\delta, r) = \frac{(r/c_\ast)^{1/m}}{\delta^2}
$$
\citep[\S2.3]{ravikumar2011high_a}.
In this case, \Cref{thm:consistency} and \Cref{thm:sign_consistency_gen} incarnate as the following.
\begin{corollary}[Elementwise error]\label{cor:polytail}
Suppose  the data matrix $\bX \in \mathbb{R}^{n \times p}$ is composed of $n$ i.i.d. copies of zero-mean continuous random vector $X=(X_1,\dotsc, X_p)\in\mathbb{R}^p$ with covariance matrix $\bSigma^*= (\Sigma^*_{ij}) \allowbreak = {\bTheta^*}^{-1}$ and each $X_j$ satisfies $\E(X_j/\sqrt{\Sigma^*_{jj}})^{4m} \leq K_m < \infty$ for a positive integer $m$.
Let $\sigma^2 = \max_{j\in[p]}\Sigma^*_{jj}$.
If further Assumption 4.1 holds, then 
for 
$\lambda = 20m^{1+1/(2m)}(K_m + 1)^{1/(2m)}\sigma^2\kappa_{\Omega^\ast}\alpha^{-1}p^{\tau/(2m)}/\sqrt{n}$, 
$\tau > 2$,
\begin{enumerate}
    \item there holds
    $\|\hat{\bOmega} - \bOmega^\ast\|_{\infty}%
			\leq%
    6m^{1+1/(2m)}(K_m+1)^{1/(2m)}\sigma^2\kappa_{\Gamma^\ast}\kappa_{\Omega^\ast}(1 + {10}/{\alpha}) p^{\tau/2m}/\sqrt{n}%
    $%
    ;
	\item the estimated support set $\hat{S}=\{(i,j)\in [p]\times [p]: \hat{\omega}_{ij} \neq 0\}$ is contained in the true support $S$ and includes all edges $(i,j)$ with 
		$$
		|\omega^\ast_{ij}| > 
			6m^{1+1/(2m)}(K_m+1)^{1/(2m)}\sigma^2\kappa_{\Gamma^\ast}\kappa_{\Omega^\ast}(1 + {10}/{\alpha}) p^{\tau/2m}/\sqrt{n}
		,
		$$
\end{enumerate}
with a probability no smaller than $1 - p^{-(\tau-2)}$,
provided that 
$n > 4m^{2+1/m}(K_m+1)^{1/m}\sigma^4\delta^{-2}p^{\tau/m}$ where
$$
    \delta = \min\left\{ \frac{\min\left\{\frac{1}{3\gamma_1}, \frac{1}{3\gamma_1^3\kappa_{\Gamma^\ast}}, \frac{\kappa_{\Omega^\ast}}{3d}\right\}}{3\kappa_{\Gamma^\ast}\kappa_{\Omega^*}\left(1 + {10}/{\alpha}\right)}
        , 
        \frac{2}{%
            27\gamma_1^3 
            \kappa_{\Gamma^\ast}^2\kappa_{\Omega^\ast}\left(1 + {10}/{\alpha}\right)^2%
        }
    \right\} 
    .
$$
\end{corollary}
\begin{corollary}[Edge selection and sign consistency]\label{cor:sign_consistency}
Assume the same conditions as \Cref{cor:polytail}. If the sample size satisfies
$$
    n > 4\sigma^4(K_m+1)^{1/m}p^{\tau/m}/\min\{\omega_{\min}/[6\kappa_{\Gamma^\ast}\kappa_{\Omega^\ast}(1+10/\alpha)],\delta\}^2
$$
where $\sigma$ and $\delta$ are as defined in \Cref{cor:polytail},
then the event
$$
    \{\operatorname{sign}(\omega_{ij}^*) = \operatorname{sign}(\hat{\omega}_{ij}) \text{ for all } (i,j)\}
$$
occurs with a probability no smaller than $1-p^{-(\tau - 2)}$.
\end{corollary}

\subsection{Proofs of technical lemmas}\label{sec:proofs:technical}
Throughout, it is convenient to note
\begin{equation}\label{eqn:veccab}
    \vect(\bA\bB\bC) = (\bC^T\otimes \bA)\vect(\bB)
    .
\end{equation}

\begin{proof}[Proof of \Cref{lem:cont}]
    Fix the sample $(X^1, \cdots, X^n),$ and rewrite the left side of the inequality as
    \begin{align*}
        \Expect_{\epsilon}&\left[\sup_{f \in \mathcal{F}} \sum_{i=1}^n \epsilon_i(\phi_i \circ f)(X^i)\right] 
        = \Expect_{\epsilon_1, \cdots, \epsilon_{n-1}}\Expect_{\epsilon_n}\sup_{f \in \mathcal{F}} \left[ m_{n-1}(f) + \epsilon_n(\phi_n \circ f)(X^n) \right] \\
        = & \frac{1}{2}\Expect_{\epsilon_1, \cdots, \epsilon_{n-1}}\sup_{f \in \mathcal{F}} \left[ m_{n-1}(f) + (\phi_n \circ f)(X^n)\right]\\
        & +
        \frac{1}{2}\Expect_{\epsilon_1, \cdots, \epsilon_{n-1}}\sup_{f \in \mathcal{F}} \left[ m_{n-1}(f) - (\phi_n \circ f)(X^n)\right],
    \end{align*}
    where $m_{n-1}(f) = \sum_{i=1}^{n-1} \epsilon_i(\phi_i \circ f)(X^i)$.
    For fixed $\epsilon_1, \cdots, \epsilon_{n-1}$, note that there exists $f_1, f_2$ such that
    \begin{align*}
        &\frac{1}{2}\sup_{f \in \mathcal{F}} \left[ m_{n-1}(f) + (\phi_n \circ f)(X^n)\right] +
        \frac{1}{2}\sup_{f \in \mathcal{F}} \left[ m_{n-1}(f) - (\phi_n \circ f)(X^n)\right] - \delta\\
        &\leq 
        \frac{1}{2}\left[ m_{n-1}(f_1) + (\phi_n \circ f_1)(X^n)\right] +
        \frac{1}{2}\left[ m_{n-1}(f_2) - (\phi_n \circ f_2)(X^n)\right],
    \end{align*}
    for some arbitrary $\delta > 0.$ Let $s = \text{sign}(f_1(X^n) - f_2(X^n))$. Then, we have
    \begin{align*}
        &\Expect_{\epsilon_n}\sup_{f \in \mathcal{F}} \left[ m_{n-1}(f) + \epsilon_n(\phi_n \circ f)(X^n) \right] -\delta \\
        &\leq \frac{1}{2}\left[m_{n-1}(f_1) + m_{n-1}(f_2) + Ls(f_1(X^n) - f_2(X^n)) \right]\quad\text{($L$-Lipschitz)}\\
        &\leq \frac{1}{2}\sup_{f\in\mathcal{F}}\left[m_{n-1}(f) + Lsf(X^n)\right] + \frac{1}{2}\sup_{f\in\mathcal{F}}\left[m_{n-1}(f) - Lsf(X^n)\right] \\
        &= \Expect_{\epsilon_n}\sup_{f \in \mathcal{F}}\left[m_{n-1}(f) + L \epsilon_n f(X^n)\right].
    \end{align*}
    Since the inequality holds for arbitrary $\delta > 0,$ we have
    $$
        \Expect_{\epsilon_n}\sup_{f \in \mathcal{F}} \left[ m_{n-1}(f) + \epsilon_n(\phi_n \circ f)(X^n) \right]
        \leq 
        \Expect_{\epsilon_n}\sup_{f \in \mathcal{F}}\left[m_{n-1}(f) + L \epsilon_n f(X^n)\right],
    $$
    and repeating this step proves the lemma by induction.
\end{proof}

\begin{proof}[Proof of \Cref{lem:inf-mom}]
    By Jensen's inequality, for any $t > 0,$
    \begin{align*}
        &\exp(t\Expect_X[\max_i|X_i|]) \leq \Expect_X \max_i\exp(t|X_i|) \\
        &\leq \sum_{i=1}^p \{\Expect_X\exp(tX_i) + \Expect_X\exp(-tX_i)\} \leq 2p\exp(t^2\sigma^2/2),
    \end{align*}
    Therefore, $\Expect_X[\max_i|X_i|] \leq \frac{\log 2p}{t} + \frac{t\sigma^2}{2},$ and setting $t = \sqrt{2\log 2p}/\sigma$ proves the inequality.
\end{proof} 

\begin{proof}[Proof of \Cref{lem:oraclecontrol}]
The optimality conditions for the oracle problem \eqref{eqn:oracle} are
\begin{subequations}\label{eqn:oracleoptimality}
    \begin{equation}
    [-\bOmega_D^{-1} + \bOmega\bS + \lambda\bZ]_S = 0,
    \quad
    \exists \bZ \in \partial \|\bOmega\|_{1}
    ,
    \label{eqn:oracleoptimality:subgrad}
    \end{equation}
    \begin{equation}
    \bOmega_{S^c} = 0
    \label{eqn:oracleoptimality:support}
    \end{equation}
\end{subequations}
For any $\bOmega \in \mathbb{R}^{p\times p}$ satisfying \eqref{eqn:oracleoptimality:support},
write $\bOmega = \bOmega^{\ast} + \bDelta$. 
Then $\bOmega = \bOmega_S = \Omega^{\ast}_S + \bDelta_S$, $\bOmega_D = \bOmega^{\ast}_D + \bDelta_D$,
and $\bDelta_{S^c} = 0$, $\bDelta = \bDelta_S$.
Then,
\begin{align*}
    &[-\bOmega_D^{-1} + \bOmega\bS + \lambda\bZ]_S = -[\bOmega_D^{-1}]_S + [\bOmega\bS]_S + \lambda\bZ_S
    \\
    &= -[(\bOmega^{\ast}_D + \bDelta_D)^{-1}]_S + [(\bOmega^{\ast}+\bDelta)(\bSigma^{\ast}+\bW)]_S + \lambda\bZ_S
    \\
    &= -[(\bOmega^{\ast}_D + \bDelta_D)^{-1}]_S + [\bOmega^{\ast}\bTheta^{\ast -1} + \bDelta\bSigma^{\ast} + \bOmega^{\ast}\bW + \bDelta\bW]_S + \lambda\bZ_S
    \\
    &= -[(\bOmega^{\ast}_D + \bDelta_D)^{-1}]_S + [\bOmega^{\ast -1}_D + \bDelta\bSigma^{\ast} + \bOmega^{\ast}\bW + \bDelta\bW]_S + \lambda\bZ_S
    \\
    &= [-(\bOmega^{\ast}_D + \bDelta_D)^{-1} + \bOmega^{\ast -1}_D ]_S + [\bDelta\bSigma^{\ast}]_S + [\bOmega^{\ast}\bW]_S + [\bDelta\bW]_S + \lambda\bZ_S
    \\
    &= [-(\bOmega^{\ast}_D + \bDelta_D)^{-1} + \bOmega^{\ast -1}_D ]_S + [\bDelta_S\bSigma^{\ast}]_S + [\bOmega^{\ast}\bW]_S + [\bDelta_S\bW]_S + \lambda\bZ_S
    \\
    &=: G(\bDelta_S)
    .
\end{align*}
But
\begin{align*}
    (\bOmega^{\ast}_D + \bDelta_D)^{-1} 
    &=
    [(\bI_p + \bDelta_D\bOmega_D^{\ast -1})\bOmega_D^{\ast}]^{-1}
    \\
    &=
    \bOmega_D^{\ast -1}(\bI + \bDelta_D\bOmega_D^{\ast -1})^{-1}
    \\
    &=
    \bOmega_D^{\ast -1}[\sum_{k=0}^{\infty}(-\bDelta_D\bOmega_D^{\ast -1})^k]
    \\
    &=
    \bOmega_D^{\ast -1}[\bI - \bDelta_D\bOmega_D^{\ast -1} + \sum_{k=2}^{\infty}(-\bDelta_D\bOmega_D^{\ast -1})^k]
    \\
    &=
    \bOmega_D^{\ast -1} - \bOmega_D^{\ast -1}\bDelta_D\bOmega^{\ast -1} + \underbrace{\bOmega_D^{\ast -1}[\sum_{k=2}^{\infty}(-\bDelta_D\bOmega_D^{\ast -1})^k]}_{=-R(\bDelta_D)} < \infty
    ,
\end{align*}
provided that the series converges.

Observing that $\bDelta_D = \sum_{i=1}^p e_i e_i^T\bDelta e_i e_i^T$, it follows that
\begin{align*}
    G(\bDelta_S)
    =& [\bOmega_D^{\ast -1}\bDelta_D\bOmega_D^{\ast -1} + \bDelta\bSigma^{\ast}]_S + R(\bDelta_D) +  [\bOmega^{\ast}\bW]_S + [\bDelta_S\bW]_S + \lambda\bZ_S
    \\
    =& [\sum_{i=1}^p\bOmega_D^{\ast -1} e_i e_i^T\bDelta e_i e_i^T\bOmega_D^{\ast -1} + \bDelta_S\bSigma^{\ast}]_S + R(\bDelta_D) +  [\bOmega^{\ast}\bW]_S \\
    &+ [\bDelta_S\bW]_S + \lambda\bZ_S
    .
\end{align*}
Consider a vectorization of $G(\bDelta_S)$:
\begin{equation}\label{eqn:vectorization}
    \begin{split}
    g(\bDelta_S)
    =& [\vect G(\bDelta_S)]_S
    \\
    =& \bGamma^{\ast}_{SS}
        \vect(\bDelta_S)_S
        + \vect R(\bDelta_D)_S +  \vect(\bOmega^{\ast}\bW)_S + \vect(\bDelta_S\bW)_S \\
        &+ \lambda\vect(\bZ_S)_S
    .
    \end{split}
\end{equation}
We recall (4.1), \eqref{eqn:veccab}, and
$$
    \sum_{i=1}^p\bOmega_D^{\ast -1} e_i e_i^T\otimes \bOmega_D^{\ast -1} e_i e_i^T 
    =
    (\bOmega_D^{\ast -1} \otimes \bOmega_D^{\ast -1})\bUpsilon
    .
$$
If we let $F:\mathbb{R}^{|S|}\to\mathbb{R}^{|S|}$ be
$$
    F(\bDelta_S) = \vect(\bDelta_S)_S - \bGamma^{\ast -1}_{SS}g(\bDelta_S) 
    ,
$$
then $F(\bDelta_S) = \vect(\bDelta_S)_S$ if and only if either $G(\bDelta_S) = 0$ or \eqref{eqn:oracleoptimality:subgrad} is satisfied.

Thus, if a fixed point of the map $F$ exists, then it yields a solution to \eqref{eqn:oracleoptimality}, which must be unique.
It follows from (4.1) that
$$
    F(\bDelta_S)
    = \bGamma_{SS}^{\ast -1}[
    - \vect R(\bDelta_D) - \vect(\bOmega^{\ast}\bW)_S - \vect(\bDelta_S\bW)_S - \lambda \vect(\bZ_S)_S]
    ,
$$
and
\begin{align*}
    \|F(\bDelta_S)\|_{\infty}
    &\leq
    \kappa_{\Gamma^{\ast}}(\|R(\bDelta_D)\|_{\infty}
    + \vvvert \bOmega^{\ast} \vvvert_{\infty} \|\bW\|_{\infty}
    + \| \bW \bDelta_S\|_{\infty}
    + \lambda )
    .
    \\
    &=
    \kappa_{\Gamma^{\ast}}(\|R(\bDelta_D)\|_{\infty}
    + \vvvert \bDelta_S \vvvert_{\infty} \|\bW\|_{\infty}
    + \kappa_{\Omega^{\ast}} \|\bW\|_{\infty}
    + \lambda )
    \\
    &\leq
    \kappa_{\Gamma^{\ast}}(\|R(\bDelta_D)\|_{\infty}
    + d \|\bDelta_S\|_{\infty}\|\bW\|_{\infty} 
    + \kappa_{\Omega^{\ast}} \|\bW\|_{\infty}
    + \lambda )
    .
\end{align*}
This is because $\bDelta=\bDelta_S$ has at most $d$ non-zeroes per row or column, and thus $\vvvert \bDelta_S \vvvert_{\infty} \leq d \| \bDelta_S \|_{\infty}$.
For any $\bDelta_S$ such that $\|\bDelta_S\|_{\infty} \leq r$ for some $r \in (0, 1/\gamma_1)$ (recall $\gamma_1 = 1/(\min_{i=1,\dotsc,p} \omega^{\ast}_{ii}) > 0$), we see
\begin{align*}
    \|R(\bDelta_D)\|_{\infty}
    &= \left\|\bOmega^{\ast -1}_D \sum_{k=2}^{\infty}(-\bDelta_D\bOmega^{\ast -1}_D)^k\right\|_{\infty}
    \\
    &\leq \sum_{k=2}^{\infty}\|\bOmega^{\ast -1}_D(\bDelta_D\bOmega^{\ast -1}_D)^k\|_{\infty}
    = \sum_{k=2}^{\infty}\max_{i=1,\dotsc,p}|\omega^{\ast -1}_{ii}(\delta_{ii}/\omega^{\ast}_{ii})^k|
    \\
    &\leq \sum_{k=2}^{\infty} \gamma_1 (\gamma_1 r)^k
    = \gamma_1^3 \frac{r^2}{1 - \gamma_1 r}
    < \infty .
\end{align*}
Thus, the series converges for the range of $r$ claimed.

If $r \leq \min\left\{\frac{1}{3\gamma_1}, \frac{1}{3\gamma_1^3\kappa_{\Gamma^{\ast}}}\right\}$, then
\begin{equation}\label{eqn:Rbound}
    \begin{split}
    \kappa_{\Gamma^{\ast}}
    \|R(\bDelta_D)\|_{\infty}
    \leq
    \kappa_{\Gamma^{\ast}}
    \gamma_1^3 \frac{r^2}{1-1/3} 
    = 
    \kappa_{\Gamma^{\ast}}
    \frac{3}{2}\gamma_1^3 r^2
    \leq
    \frac{3\gamma_1^3\kappa_{\Gamma^{\ast}}}{2}\frac{1}{3\gamma_1^3\kappa_{\Gamma^{\ast}}}r 
    = \frac{r}{2}
    .
    \end{split}
\end{equation}
Now equate $r = 3\kappa_{\Gamma^{\ast}}(\kappa_{\Omega^{\ast}}\|\bW\|_{\infty} + \lambda)$.
If further $r \leq \frac{\kappa_{\Omega^{\ast}}}{3d}$, then
\begin{align*}
    &\kappa_{\Gamma^{\ast}} d \|\bW\|_{\infty}\|\bDelta_S\|_{\infty}
    \leq
    \kappa_{\Gamma^{\ast}} d \|\bW\|_{\infty}r
    \leq
    \kappa_{\Gamma^{\ast}} d \|\bW\|_{\infty} \frac{\kappa_{\Omega^{\ast}}}{3d}
    \\
    \quad &=
    \frac{1}{3} \kappa_{\Gamma^{\ast}}\kappa_{\Omega^{\ast}}\|\bW\|_{\infty}
    \leq
    \frac{1}{3} \kappa_{\Gamma^{\ast}}(\kappa_{\Omega^{\ast}}\|\bW\|_{\infty} + \lambda)
    =
    \frac{r}{9}
    .
\end{align*}
Thus from \eqref{eqn:Rbound},
$$
    \|F(\bDelta_S)\|_{\infty} 
    \leq 
    \frac{r}{2} + \frac{r}{9} + \frac{r}{3}
    \leq 
    r
$$
and $F$ maps $B(r) = \{\Delta_S: \|\Delta_S\|_{\infty} \leq r\}$ to itself.
Then, by Brouwer's fixed point theorem, $F$ has a fixed point $\bar{\bDelta}_S$ in $B(r)$.
It follows that $\bar{\bOmega}_S = \bOmega^{\ast}_S + \bar{\bDelta}_S$ satisfies \eqref{eqn:oracleoptimality}.
By the uniqueness of the solution, we have $\tilde{\bOmega} = \bar{\bOmega}_S$, and
$$
    \|\bar{\bDelta}_S\|_{\infty}
    =
    \|\tilde{\bOmega} - \bOmega^{\ast}\|_{\infty}
    \leq
    r 
    := 3\kappa_{\Gamma^{\ast}}(\kappa_{\Omega^{\ast}}\|\bW\|_{\infty} + \lambda)
    ,
$$
provided that
$$
    r 
    \leq
    \min\left\{\frac{1}{3\gamma_1}, \frac{1}{3\gamma_1^3\kappa_{\Gamma^{\ast}}}, \frac{\kappa_{\Omega^{\ast}}}{3d} \right\}
    .
$$
\end{proof}

\begin{proof}[Proof of \Cref{lem:remainder}]
    It follows immediately from the derivation of inequality \eqref{eqn:Rbound} in the proof of \Cref{lem:oraclecontrol}.
\end{proof}

\begin{proof}[Proof of \Cref{lem:strictdual}]
From \eqref{eqn:oracleoptimality} and \eqref{eqn:vectorization},
\begin{align*}
    \vect(\bDelta)_S
    =&
    -\bGamma^{\ast -1}_{SS}[
        \vect R(\bDelta_D)_S +  \vect(\bOmega^{\ast}\bW)_S + \vect(\bDelta\bW)_S + \lambda\vect(\tilde{\bZ})_S],
    \\
    0
    =&
    \bGamma^{\ast}_{S^c S} \vect(\bDelta)_S
    + [\vect R(\bDelta_D) +  \vect(\bOmega^{\ast}\bW) + \vect(\bDelta_S\bW) \\
    &+ \lambda\vect(\tilde{\bZ})]_{S^c}.
\end{align*}
Therefore,
\begin{align*}
    \vect(\tilde{\bZ}_{S^c})
    =& \frac{1}{\lambda}\bGamma^{\ast}_{S^cS}\bGamma^{\ast -1}_{SS}[\vect R(\bDelta_D)_S +  \vect(\bOmega^{\ast}\bW)_S + \vect(\bDelta\bW)_S 
    \\
    &+ \lambda\vect(\tilde{\bZ})_S]
    - \frac{1}{\lambda}[\vect R(\bDelta_D)_{S^c} 
    + \vect(\bOmega^{\ast}\bW)_{S^c}
    + \vect(\bDelta_S\bW)_{S^c}]
    \\
    =& \frac{1}{\lambda}\bGamma^{\ast}_{S^cS}\bGamma^{\ast -1}_{SS}[\vect R(\bDelta_D)_S +  \vect(\bOmega^{\ast}\bW)_S + \vect(\bDelta\bW)_S]
    \\
    &+ \bGamma^{\ast}_{S^cS}\bGamma^{\ast -1}\vect(\tilde{\bZ})_S
    - \frac{1}{\lambda}[\vect R(\bDelta_D)_{S^c}
    + \vect(\bOmega^{\ast}\bW)_{S^c}\\
    &+ \vect(\bDelta_S\bW)_{S^c}
    ]
    ,
\end{align*}
which entails
\begin{align*}
    \|\tilde{\bZ}_{S^c}\|_{\infty}
    & \leq
    \frac{1}{\lambda}\vvvert\bGamma^{\ast}_{S^cS}\bGamma^{\ast -1}_{SS}\vvvert_{\infty} [\| R(\bDelta_D)_S \|_{\infty} +  \|(\bOmega^{\ast}\bW)_S\|_{\infty} + \|(\bDelta\bW)_S\|_{\infty} ]
    \\
    &\quad 
    + \vvvert \bGamma^{\ast}_{S^cS}\bGamma^{\ast -1}_{SS}\vvvert_{\infty} \|\vect(\tilde{\bZ})_S\|_{\infty}
    + \frac{1}{\lambda}(\|R(\bDelta_D)_{S^c}\|_{\infty} + \|(\bOmega^{\ast}\bW)_{S^c}\|_{\infty} 
    \\
    &\quad + \|(\bDelta_S\bW)_{S^c}\|_{\infty})
    \\
    &\leq
    \frac{1}{\lambda}(1-\alpha)[\| R(\bDelta_D)_S \|_{\infty} +  \|(\bOmega^{\ast}\bW)_S\|_{\infty} + \|(\bDelta\bW)_S\|_{\infty} ]
    \\
    &\quad
    + 1-\alpha
    + \frac{1}{\lambda}(\|R(\bDelta_D)_{S^c}\|_{\infty} + \|(\bOmega^{\ast}\bW)_{S^c}\|_{\infty} + \|(\bDelta_S\bW)_{S^c}\|_{\infty})
    \\
    &\leq
    \frac{2-\alpha}{\lambda}[\| R(\bDelta_D) \|_{\infty} +  \|\bOmega^{\ast}\bW\|_{\infty} + \|\bDelta\bW\|_{\infty} ]
    +
    1 - \alpha
    \\
    &\leq
    \frac{2-\alpha}{\lambda}[\| R(\bDelta_D) \|_{\infty} +  \vvvert\bOmega^{\ast}\vvvert_{\infty}\|\bW\|_{\infty} + \vvvert\bDelta \vvvert_{\infty} \|\bW\|_{\infty} ]
    +
    1 - \alpha
    \\
    &\leq
    \frac{2-\alpha}{\lambda}[\| R(\bDelta_D) \|_{\infty} +  \kappa_{\Omega^{\ast}}\|\bW\|_{\infty} + d\|\bDelta \|_{\infty} \|\bW\|_{\infty} ]
    +
    1 - \alpha
    \\
    &\leq
    \frac{2-\alpha}{\lambda} \frac{\alpha\lambda}{4} + 1 - \alpha
    \leq
    \frac{\alpha}{2} + 1 - \alpha
    < 1
    .
\end{align*}
The second inequality is due to the irrepresentability assumption (4.2).
The fifth inequality holds since $\bDelta$ has at most $d$ non-zeroes per row or column, and thus $\vvvert \bDelta \vvvert_{\infty} \leq d \| \bDelta \|_{\infty}$.
\end{proof}

\section{Additional Figures from Numerical Experiments}\label{sec:debiasing}
\subsection{Edge detection}

\begin{figure}[h]
    \centering
    \begin{subfigure}[t]{0.95\textwidth}
        \centering
        \includegraphics[width=\linewidth]{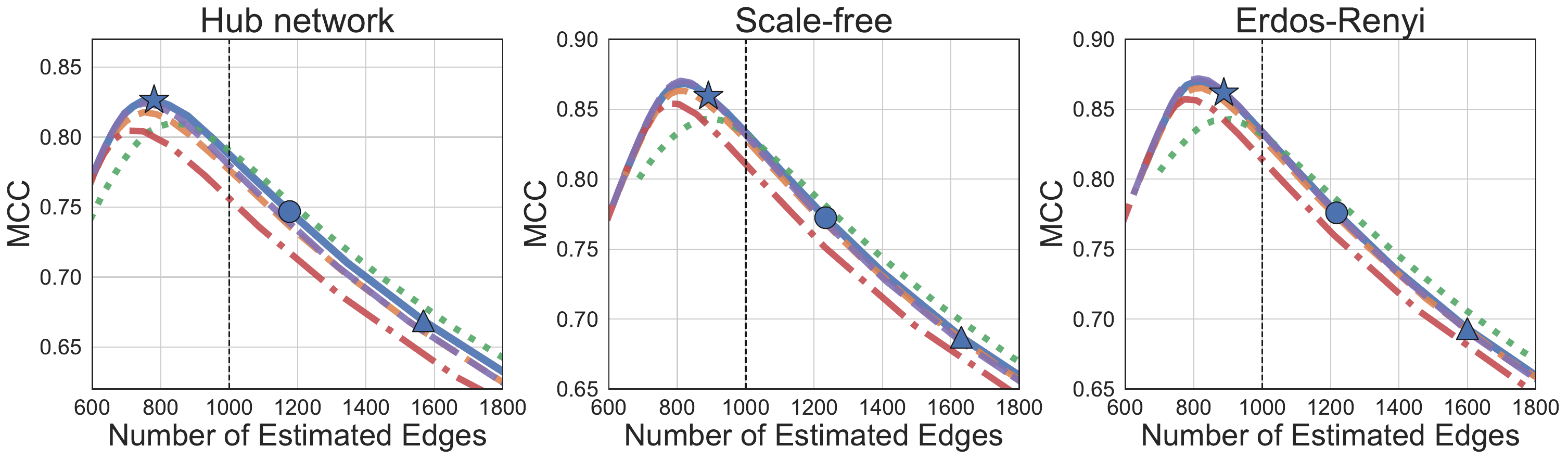}
        \caption{Matthews correlation coefficient curve by estimated edges.}
    \end{subfigure}
    \\ \vspace{5pt}
    \begin{subfigure}[t]{0.95\textwidth}
        \centering
        \includegraphics[width=\linewidth]{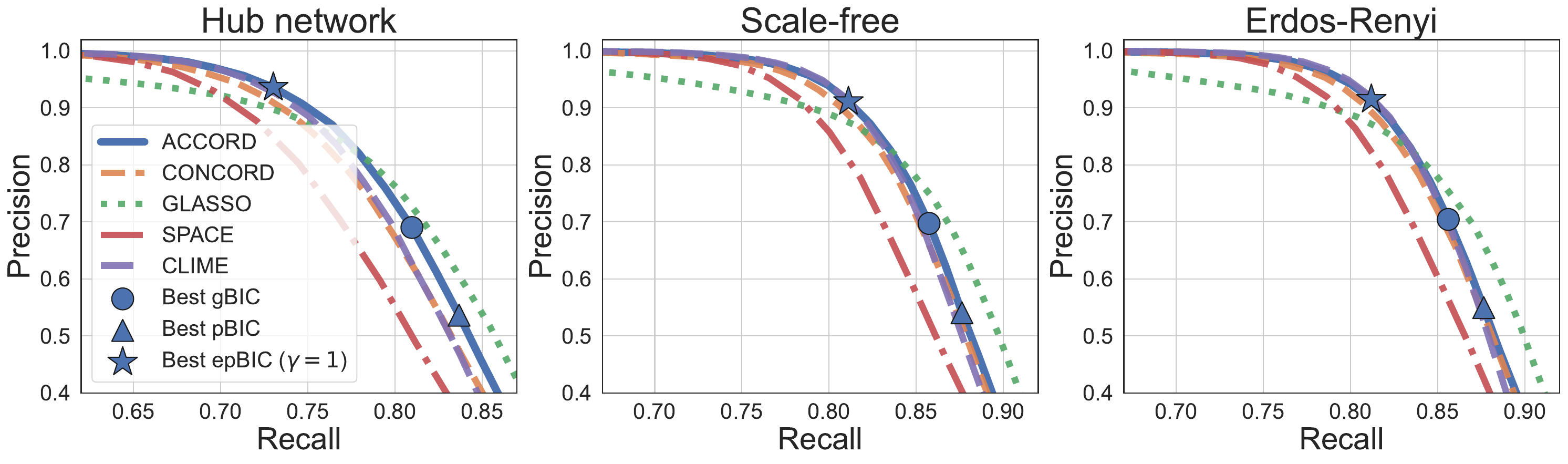}
        \caption{Precision-recall curve.}
    \end{subfigure}
    \caption{Edge detection performance comparison with various regularization parameter.}
    \label{fig:edge-detection}
\end{figure}

\Cref{fig:edge-detection} shows the Matthews correlation coefficient (MCC) and precision-recall curves generated by varying regularization parameter $\lambda$, where each point represents averaged value of 50 replications in same $\lambda$. 
Overall, the mean edge detection performance of ACCORD was slightly better compared to CONCORD, CLIME and SPACE; this gap was bigger in Hub Network and Scale-free graph, which are more complex structures.
Compared with graphical LASSO, ACCORD showed better MCC when the estimates are sparsely selected. In the precision-recall curves, ACCORD showed the best AUC in Hub Network and Scale-free graph, as shown in Table 1.
Also, we marked the most frequently selected model among all $\lambda$ grid in \Cref{fig:edge-detection} for different criteria. The models selected by the plain pseudo-BIC or Gaussian BIC were denser than the models selected by extended pseudo-BIC. The sparser estimates selected with extended pseudo-BIC showed better results in terms of MCC.
Hence, these results shows practical merit of using the extened pseudo-BIC in terms of tuning parameter selection; the selected models retain estimates with far less false positive edges in the expense of few true positive edges.

\subsection{Parameter estimation}

\begin{figure}[tp]
\centering
\subfloat[Hub Network]{%
  \includegraphics[width=0.95\textwidth]{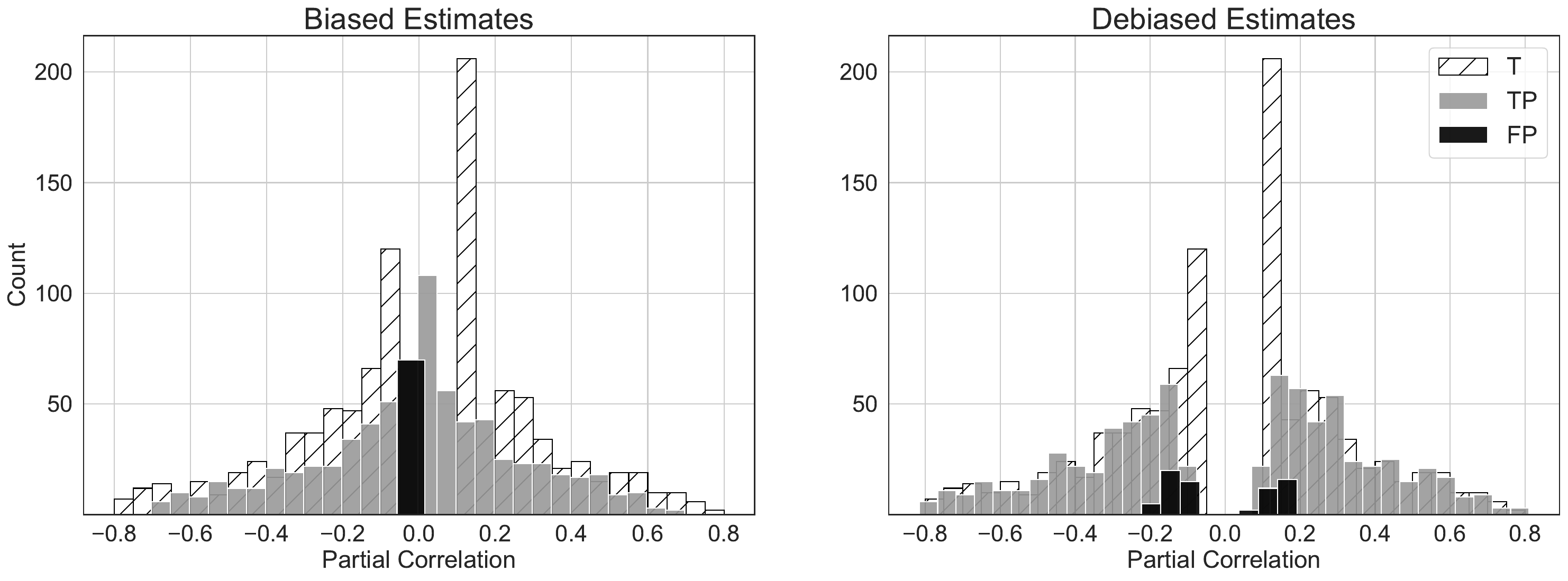}%
}
\\
\subfloat[Scale-free Graph]{%
  \includegraphics[width=0.95\textwidth]{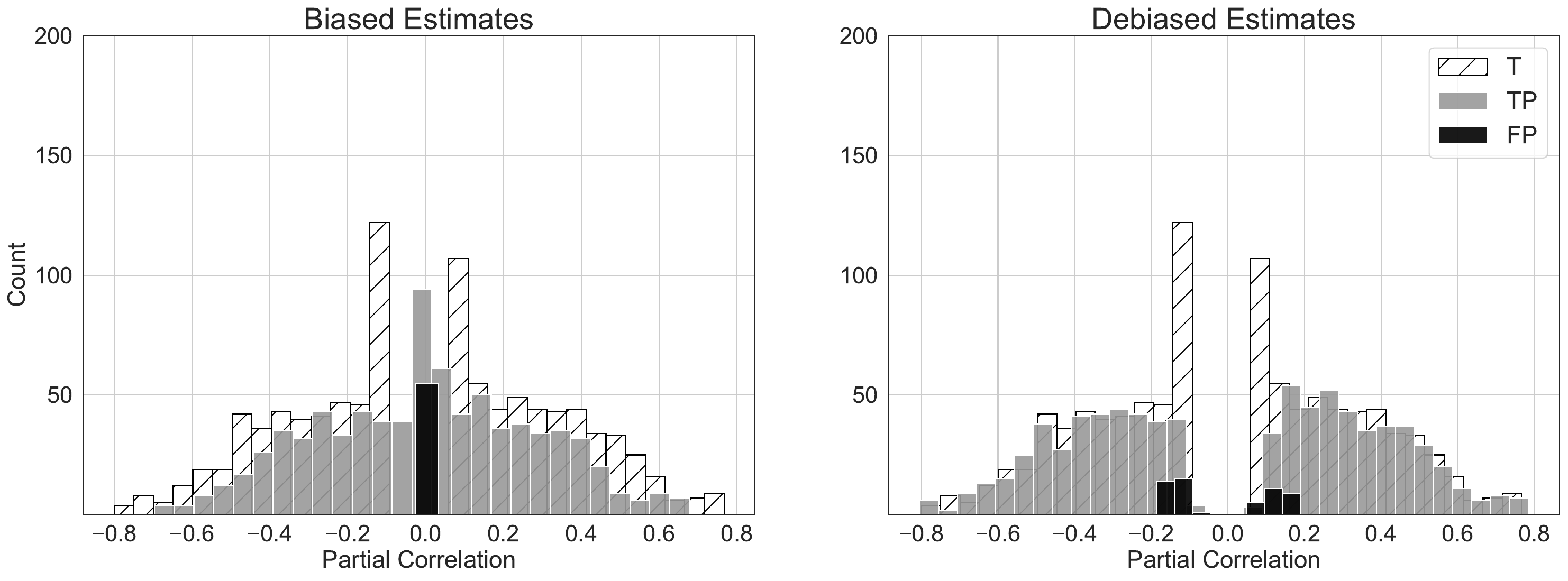}%
}
\\
\subfloat[Erd\"os-R\'enyi Graph]{%
  \includegraphics[width=0.95\textwidth]{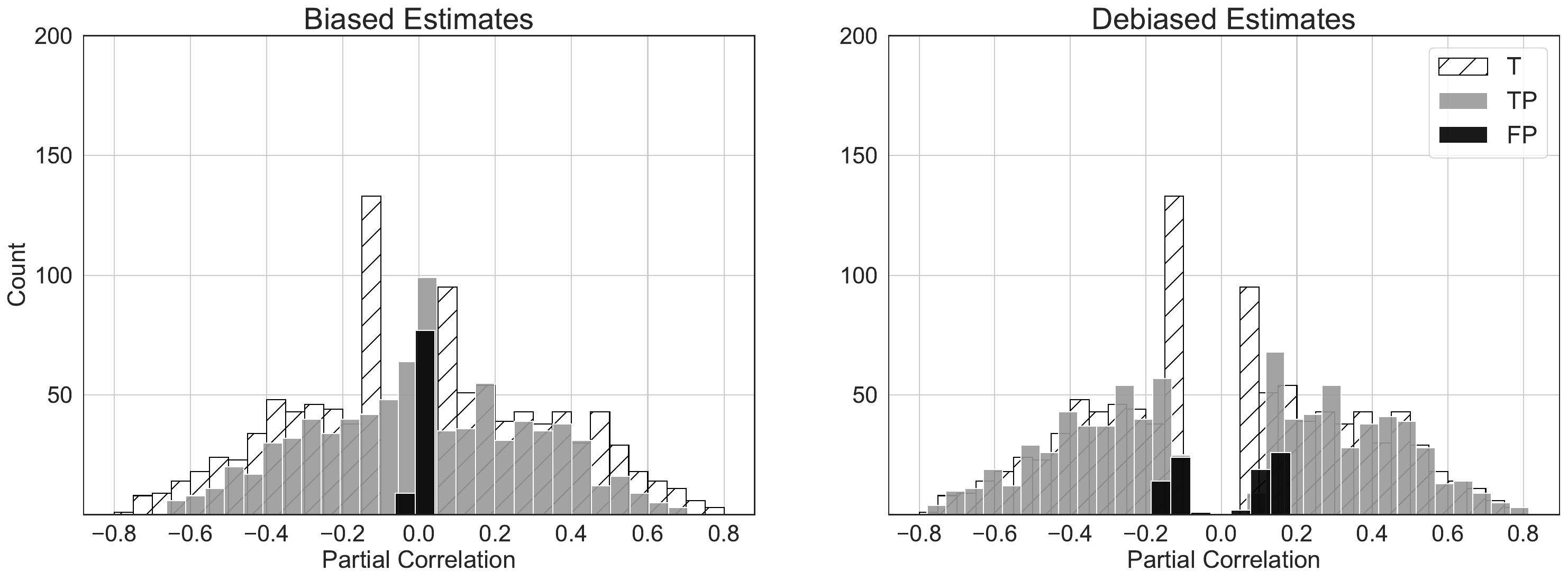}%
}
\caption{Distributions of partial correlation estimates of ACCORD}
\label{fig:debiasing-hist}
\end{figure}

The impact of debiasing refit in ACCORD is demonstrated in \Cref{fig:debiasing-hist} for the suggested simulation settings. Each plot is a histogram showing the true partial correlations (T) and estimated partial correlations. Different colors are used to distinguish the estimated values at correctly detected nonzero locations (true positives or TP) and falsely detected nonzero locations (false positives or FP). The results clearly indicate that debiasing (3.8) consistently improves the accuracy of the estimates at TP locations while effectively controlling the magnitude of the FP, ensuring it remains relatively small. 

\section{Comparison of the computational complexity of graphical model selection methods}\label{sec:addfig}
\begin{table}[h]
\centering
\begin{tabular}{ccccc}
    \toprule
    Method & Algorithm & Flops per iteration & Iteration number & Memory\\
    \midrule
    CLIME & Interior-point method & $O(p^3)$ & $O(\sqrt{p}\log(1/\epsilon))$ & $\Omega(p^2)$ \\
    QUIC & Proximal Newton& $O(p^3)$ & $O(\log\log(1/\epsilon))^\dagger$ & $\Omega(p^2)$\\
    ACCORD & Proximal gradient& $O(np^2)^*$ & $O(\log(1/\epsilon))$ & $\Omega(p^2)^*$\\
    \bottomrule
\end{tabular}
\caption{Complexity comparison of graphical model selection methods. Remarks: $\dagger$, based on local convergence rate;
$*$, computation and intermediate values can be distributed across multiple computational nodes.}
\label{tab:complexity}
\end{table}

\Cref{tab:complexity} compares the computational complexity of ACCORD with other graphical moodel selection methods. 
``Flops per iteration'' denotes the number of floating-point operations required for each iteration. 
``Iteration number'' denotes the number of iterations required to 
reach within $\epsilon>0$ of the optimal value.
``Memory'' indicates the amount of memory required to run the iteration, also known as space complexity; the $\Omega(\cdot)$ refers to the asymptotic lower bound.
The complexity measures for CLIME are based on the estimated complexity of the interior-point method to solve linear programming. 
For QUIC, $O(p)$ flops are needed for each element of the iterate matrix, where other intermediate values are also updated alongside with it. A Cholesky factorization then follows to compute the inverse of the iterate matrix.

When $p \gg n$ and $p$ is massively large as we aim,  the per-iteration complexity of $O(p^3)$ is prohibitive, as we reported in the experiments section. This drawback annuls the attractive locally quadratic convergence rate of QUIC.

In practice, more critical than flop counts is the feasibility of computation in the context of current computer technology. 
First of all, when $p$ is massively large, the memory limitations alone may prevent the algorithm from running on a single computational node, since all of the methods require at least $\gtrsim p^2$ space to store the sample covariance $\bS$ or other intermediate values.
Next, the per-iteration complexity of ACCORD, $O(np^2)$, is dominated by sparse-dense matrix multiplication, which is easy to scale up with multiple computational nodes. On the other hand, the Cholesky factorization involved with QUIC to compute the inverse of a $p\times p$ matrix is not easy to scale, not alone the preceding coordinate descent steps. %

\bibliographystyle{abbrvnat} 
\bibliography{ref}